\documentclass[lettersize,journal]{IEEEtran}
\usepackage{amsmath,amsfonts}
\usepackage{algorithmic}
\usepackage{algorithm}
\usepackage{array}
\usepackage[caption=false,font=normalsize,labelfont=sf,textfont=sf]{subfig}
\usepackage{textcomp}
\usepackage{stfloats}
\usepackage{url}
\usepackage{verbatim}
\usepackage{graphicx}
\usepackage{cite}
\usepackage{xcolor}
\usepackage{soul}
\usepackage{amssymb}
\usepackage{amsthm}
\usepackage{multirow}
\usepackage{booktabs}
\usepackage{tabularx}
\usepackage{mathrsfs}
\newtheorem{theorem}{Theorem}
\newtheorem{lemma}{Lemma}
\newtheorem{proposition}{Proposition}
 
\newtheorem{corollary}{Corollary}
\theoremstyle{remark}
\newtheorem{remark}{Remark}
\newtheorem{assumption}{Assumption}

\usepackage{hyperref}

\begin{document}
	
	\title{FAIR-Pruner: A Flexible Framework for Automatic Layer-Wise Pruning via Tolerance of Difference}
	
	\author{Chenqing Lin, Mostafa Hussien, Chengyao Yu, Bingyi Jing, Ruixing Ming$^*$,\\ Kim Khoa Nguyen$^*$, and Mohamed Cheriet%
		\thanks{Chenqing Lin and Ruixing Ming are with the School of Statistics and Mathematics, Zhejiang Gongshang University, Hangzhou, China (e-mail: 24020040061@pop.zjgsu.edu.cn; rxming@zjgsu.edu.cn).}%
		\thanks{Mostafa Hussien, Kim Khoa Nguyen and Mohamed Cheriet are with the École de technologie supérieure (ÉTS), Université du Québec, Montreal, QC, Canada (e-mail: mostafa.hussien@etsmtl.ca; Mohamed.Cheriet@etsmtl.ca).}%
		\thanks{Chengyao Yu and Bingyi Jing are with the Southern University of Science and Technology, Shenzhen, China (e-mail: 12532239@mail.sustech.edu.cn; jingby@sustech.edu.cn).}%
		\thanks{Corresponding author: Ruixing Ming and Kim Khoa Nguyen.}%
	}

%


\maketitle

\begin{abstract}
Structured pruning is a standard tool for compressing deep neural networks, but
its practical performance depends on how sparsity is allocated across layers.
We propose FAIR-Pruner, a search-free framework for adaptive layer-wise
structured pruning. FAIR-Pruner uses two within-layer rankings: a
removal-oriented signal that proposes candidate units and a protection-oriented
signal that identifies task-sensitive units. Its core component, Tolerance of
Difference (ToD), measures the overlap between the removal prefix and the
protected tail, and uses a shared tolerance level to induce non-uniform pruning
depths across layers. As a default vision instantiation, FAIR-Pruner combines a
Wasserstein-based U-Score for class-conditional unit separability with a
Taylor-based R-Score for task-level sensitivity; the same ToD allocation rule
can also be paired with alternative removal signals. Theoretically, we analyze
ToD through the population R-Score, derive rank-based control of the
high-R-Score mass entering the pruning set, and identify an additive exchange
condition for same-budget comparison with uniform pruning. Experiments on
CIFAR-10, CIFAR-100, SVHN, and ImageNet across VGG, ResNet, DenseNet, ConvNeXt,
and DeiT show strong accuracy--compression trade-offs. Prune-only experiments
on routed-expert Qwen1.5-MoE-A2.7B-Chat further examine architectural
extensibility under matched expert budgets. FAIR-Pruner is released as a
pip-installable open-source package.
\end{abstract}


\section{Introduction}
\label{sec:intro}

\IEEEPARstart{N}{eural} networks have achieved remarkable success across a wide range of applications, but their increasing scale often leads to substantial computational and memory costs~\cite{lecun2015deep}. This has made \emph{model compression} an important practical problem, especially for deployment on resource-limited devices. Among existing compression techniques, pruning is one of the most widely adopted approaches, as it aims to remove units (e.g., neurons, filters, heads, or channels) with minimal contribution to model performance while preserving predictive accuracy as much as possible~\cite{cheng2024survey,xu2020convolutional,he2023structured}.

A central challenge in structured pruning is converting unit-level evidence
into layer-wise pruning budgets.  A pruning criterion may reliably rank units
inside one layer, but this alone does not determine how much sparsity each
layer should receive, and comparing saliency values across heterogeneous layers
can be fragile.  We therefore separate the pruning problem into two connected
questions: how to rank candidate units within each layer, and how to translate
these within-layer rankings into non-uniform layer-wise pruning depths without
architecture search or cross-layer score calibration.

For within-layer unit ranking, existing criteria can be broadly viewed from two
complementary perspectives:
\emph{performance preservation}, where a unit is important if its removal
causes a large task-level loss change~\cite{lecun1989optimal,
	molchanov2019importance}, and \emph{architectural utility}, where importance
is associated with structural or representational properties such as activation
magnitude, geometric redundancy, or feature-map rank~\cite{li2017pruning,
	yeom2021pruning,he2018SFP,he2019filter,lin2020hrank}.  These two views serve
different roles: a task-sensitivity criterion is suitable for identifying risky
deletions, whereas a structural criterion is more natural for ranking candidate
removable units.

The allocation problem is difficult because redundancy is rarely uniformly
distributed across a network.  Uniform pruning applies the same sparsity ratio
to all layers and is simple to implement~\cite{Liu_2017_ICCV,
	lin2020hrank,li2017pruning}.  However, fixed-ratio strategies often
deteriorate under aggressive compression~\cite{blalock2020state,
	ye2018rethinking}.  More adaptive methods,
including regularization-based, reinforcement-learning-based, evolutionary, and
search-based approaches, can produce non-uniform sparsity profiles, but often
require additional optimization, repeated evaluation, or careful tuning
\cite{han2015learning,louizos2017learning,lee2025pruning,he2018amc,
	Yu_2021_ICCV,liu2019metapruning}.  Global-ranking methods provide another
route by thresholding a merged score list across layers
\cite{molchanov2019importance,wang2021neural}, but they require score
values from heterogeneous layers to be comparable on a common scale.  Moreover,
a single scalar score must simultaneously describe how removable a unit appears
and how risky its deletion is, although these two roles may conflict.

To address this problem, we propose \textbf{FAIR-Pruner} (Flexible Automatic
Identification and Removal Pruner), a search-free framework for adaptive
layer-wise structured pruning. Its core criterion, \textbf{Tolerance of
	Difference (ToD)}, turns the conflict between two within-layer rankings into a
layer-wise pruning depth. A removal-oriented signal proposes a prefix of
candidate units, while a protection-oriented signal identifies a
high-sensitivity tail. ToD measures the overlap between these two sets and
allows the removal prefix to grow only while this conflict remains below a
prescribed tolerance. Applying the same tolerance across layers yields
non-uniform pruning depths without architecture search or direct cross-layer
saliency calibration.

In the default vision setting, we instantiate the removal signal by a
Wasserstein-based U-Score that measures class-conditional unit separability, and
instantiate the protection signal by a Taylor-based R-Score that estimates
task-level deletion sensitivity. These choices are practical defaults rather
than necessary components of the framework: ToD only requires a removal ranking
and a protection ranking, and can therefore be paired with other saliency
metrics or architecture-specific signals.

The theory follows this dual-signal view. We define the population R-Score as
the loss change caused by deleting one unit, and show that simultaneous pruning
loss is controlled by singleton deletion sensitivity and a structural
interaction envelope. This decomposition explains why a protection signal alone
is not sufficient and why a separate removal signal is useful. We then show
that the ToD constraint gives an explicit rank-based envelope for the
high-R-Score mass admitted into the removal set, with sharpness over the
nonempty feasible class, and identify an additive compatibility condition for
same-budget comparison with uniform pruning.

\begin{figure*}
	\centering
	\includegraphics[width=\linewidth]{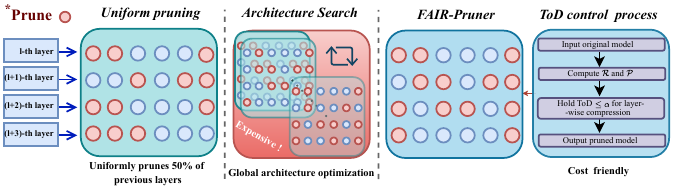}
	\caption{Comparison of three layer-wise pruning paradigms. Uniform pruning applies a fixed sparsity ratio across layers and is computationally cheap, but ignores heterogeneous redundancy. Architecture-search-based methods optimize layer widths more flexibly, but at substantial computational cost. FAIR-Pruner uses ToD to adaptively determine layer-wise sparsity from a preset parameter $\alpha$, achieving search-free global compression control with negligible overhead.}
	\label{fig:framework_compare}
\end{figure*}

Figure~\ref{fig:framework_compare} summarizes the position of FAIR-Pruner among
representative pruning paradigms. Uniform pruning is cheap but fixes the same
sparsity ratio for every layer; architecture-search methods are adaptive but
expensive; FAIR-Pruner lies between these extremes by using score conflict to
obtain adaptive layer-wise sparsity while remaining search-free.

The contribution of FAIR-Pruner is therefore a framework-level allocation
principle: given any removal ranking and protection ranking inside each layer,
ToD converts their conflict into an adaptive sparsity profile. This separates
the design of pruning scores from the design of layer-wise budgets and makes the
same allocation mechanism reusable across different saliency scores,
architectures, and deployment constraints.

The main contributions of this work are summarized as follows.
\begin{itemize}
	\item We propose FAIR-Pruner, a search-free framework for adaptive layer-wise
	structured pruning. Its core mechanism, Tolerance of Difference (ToD),
	separates within-layer removal ranking from cross-layer sparsity allocation
	and induces non-uniform pruning rates by coordinating a removal-oriented
	signal with a protection-oriented signal.
	
	\item We formulate ToD as a modular allocation principle rather than a
	score-specific heuristic. In the default vision setting, FAIR-Pruner uses a
	Wasserstein-based U-Score as the removal signal and a Taylor-based R-Score
	as the protection signal. The same ToD rule can also be paired with
	alternative removal signals, including L1-norm, FPGM, and soft activation,
	and consistently improves over their corresponding uniform pruning variants
	in our matched-budget experiments.
	
	\item We provide an R-Score-based theoretical analysis of ToD. The population
	R-Score \(R_j^{(l)}\) is defined as the loss change caused by removing one
	unit. We show that simultaneous pruning loss can be bounded by the R-Scores
	of the pruned units and a structural interaction envelope, prove that ToD
	controls the high-R-Score mass entering the pruning set, and derive a
	ToD-induced structural loss envelope for the selected pruning set. We further
	give an additive exchange condition under which a ToD-induced allocation has
	no larger surrogate pruning cost than any same-budget allocation, including
	uniform allocation.
	
	\item We conduct extensive experiments across classical CNNs, modern
	block-structured and transformer architectures, and routed-expert Mixture-of-Experts (MoE)
	language models. Across VGG, AlexNet, ResNet, DenseNet, ConvNeXt, and
	DeiT-B, FAIR-Pruner achieves competitive accuracy--compression trade-offs
	against recent structured pruning methods. On Qwen1.5-MoE, prune-only
	experiments examine how the same allocation principle transfers to
	routed experts. We also evaluate practical efficiency beyond FLOPs, including
	CPU latency, GPU memory, CPU resident memory, and implementation-matched
	inference carbon estimates, and release FAIR-Pruner as a pip-installable
	open-source package.
\end{itemize}

\section{Related Work}
\label{sec:related}

This section compares FAIR-Pruner with existing pruning methods from two perspectives: layer-wise sparsity allocation and unit importance estimation.

\subsection{Layer-wise Sparsity Allocation Strategies}

A major line of pruning research focuses on how to allocate sparsity across layers. One common strategy is to apply a uniform pruning ratio to all layers~\cite{Liu_2017_ICCV,li2017pruning,lin2020hrank}. Although computationally simple, uniform pruning ignores the fact that different layers often exhibit different redundancy and functional importance, and can therefore perform poorly under aggressive compression~\cite{blalock2020state,ye2018rethinking}.

To overcome this limitation, several methods explicitly seek non-uniform layer-wise allocation. Search-based approaches, including reinforcement-learning-based methods~\cite{he2018amc,Yu_2021_ICCV} and heuristic or evolutionary strategies~\cite{liu2019metapruning}, optimize layer-wise sparsity through global search. These methods can be effective, but they often incur substantial computational cost and require repeated evaluation during the search process. Differentiable approaches such as DMCP~\cite{guo2020dmcp} reduce search cost by introducing continuous mask relaxation, but still require additional optimization machinery and careful hyperparameter tuning.

Other methods address layer-wise allocation from more direct global criteria.
ITPruner~\cite{zheng2025information} estimates layer-wise redundancy through
normalized HSIC and determines layer widths by solving a resource-constrained
architecture selection problem; its theoretical motivation is tied to an
information-bottleneck view of inter-layer representation dependence.  Taylor
pruning, such as Molchanov et al.~\cite{molchanov2019importance}, can also
induce non-uniform sparsity by applying a global threshold to gradient-based
saliency scores.  In contrast, FAIR-Pruner does not assign layer importance
solely from inter-layer dependence, nor does it allocate sparsity through a
single globally merged saliency ranking.  ToD only requires a removal-oriented
ranking and a protection-oriented sensitivity ranking, and determines how far
the former can proceed before it conflicts with the latter.  This
conflict-control view makes the framework naturally applicable to heterogeneous
pruning units, including channels, hidden dimensions, attention heads, and
routed experts.

\subsection{Unit Importance Estimation}

Estimating unit importance is a central problem in neural network pruning. Existing criteria can be broadly grouped into three categories. \emph{Magnitude-based} methods prune units with small norms and are attractive for their simplicity and efficiency, but may overlook the functional role of a unit in prediction~\cite{li2017pruning,han2015learning}. \emph{Representation-oriented} methods evaluate importance through activation statistics, geometric redundancy, or feature-map rank, aiming to identify units that contribute little to the internal representation structure~\cite{he2019filter,lin2020hrank}. \emph{Performance-oriented} methods estimate the loss increase caused by removing a unit, for example through saliency measures or Taylor approximations, and are therefore more directly connected to task accuracy~\cite{lecun1989optimal,molchanov2017pruning,dong2017learning}.

FAIR-Pruner is related to these lines of work, but differs in two important aspects. First, instead of relying on a single importance signal, it explicitly combines two complementary perspectives: architectural utility and task sensitivity. Second, these signals are not used only for within-layer ranking; they are further integrated through ToD to determine adaptive layer-wise pruning rates. Thus, the main novelty of FAIR-Pruner lies not only in proposing a new importance criterion, but also in providing a unified framework for converting complementary importance signals into adaptive non-uniform pruning decisions. 

Recent work has also recognized that pruning decisions may benefit from
multiple types of evidence. Wang and Jiang~\cite{wang2026neural} propose
the Neural Differentiation Index (NDI), which integrates spectral diversity,
entropy-derived informativeness, and second-order sensitivity into a unified
neuron-level importance score. This is aligned with our motivation that
structural redundancy and task sensitivity are complementary. However, the
mathematical form of the pruning rule is different. NDI follows a pointwise
scalarization paradigm: each unit is assigned a fused importance score and
pruning is performed by ranking or thresholding this score. FAIR-Pruner does
not construct such a fused score. For each layer, the U-Score defines a
removal prefix, the R-Score defines a protected high-sensitivity prefix, and
ToD selects the largest removal prefix whose overlap with the protected set is
below the tolerated conflict level. Therefore, ToD is a rank-based
set-level allocation rule rather than a special case of pointwise scalar
score fusion. This design avoids choosing scalar fusion weights between
heterogeneous evidence sources and directly turns their conflict into a
layer-wise pruning depth.

The default U-Score in our vision experiments is one transport-based
instantiation of the removal signal. Wasserstein-type distances have been used
in pruning to compare channel or feature
distributions~\cite{shen2020cpot,duan2020channel,you2023swap}, and sliced Wasserstein
distances provide scalable approximations for high-dimensional
distributions~\cite{bonneel2015sliced}. In FAIR-Pruner, this distance is used more modestly:
it measures class-conditional unit separability and supplies a within-layer
removal ranking, while ToD remains the allocation rule.

\section{Methodology}
\label{sec:proposed-method}
Consider a trained network \(\hat f\) for a \(K\)-class classification task.
Let \(X\) denote the input and \(Y\in[K]:=\{1,2,\cdots,K\}\) its class label. For each class
\(k\in[K]\), let \(Z_k\) denote a random variable following the conditional
distribution of \(X\mid(Y=k)\). We assume access to a labeled pruning dataset
\(\{(x_i,y_i)\}_{i=1}^n\), and write \(n_k\) for the number of pruning samples
with label \(k\).
Let \( L\) denote the set of prunable layers or blocks.  For each
\(l\in L\), let \(\mathcal U^{(l)}=[J^{(l)}]\) be the set of
prunable units in that layer.  A unit may correspond to a convolutional
channel, an attention head, a hidden dimension in an MLP block, or a routed
expert in an MoE layer, depending on the architecture.  We write
\(\mathcal U=\{(l,j):l\in L,\,j\in\mathcal U^{(l)}\}\) for the collection
of all prunable units.

Each unit \((l,j)\) is assigned two scores.  The removal-oriented score
\(U_j^{(l)}\) measures how dispensable the unit appears from a structural or
representational perspective; smaller values indicate stronger candidates for
removal.  The protection-oriented score \(R_j^{(l)}\) measures the task-level
sensitivity of removing the unit; larger values indicate units that should be
protected.  Their empirical estimates computed from the pruning dataset are
written as \(\widehat U_j^{(l)}\) and \(\widehat R_j^{(l)}\), respectively.

\begin{figure*}
	\centering
	\includegraphics[width=\linewidth]{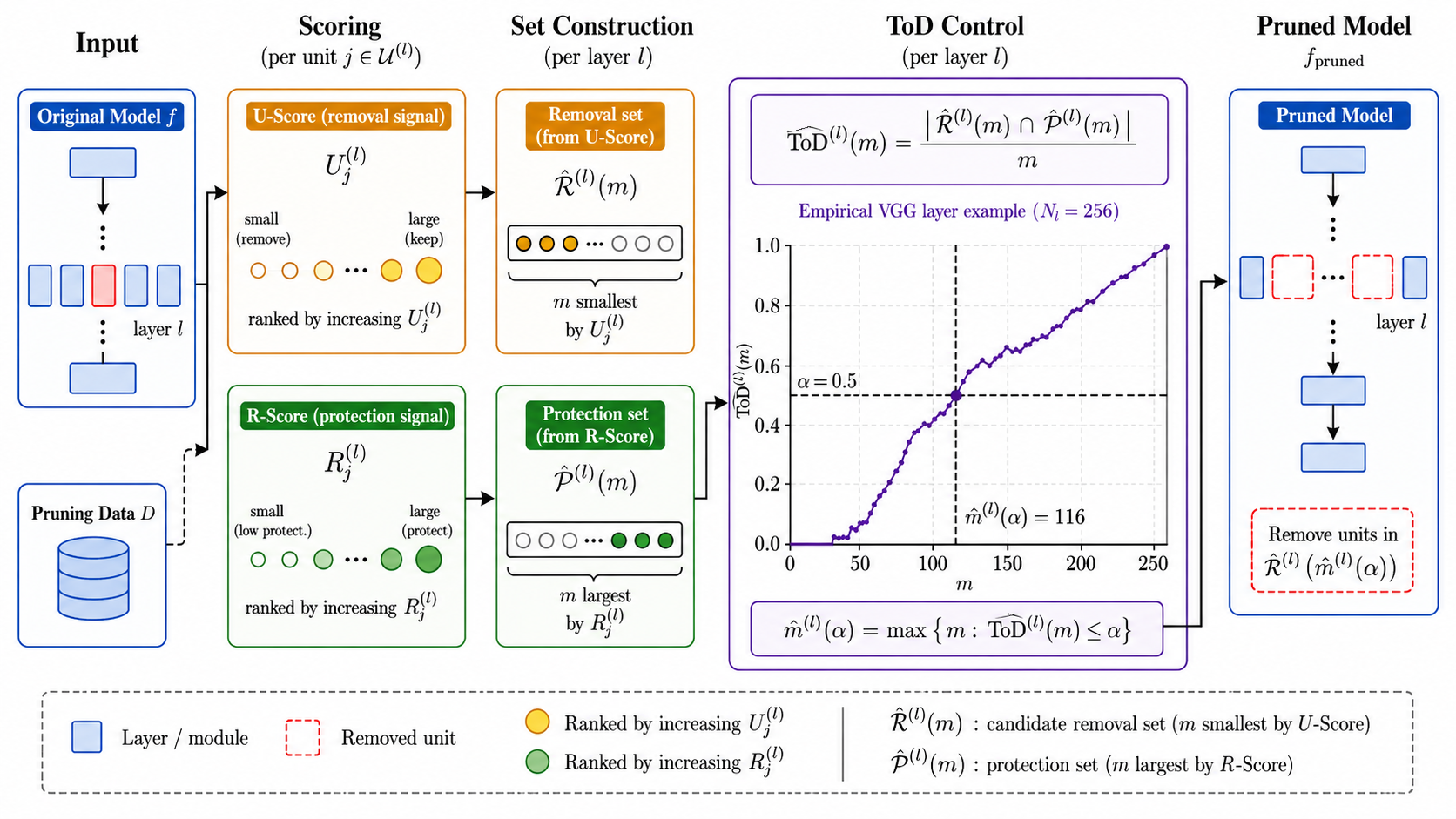}
	\caption{
		\textbf{Overview of the ToD-based pruning decision in one layer.} 
		The removal-oriented score ranks units from most removable to least removable,
		while the protection-oriented score ranks units from most sensitive to least sensitive.
		For each candidate pruning depth \(m\), the removal prefix contains the \(m\)
		lowest-removal-score units and the protected tail contains the \(m\)
		highest-protection-score units. The empirical ToD
		\(\widehat{\mathrm{ToD}}^{(l)}(m)\) is the normalized overlap between these two sets.
		FAIR-Pruner scans all candidate depths and selects the largest feasible depth
		whose conflict is no greater than the tolerance level \(\alpha\); only the units
		in the corresponding removal prefix are pruned.
	}
	\label{fig:algorithm-overview}
\end{figure*}

Building upon these notations, we introduce FAIR-Pruner, which achieves non-uniform layer-wise pruning by controlling the Tolerance of Difference (ToD).
Figure~\ref{fig:algorithm-overview} provides a high-level overview of FAIR-Pruner.

\subsection{Tolerance of Difference (ToD)-based Pruning}

Unit importance in a neural network is inherently multi-dimensional.
We characterize each unit from two complementary perspectives:
the Utilization Score (U-Score), which provides a removal-oriented
signal within a layer, and the Reconstruction Score (R-Score),
which provides a protection-oriented signal by measuring the
task-level sensitivity magnitude associated with removing that unit.
The formal definitions of the U-Score and R-Score are given in
Subsections~\ref{subsec:u-s} and~\ref{subsec:r-s}, respectively.

\begin{figure*}
	\centering
	
	\subfloat[AlexNet: a shallow layer close to the input]{%
		\includegraphics[width=0.48\linewidth]{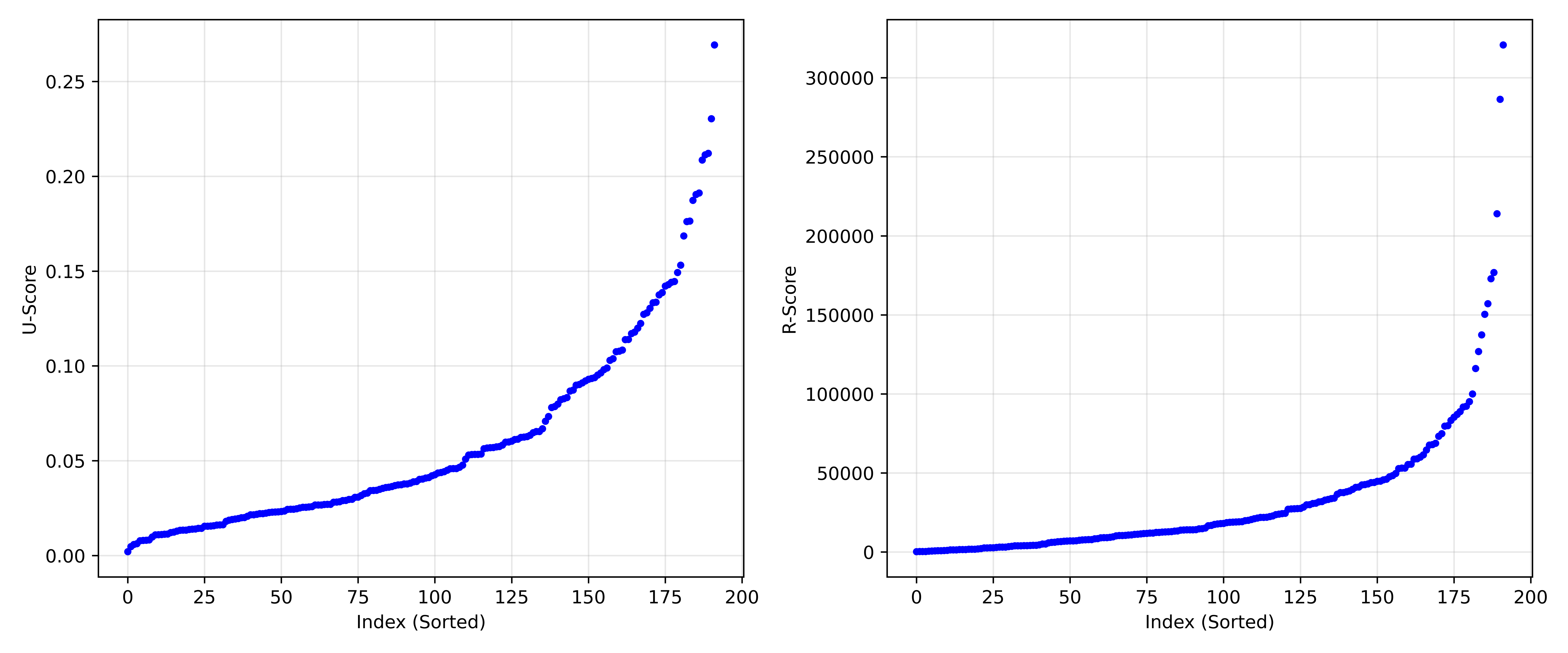}%
		\label{fig:score-alex-shallow}
	}
	\hfill
	\subfloat[VGG-16: a shallow layer close to the input]{%
		\includegraphics[width=0.48\linewidth]{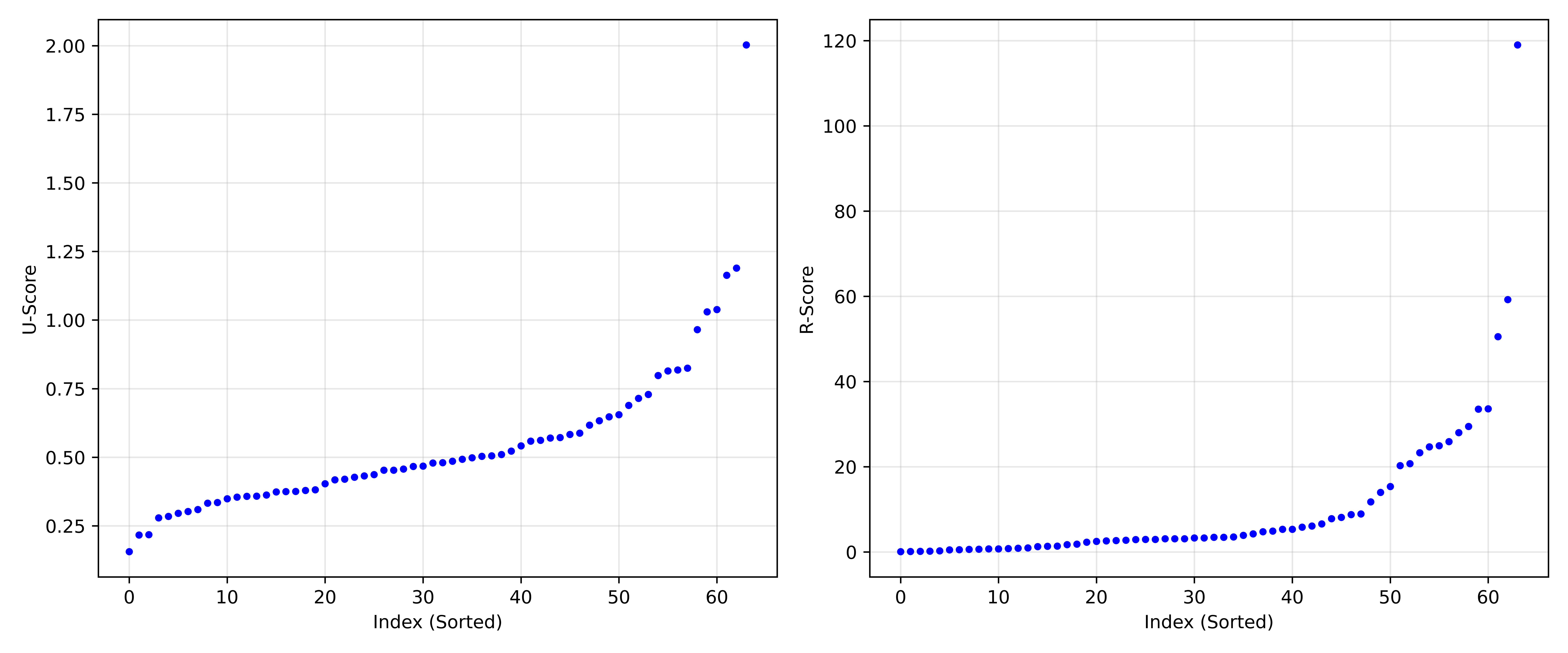}%
		\label{fig:score-vgg-shallow}
	}\\[2mm]
	
	\subfloat[AlexNet: a deep layer close to the output]{%
		\includegraphics[width=0.48\linewidth]{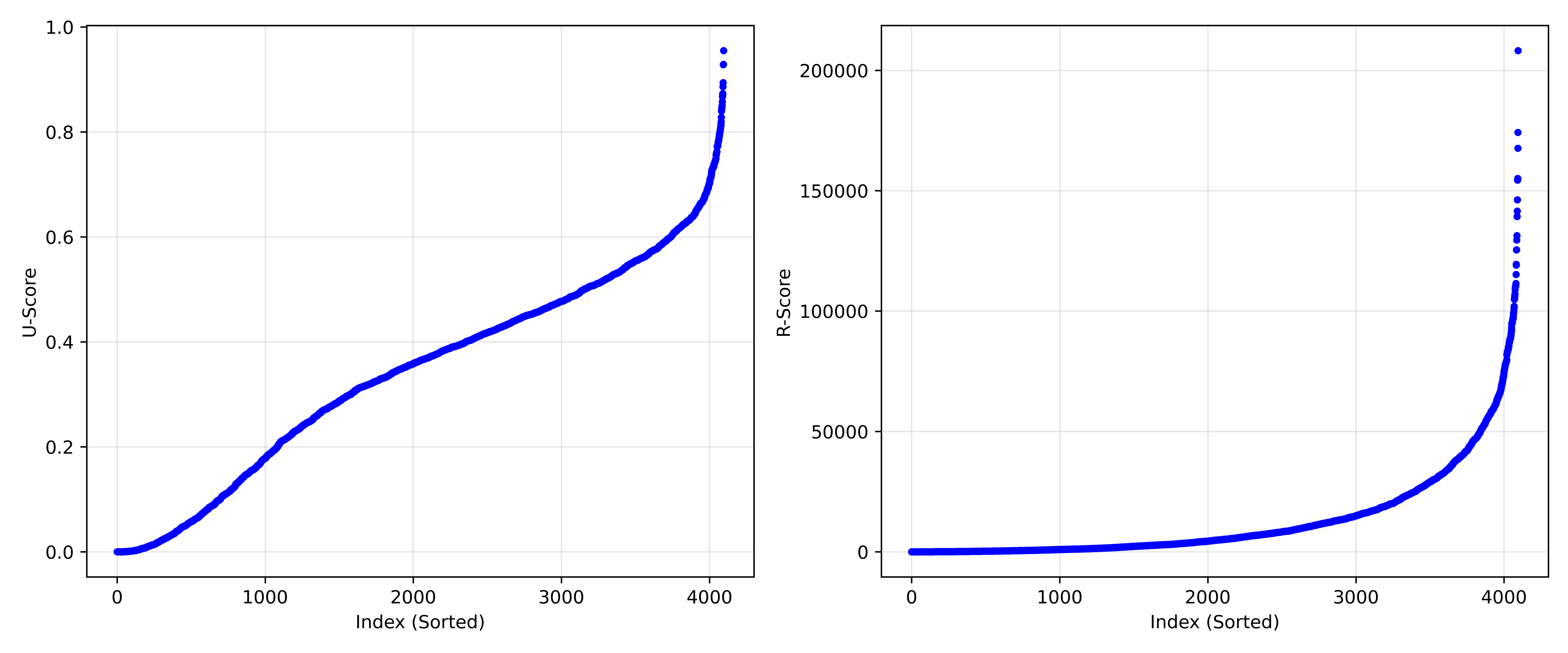}%
		\label{fig:score-alex-deep}
	}
	\hfill
	\subfloat[VGG-16: a deep layer close to the output]{%
		\includegraphics[width=0.48\linewidth]{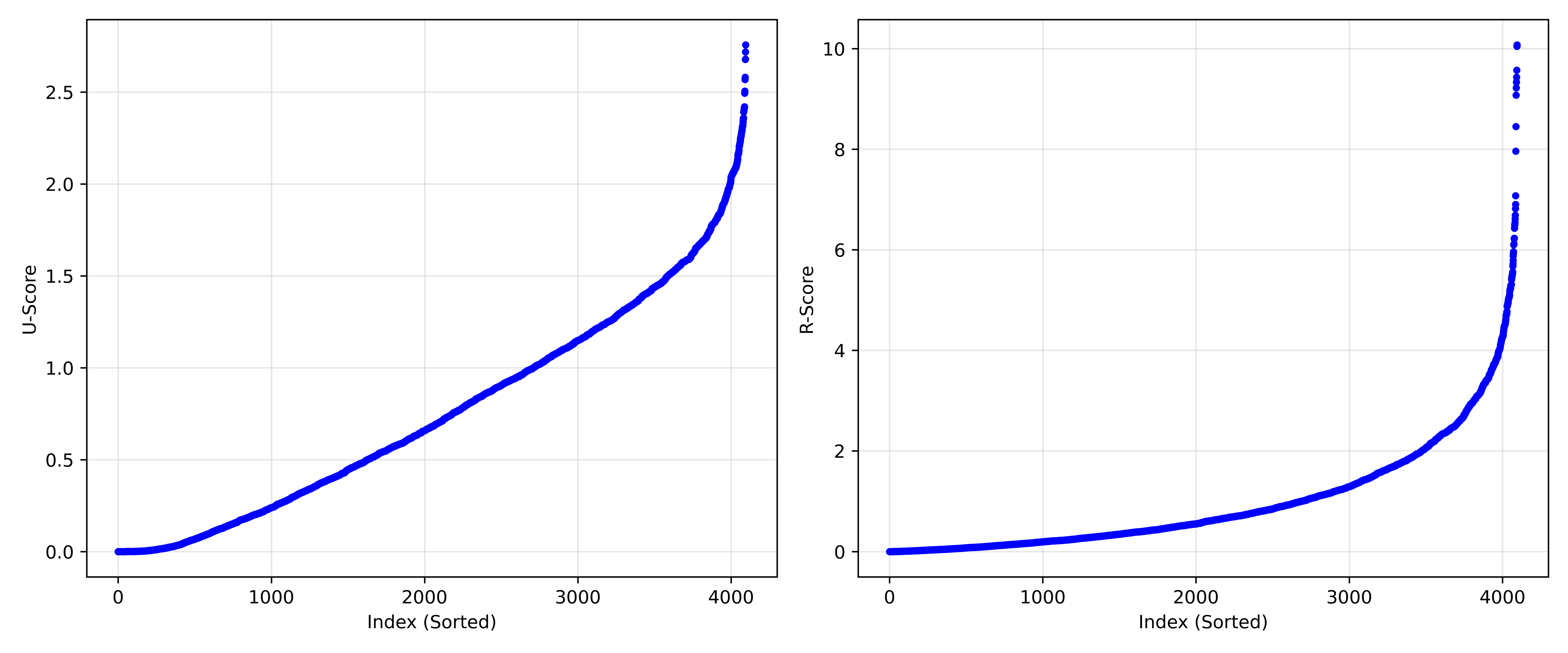}%
		\label{fig:score-vgg-deep}
	}
	
	\caption{Comparison of U-Score (left panel within each subfigure) and R-Score (right panel within each subfigure) across representative shallow and deep layers of AlexNet and VGG-16. Across both architectures, U-Scores exhibit smoother and more globally discriminative ranking structures, while R-Scores mainly separate a small subset of highly influential units. This complementary behavior motivates using U-Scores to identify redundant units and R-Scores to protect globally critical ones in the proposed ToD framework.}
	\label{fig:score-distributions}
\end{figure*}

The use of two scores is motivated by the empirical asymmetry shown in
Figure~\ref{fig:score-distributions}. Intuitively, the two scores evaluate
unit importance from different and complementary perspectives. Across
representative shallow and deep
layers, R-Scores mainly separate a small protected tail of highly influential
units, but are often much less discriminative over the larger population of
low-sensitivity units. This plateau-like behavior is expected under finite
pruning data, gradient noise, and the fact that many units have similarly small
individual deletion effects. By contrast, U-Scores exhibit a smoother ranking
structure over the unit population. We therefore use the R-Score as a
protection signal and the U-Score as the removal ranking.

This motivates the proposed Tolerance of Difference (ToD)
criterion. ToD quantifies the overlap between a removal set induced by the U-Score and a protection set induced by the R-Score within each layer. A small overlap indicates that the units regarded as removable by the architectural signal are largely different from those regarded as sensitive by the task-level signal, and hence a deeper pruning is acceptable. A large overlap indicates substantial conflict between the two views and therefore leads to a more conservative pruning. The role of ToD should be understood at the layer-allocation level:
different layers may exhibit different degrees of conflict between the removal and protection rankings, and ToD converts these conflict patterns
into layer-specific pruning depths.

In the actual pruning procedure, all layer-wise decisions are made from the empirical score estimates computed on the pruning dataset. For each layer $l\in L$ and pruning depth $m\in [J^{(l)}]$, let $\widehat{\mathcal R}^{(l)}(m)$ denote the empirical removal set containing the $m$ units with the smallest empirical U-Scores, and let $\widehat{\mathcal P}^{(l)}(m)$ denote the empirical protection set containing the $m$ units with the largest empirical R-Scores. We set $\widehat{\mathcal R}^{(l)}(0)=\widehat{\mathcal P}^{(l)}(0)=\varnothing$. We quantify the conflict between these two sets by the empirical \textbf{Tolerance of Difference}
\begin{equation}
	\label{eq:tau}
	\widehat{\mathrm{ToD}}^{(l)}(m)
	=
	\frac{\bigl|\widehat{\mathcal R}^{(l)}(m)\cap \widehat{\mathcal P}^{(l)}(m)\bigr|}{\max\{m,1\}}.
\end{equation}
The denominator term $\max\{m,1\}$ avoids division by zero and allows the convention $\widehat{\mathrm{ToD}}^{(l)}(0)=0$.
The empirical ToD therefore quantifies the normalized conflict, or overlap,
between the removal and protection views.

Given the preset level $\alpha \in (0,1)$, the number of units to be pruned in layer $l$ is automatically determined as
\begin{equation}
	\label{eq:hat_m}
	\widehat m^{(l)}(\alpha)
	=
	\max\bigl\{ m \in \{0,1,\ldots,J^{(l)}\} : \widehat{\mathrm{ToD}}^{(l)}(m)\le \alpha \bigr\}.
\end{equation}
A larger value of $\alpha$ allows more overlap with the protected R-Score tail
and therefore corresponds to a larger pruning-risk tolerance.
The choice of taking the maximum feasible $m$ ensures the monotonicity of the selected pruning depth with respect to $\alpha$; this property is formalized in Subsection~\ref{subsec:alpha_monotonicity}.

\subsection{Utilization Score}
\label{subsec:u-s}
In FAIR-Pruner, the U-Score is a within-layer removal signal: units with small
U-Scores are proposed as candidates for pruning.  The theory in
Subsection~\ref{subsec:theory} only requires such a signal to order units so
that low-score prefixes are structurally mild to remove.  In vision models, we
instantiate this idea by class-conditional separability.  If a unit has nearly
the same output distribution across classes, then it carries little
class-discriminative structure and is a natural pruning candidate; if its
responses vary strongly across classes, it is treated as more useful and moved
later in the removal ranking.

Denote the output of the $j$-th unit in the $l$-th layer corresponding to $Z_k$ by $O^{(l)}_{j}(Z_k)$, where $j \in [J^{(l)}]$, $l\in L$, and $k \in [K]$. Let $d(\cdot, \cdot)$ represent a distance between probability distributions of random variables. We define the population U-Score of the $j$-th unit in the $l$-th layer by
\begin{equation}
	U_{j}^{(l)} := \sup_{\substack{k_1, k_2 \in [K] \\ k_1 \neq k_2}} 
	d\big(O^{(l)}_{j}(Z_{k_1}), O^{(l)}_{j}(Z_{k_2})\big).
	\label{eq:utility-score}
\end{equation}
We use the Wasserstein distance to quantify the dissimilarity between the output distributions of units. Specifically, the Wasserstein distance between random variables $P$ and $Q$ is defined as
\begin{equation*}
	d(P, Q) = \int_0^1 |F_P^{-1}(z) - F_Q^{-1}(z)| \, \mathrm{d}z,
\end{equation*}
where $F_{X}^{-1}(\cdot)$ represents the inverse of the cumulative distribution function of random variable $X$. For convolutional layers where the output of a channel is a matrix, we flatten the output and then use sliced Wasserstein distance~\cite{bonneel2015sliced}.

Below we provide a computable version of the U-Score. For any $l\in L$ and $j\in[J^{(l)}]$, let
\begin{equation}
	\widehat U_j^{(l)}
	:=
	\sup_{\substack{k_1\neq k_2 \in [K]}}
	d\bigl(\widehat{O}^{(l)}_{j, n_{k_1}}, \widehat{O}^{(l)}_{j, n_{k_2}}\bigr)
\end{equation}
be the empirical estimator of $U_j^{(l)}$. Under mild moment conditions, the empirical U-Score consistently estimates its population counterpart. The details of computing the empirical score vector \(\{\widehat U_j^{(l)}\}_{j=1}^{J^{(l)}}\) are given in Algorithm~\ref{algo:US}.

\begin{algorithm}[t]
	\caption{Wasserstein-based U-Score}
	\label{algo:US}
	\begin{algorithmic}[1]
		\REQUIRE Model $\hat{f}$; pruning dataset $\{x_i,y_i\}_{i=1}^n$; $l \in L$.
		\ENSURE Empirical Utilization Scores \(\{\widehat U_j^{(l)}\}_{j=1}^{J^{(l)}}\).
		\STATE Collect $\widehat{F}_{O^{(l)}_{j},n_k},k\in [K]$ by using $\hat{f}$ and $\{x_i,y_i\}_{i=1}^n$.
		\STATE $\widehat U_j^{(l)}\gets 0$, $j\in [J^{(l)}]$.
		\FOR{$k_1 = 2$ to $K$}
		\FOR{$k_2 = 1$ to $k_1-1$}
		\STATE $\widehat U_j^{(l)}\gets \max\left\{d(\widehat{O}^{(l)}_{j, n_{k_1}}, \widehat{O}^{(l)}_{j, n_{k_2}}), \widehat U_j^{(l)}\right\},\ j\in [J^{(l)}]$.
		\ENDFOR
		\ENDFOR
		\RETURN \(\{\widehat U_j^{(l)}\}_{j=1}^{J^{(l)}}\).
	\end{algorithmic}
\end{algorithm}

\subsection{Reconstruction Score}
\label{subsec:r-s}

Estimating the loss change caused by removing a unit traces back to
Optimal Brain Damage~\cite{lecun1989optimal}, which framed pruning
through a Taylor expansion of the training loss. In FAIR-Pruner, the R-Score is
the absolute population loss change caused by removing one unit. The
finite-deletion analysis in Subsection~\ref{subsec:theory} shows that the sum of
R-Scores is the first term controlling simultaneous pruning loss, and further
explains the additional interaction terms that arise when multiple units are
pruned together.

Let $\hat f^{(l)}_{-j}$ denote the model obtained after removing the
$j$-th unit in the $l$-th layer of $\hat f$. We define the population
R-Score as the loss change caused by deleting that unit:
\begin{equation}
	R_j^{(l)} :=
	\left|
	\mathbb E\!\left[
	\mathcal{L}(\hat f^{(l)}_{-j}(X),Y) - \mathcal{L}(\hat f(X),Y)
	\right]
	\right|,
\end{equation}
which measures the population task-level sensitivity magnitude
induced by removing that unit.

In practice, we estimate this quantity through a Taylor approximation
of the loss change. Following the classical pruning literature
\cite{lecun1989optimal,molchanov2019importance,molchanov2017pruning,
	hassibi1992second,wang2019eigendamage}, we adopt the first-order
Taylor approximation as the default implementation of the R-Score,
mainly because it can be computed efficiently through a single
backward pass and is substantially cheaper than second-order
alternatives. This choice should not be interpreted as claiming that
the first-order approximation is always the most accurate. Higher-order
Taylor approximations can also be incorporated into the same framework,
and, as shown in Appendix~\ref{app:taylor_order}, second-order variants
can in some cases yield better post-pruning accuracy. We therefore view
the first-order approximation as a practical default, while treating
higher-order variants as accuracy-oriented alternatives within the same
framework. More generally, any fast estimator of the same singleton deletion
loss quantity can be used in this role, provided that it yields a reliable
protection ranking.

At the theoretical level, however, our main consistency result is not intended as an approximation theorem for a specific Taylor truncation order. Instead, it is a stability result for the ToD-based pruning rule. For this reason, the theorem in Subsection~\ref{subsec:theory} only requires that the resulting empirical R-Score consistently estimates its population counterpart. In this way, the analysis covers first-order as well as higher-order Taylor-based implementations, provided that the approximation is sufficiently accurate.

The first-order estimator used in our implementation is as follows.
Let \(\theta_j^{(l)}\) collect all parameters associated with unit \((l,j)\),
for example the incoming and outgoing weights of a channel, the projection
parameters of an attention head, or the routed-expert parameters of an MoE
expert.  Removing the unit can be viewed as the block perturbation
\(\theta_j^{(l)}\mapsto 0\), whose first-order Taylor loss variation is
\(-\langle \nabla_{\theta_j^{(l)}}\mathcal L_i,\theta_j^{(l)}\rangle\) for sample
\(i\).  We therefore use
\begin{equation}
	\widehat R_j^{(l)}
	:=\left|
	\frac{1}{n}\sum_{i=1}^n
	\left\langle
	\nabla_{\theta_j^{(l)}}\mathcal L(\hat f(x_i),y_i),
	\theta_j^{(l)}
	\right\rangle
	\right|.
\end{equation}
Equivalently, for tensor-valued parameter blocks, the inner product denotes the
sum of elementwise products over all tensors in the block.
A large R-Score suggests that removing the $j$-th unit may induce
a larger task-level loss variation in magnitude, and such units are
therefore treated as candidates for protection. The details of calculating the empirical R-Score vector
\(\{\widehat R_j^{(l)}\}_{j=1}^{J^{(l)}}\) are given in Algorithm~\ref{algo:RS}.

\begin{algorithm}[t]
	\caption{First-order Taylor approximation R-Score}
	\label{algo:RS}
	\begin{algorithmic}[1]
		\REQUIRE Model $\hat{f}$; pruning dataset $\{x_i,y_i\}_{i=1}^n$; $l \in L$.
		\ENSURE Empirical Reconstruction Scores \(\{\widehat R_j^{(l)}\}_{j=1}^{J^{(l)}}\).
		\STATE $\widehat R_j^{(l)}\gets 0$, $j\in [J^{(l)}]$.
		\FOR{$i = 1$ to $n$}
		\STATE $\widehat R_j^{(l)}\gets \widehat R_j^{(l)}+\left\langle \nabla_{\theta_j^{(l)}}\mathcal L_i,\theta_j^{(l)}\right\rangle$.
		\ENDFOR
		\STATE $\widehat R_j^{(l)} \gets |\widehat R_{j}^{(l)}|/n$.
		\RETURN \(\{\widehat R_j^{(l)}\}_{j=1}^{J^{(l)}}\).
	\end{algorithmic}
\end{algorithm}

\subsection{Pruning Strategy}

We now present the formal implementation of FAIR-Pruner. Building upon~\eqref{eq:tau} and~\eqref{eq:hat_m}, we describe how the empirical U-Scores and R-Scores jointly determine the pruning decision at each layer. For each candidate pruning depth $m\in[J^{(l)}]$ in the $l$-th layer, define the empirical \textbf{R}emoval and \textbf{P}rotection sets by
\begin{equation}
	\begin{aligned}
		&\widehat{\mathcal R}^{(l)}(m)
		=
		\left\{
		k\in[J^{(l)}]:
		\widehat U_k^{(l)}
		\le
		\widehat U_{(m)}^{(l)}
		\right\},\\
		&\widehat{\mathcal P}^{(l)}(m)
		=
		\left\{
		k\in[J^{(l)}]:
		\widehat R_k^{(l)}
		\ge
		\widehat R_{(J^{(l)}-m+1)}^{(l)}
		\right\},
	\end{aligned}
\end{equation}
where
\begin{equation}
	\widehat U_{(1)}^{(l)}\le \cdots \le \widehat U_{(J^{(l)})}^{(l)},
	\qquad
	\widehat R_{(1)}^{(l)}\le \cdots \le \widehat R_{(J^{(l)})}^{(l)}
\end{equation}
denote the order statistics of the empirical U-Scores and R-Scores in layer $l$, respectively. The empirical ToD level of $\widehat{\mathcal R}^{(l)}(m)$ and $\widehat{\mathcal P}^{(l)}(m)$ is then computed by~\eqref{eq:tau}. Finally, the number of units to prune is determined by~\eqref{eq:hat_m}. This criterion adaptively adjusts the pruning depth according to the tolerated overlap level $\alpha$, yielding a data-dependent compression--accuracy trade-off in practice.

In implementation, all rankings are computed by a stable deterministic
tie-breaking rule: units are sorted lexicographically by
\((\widehat U_j^{(l)},j)\) for the removal ranking and by
\((-\widehat R_j^{(l)},j)\) for the protection ranking. This rule makes
the empirical sets uniquely defined even when numerical ties occur.

In summary, the pruning decision is governed by three key components: (1) the \emph{Utilization Score}, ranking candidate removable units; (2) the \emph{Reconstruction Score}, protecting task-sensitive units; and (3) the \emph{ToD control}, which reconciles the two perspectives through the preset level $\alpha$.

\begin{algorithm}[t]
	\caption{FAIR-Pruner}
	\label{algorithm:FAIR}
	\begin{algorithmic}[1]
		\REQUIRE Model $\hat{f}$; pruning dataset $\{x_i,y_i\}_{i=1}^n$; the index set of layers $L$; preset ToD level $\alpha$.
		\ENSURE The pruned model $\hat f^{p}$.
		\STATE Compute \(\{\widehat U_j^{(l)}\}_{j=1}^{J^{(l)}}\) and \(\{\widehat R_j^{(l)}\}_{j=1}^{J^{(l)}}\), \(l\in L\), through Algorithms~\ref{algo:US} and~\ref{algo:RS}.
		\FOR{$l \in L$}
		\STATE $\widehat m^{(l)}(\alpha) \gets 0$.
		\FOR {$m = 1$ to $J^{(l)}$}
		\IF {$\widehat{\mathrm{ToD}}^{(l)}(m) \le \alpha$}
		\STATE $\widehat m^{(l)}(\alpha) \gets m$.
		\ENDIF
		\ENDFOR
		\ENDFOR
		\FOR{$l \in L$}
		\STATE Remove corresponding units in $\widehat{\mathcal R}^{(l)}(\widehat m^{(l)}(\alpha))$ from $\hat{f}$.
		\ENDFOR
		\STATE $\hat f^{p} \gets \hat f$.
		\RETURN The pruned model $\hat f^{p}$.
	\end{algorithmic}
\end{algorithm}

The pruning procedure for one layer has been described above. The overall
pruning process for all layers is summarized in Algorithm~\ref{algorithm:FAIR}.
For each layer, pruning proceeds under a given ToD level $\alpha$, thereby
inducing globally coordinated control over the compression--performance
trade-off.

In deployment scenarios, the desired compression level is often specified by a
target budget, such as the number of remaining channels, parameters, FLOPs, or
routed experts, rather than by a ToD level itself. We select the ToD level by
sweeping the finite set of empirical conflict values after the U-Scores and
R-Scores have been computed. This budget-matching step does not require
retraining or re-estimating scores, and is described in
Appendix~\ref{app:target_budget}. Different choices of $\alpha$ correspond to
different compression--accuracy trade-offs. For higher compression,
FAIR-Pruner can also be applied iteratively with a fixed $\alpha$, following
the iterative refinement strategy suggested by the Lottery Ticket
Hypothesis~\cite{frankle2018lottery}.

\subsection{Monotonicity and Stepwise Behavior of the ToD Control}
\label{subsec:alpha_monotonicity}

The ToD level $\alpha$ serves as the global control parameter in FAIR-Pruner. To better understand its role, we analyze how the selected pruning depth changes as $\alpha$ varies.

\begin{proposition}
	\label{prop:alpha_monotone}
	For each layer $l \in L$, the function $\widehat m^{(l)}(\alpha)$ is nondecreasing in $\alpha \in (0,1)$. Consequently, the overall pruning ratio induced by FAIR-Pruner is also nondecreasing in $\alpha$.
\end{proposition}

The proof is immediate from the nesting relation
\begin{equation}
	\begin{aligned}
		&\{m:\widehat{\mathrm{ToD}}^{(l)}(m)\le \alpha_1\}
		\subseteq
		\{m:\widehat{\mathrm{ToD}}^{(l)}(m)\le \alpha_2\},
	\end{aligned}
\end{equation}
whenever $\alpha_1\le \alpha_2$, and is deferred to Appendix~\ref{app:proof_prop2}.
Appendix~\ref{app:alpha_empirical_behavior} further reports empirical
diagnostics showing how the achieved pruning rate and one-shot degradation vary
with the ToD level.

\subsection{Theoretical Guarantees}
\label{subsec:theory}

Our theory explains what each component of ToD is responsible for at the
population level. The empirical stability result shows that the ToD allocation
rule is well defined under consistent ranking estimates. The finite-deletion
decomposition then shows why pruning loss has two distinct sources: singleton
deletion sensitivity, measured by the R-Score, and a structural interaction
term caused by deleting multiple units together. The R-mass result shows what
the ToD constraint directly controls. Finally, a structural loss envelope and an additive exchange result characterize when the ToD-induced profile can improve over same-budget uniform allocation.

\paragraph*{Population removal/protection sets.}
Fix a prunable layer $l\in L$. Set
\(\mathcal R^{(l)}(0)=\mathcal P^{(l)}(0)=\varnothing\). For each
\(m\in\{1,\ldots,J^{(l)}\}\), define the population removal set
\begin{equation}
	\mathcal R^{(l)}(m)
	:=
	\left\{
	j\in[J^{(l)}]: U^{(l)}_j \le U^{(l)}_{(m)}
	\right\},
\end{equation}
where
\begin{equation}
	U^{(l)}_{(1)} < U^{(l)}_{(2)} < \cdots < U^{(l)}_{(J^{(l)})}
\end{equation}
are the ordered population U-Scores. Similarly, define the population
protection set, for \(m\in\{1,\ldots,J^{(l)}\}\), by
\begin{equation}
	\mathcal P^{(l)}(m)
	:=
	\left\{
	j\in[J^{(l)}]: R_j^{(l)} \ge R_{(J^{(l)}-m+1)}^{(l)}
	\right\},
\end{equation}
where
\begin{equation}
	R_{(1)}^{(l)} < R_{(2)}^{(l)} < \cdots < R_{(J^{(l)})}^{(l)}
\end{equation}
are the ordered population R-Scores defined
in Subsection~\ref{subsec:r-s}. Define the population ToD by
\begin{equation}
	\begin{aligned}
		\mathrm{ToD}^{(l)}(m)
		&:=
		\frac{
			\left|
			\mathcal R^{(l)}(m)\cap \mathcal P^{(l)}(m)
			\right|
		}{
			\max\{m,1\}
		},
		\,
		m\in\{0,1,\ldots,J^{(l)}\},
	\end{aligned}
\end{equation}
and define the corresponding population pruning depth by
\begin{equation}
	\begin{aligned}
		m^{*(l)}(\alpha)
		&:=
		\max\bigl\{
		m\in\{0,1,\ldots,J^{(l)}\}:\\
		&\hspace{4.2em}
		\mathrm{ToD}^{(l)}(m)\le \alpha
		\bigr\},
		\qquad \alpha\in(0,1).
	\end{aligned}
\end{equation}

\begin{assumption}[U-Score consistency]
	\label{ass:u_score_consistency}
	For each layer $l\in L$,
	\begin{equation}
		\max_{j\in[J^{(l)}]}
		\left|
		\widehat U_j^{(l)} - U_j^{(l)}
		\right|
		\overset{a.s.}{\longrightarrow} 0,
		\qquad \text{as } \min_{k\in[K]}n_k\to\infty.
	\end{equation}
\end{assumption}

Under mild moment conditions, the empirical U-Score consistently estimates
its population counterpart. In particular, Proposition~\ref{prop:uscore_consistency}
in Appendix~\ref{app:uscore_consistency} gives a sufficient condition for
Assumption~\ref{ass:u_score_consistency}.

\begin{assumption}[R-Score consistency]
	\label{ass:protection_consistency}
	For each layer $l\in L$,
	\begin{equation}
		\max_{j\in[J^{(l)}]}
		\left|
		\widehat R_j^{(l)} - R_j^{(l)}
		\right|
		\overset{a.s.}{\longrightarrow} 0,
		\qquad \text{as } n\to\infty.
	\end{equation}
\end{assumption}

\begin{assumption}[Rank separation]
	\label{ass:rank_separation}
	For each layer $l\in L$,
	\begin{equation}
		\min_{i\ne j}|U^{(l)}_i-U^{(l)}_j|>0,
		\qquad
		\min_{i\ne j}|R_i^{(l)}-R_j^{(l)}|>0.
	\end{equation}
\end{assumption}

\begin{theorem}[Selection consistency of empirical ToD]
	\label{thm:selection_consistency}
	Under Assumptions~\ref{ass:u_score_consistency}--\ref{ass:rank_separation},
	for any fixed $\alpha\in(0,1)$ and each fixed layer $l\in L$,
	\begin{equation}
		\Pr\!\left(
		\widehat{\mathcal R}^{(l)}(\widehat m^{(l)}(\alpha))
		\ne
		\mathcal R^{(l)}(m^{*(l)}(\alpha))
		\right)
		\to 0.
	\end{equation}
	If, in addition, $|L|<\infty$, then
	\begin{equation}
		\Pr\!\left(
		\exists\, l\in L:
		\widehat{\mathcal R}^{(l)}(\widehat m^{(l)}(\alpha))
		\ne
		\mathcal R^{(l)}(m^{*(l)}(\alpha))
		\right)
		\to 0.
	\end{equation}
\end{theorem}

Theorem~\ref{thm:selection_consistency} is a stability result for the ToD
allocation rule. Once the empirical removal and protection rankings recover
their population rankings, the empirical conflict curve and selected pruning set
recover their population counterparts. The result is not tied to the default
Wasserstein U-Score or the first-order Taylor R-Score; it only requires
consistent estimation of the two rankings used by ToD. Tie handling and
near-tie interpretations are discussed in Appendix~\ref{app:ties}.

We next turn from the empirical score asymmetry to a loss-level explanation of why the two signals play different roles. The key point is that simultaneous pruning loss is not determined only by the singleton loss change of each removed unit. It also contains structural interaction terms induced by deleting multiple units
together. The following envelope makes this separation explicit.

\paragraph*{Population R-Score and simultaneous pruning loss.}
For any finite pruning set $S\subseteq\mathcal U$, let $\hat f_{-S}$ be the
network obtained by removing all units in $S$, and define
\begin{equation}
	\begin{aligned}
		\Delta\mathcal L(S)
		&:=
		\mathbb E\left[
		\mathcal L(\hat f_{-S}(X),Y)-\mathcal L(\hat f(X),Y)
		\right].
	\end{aligned}
\end{equation}
By the definition of the population R-Score in Subsection~\ref{subsec:r-s}, for
every singleton pruning set \(\{(l,j)\}\),
\begin{equation}
	R_j^{(l)}
	=
	|\Delta\mathcal L(\{(l,j)\})|.
\end{equation}

\begin{theorem}[Finite-deletion pruning loss envelope]
	\label{thm:finite_deletion_envelope}
	Fix a finite pruning set \(S\subseteq\mathcal U\). Suppose that there exist
	nonnegative structural perturbation magnitudes \(\{b_u:u\in S\}\) and a
	constant \(\lambda\ge0\) such that, for all distinct \(u,v\in S\) and all
	\(T\subseteq S\setminus\{u,v\}\),
	\begin{equation}
		\begin{aligned}
			&\big|
			\Delta\mathcal L(T\cup\{u,v\})
			-\Delta\mathcal L(T\cup\{u\})\\
			&\qquad
			-\Delta\mathcal L(T\cup\{v\})
			+\Delta\mathcal L(T)
			\big|
		\end{aligned}
		\le
		\lambda b_ub_v.
	\end{equation}
	Then the non-additive deletion interaction satisfies
	\begin{equation}
		\begin{aligned}
			&\left|
			\Delta\mathcal L(S)
			-
			\sum_{(l,j)\in S}\Delta\mathcal L(\{(l,j)\})
			\right|\\
			\le&
			\lambda\sum_{\{u,v\}\subseteq S}b_ub_v
			\le
			\frac{\lambda}{2}
			\left(\sum_{(l,j)\in S}b_{(l,j)}\right)^2.
		\end{aligned}
	\end{equation}
	Consequently,
	\begin{equation}
		|\Delta\mathcal L(S)|
		\le
		\sum_{(l,j)\in S}R_j^{(l)}
		+
		\frac{\lambda}{2}
		\left(\sum_{(l,j)\in S}b_{(l,j)}\right)^2.
	\end{equation}
\end{theorem}

Theorem~\ref{thm:finite_deletion_envelope} separates two sources of pruning
loss. The first term is the sum of population R-Scores of the removed units,
and therefore corresponds to singleton deletion sensitivity. The second term is
a structural interaction envelope that appears only when multiple units are
deleted jointly. The quantities \(b_u\) and \(\lambda\) are analysis quantities
rather than algorithmic inputs: \(b_u\) measures the structural perturbation
induced by deleting unit \(u\), while \(\lambda\) controls the worst-case
pairwise deletion interaction under arbitrary previously deleted contexts. This
is a finite-difference condition on the actual deletion losses, not a global
smoothness assumption on the network. It is mild in the same local-curvature
sense used by Taylor- and Hessian-based pruning: those methods control the loss
change of deleting weights or units through first- or second-order local
variations
\cite{lecun1989optimal,hassibi1992second,molchanov2017pruning,
	molchanov2019importance,dong2017learning,wang2019eigendamage}, as summarized
in pruning surveys~\cite{cheng2024survey,he2023structured}, whereas
Theorem~\ref{thm:finite_deletion_envelope} only requires the pairwise
second-order finite difference between deletion operations to be bounded by
\(\lambda b_ub_v\). Appendix~\ref{app:finite_deletion_envelope} records a
sufficient route to this finite-difference bound, and
Appendix~\ref{app:interaction_derivative_interpretation} gives a
perturbation-based interpretation of the interaction term.

This decomposition gives the loss-level reason for using two signals. The
R-Score protects units with large singleton deletion loss, but it does not
provide a removal ordering that controls the structural interaction term.
Conversely, a structural removal signal can rank redundant units, but
without R-Score protection it may remove units with large task-level deletion
loss. FAIR-Pruner therefore uses the U-Score to supply a removal-oriented prefix
and the R-Score to define a protected high-sensitivity tail; ToD links the two
by allowing the removal prefix to grow only while its overlap with the protected
tail remains below the tolerated conflict level.

To relate the removal ranking to the interaction term, the qualitative envelope
below uses a structural compatibility condition: for each layer \(l\in L\),
the U-Score prefix satisfies
\begin{equation}
	\sum_{j\in\mathcal R^{(l)}(m)} b_{(l,j)}
	\le
	B_l^U(m),
	\qquad m=0,1,\ldots,J^{(l)},
	\label{eq:structural_compatibility}
\end{equation}
for some nondecreasing layer-wise envelope \(B_l^U(m)\). This condition is not
implied by the Wasserstein definition alone and is not required to run ToD. It
only states when the chosen removal signal is compatible with the structural
interaction term in Theorem~\ref{thm:finite_deletion_envelope}. Sufficient
examples and the relation to norm, FPGM, and rank-based pruning criteria are
discussed in Appendix~\ref{app:structural_proxy_sufficient}.

We now connect the ToD constraint to the first term in
Theorem~\ref{thm:finite_deletion_envelope}. Since ToD limits the overlap between
the U-Score removal prefix and the protected high-\(R\)-Score tail, it directly
controls how much high-sensitivity R-Score mass can enter the selected removal
set.

\begin{lemma}[Feasibility of a ToD-constrained depth]
	\label{lem:tod_feasibility}
	Fix a layer \(l\) with \(J^{(l)}\) prunable units and a depth
	\(m\in\{0,1,\ldots,J^{(l)}\}\). For \(\alpha\in[0,1]\), let
	\begin{equation}
		\begin{aligned}
			\mathcal C_l(m,\alpha)
			:=
			\{S\subseteq[J^{(l)}]:
			& |S|=m,\ 
			\\[-1pt]
			& |S\cap\mathcal P^{(l)}(m)|\le \lfloor\alpha m\rfloor\}.
		\end{aligned}
	\end{equation}
	Then \(\mathcal C_l(m,\alpha)\neq\varnothing\) if and only if
	\begin{equation}
		\max\{0,2m-J^{(l)}\}\le \lfloor\alpha m\rfloor .
	\end{equation}
	Thus, up to integer rounding, any feasible depth must satisfy
	\(m\le J^{(l)}/(2-\alpha)\).
\end{lemma}
The proof is given in Appendix~\ref{app:tod_feasibility}. This lemma concerns
the existence of some size-\(m\) set satisfying the ToD overlap constraint. The
actual FAIR-Pruner depth additionally requires the U-Score prefix itself to
satisfy \(\mathrm{ToD}^{(l)}(m)\le\alpha\).

\begin{proposition}[ToD control of R-Score mass]
	\label{prop:tod_sensitivity_mass}
	Fix a layer \(l\in L\) and \(m\in\{0,1,\ldots,J^{(l)}\}\). Let
	\begin{equation}
		R_{[1]}^{(l)}
		\ge R_{[2]}^{(l)}
		\ge\cdots\ge R_{[J^{(l)}]}^{(l)}
	\end{equation}
	be the population R-Scores in layer \(l\) sorted in
	non-increasing order, and set \(R_{[r]}^{(l)}=0\) for
	\(r>J^{(l)}\). We also use the convention that empty sums are zero. Define
	\begin{equation}
		\mathcal A_l(0,\alpha):=0,
	\end{equation}
	and, for \(m\ge1\),
	\begin{equation}
		\begin{aligned}
			\mathcal A_l(m,\alpha)
			&:=
			\sum_{r=1}^{\lfloor \alpha m\rfloor}R_{[r]}^{(l)}
			+
			\sum_{r=m+1}^{2m-\lfloor \alpha m\rfloor}R_{[r]}^{(l)}.
		\end{aligned}
	\end{equation}
	If \(\mathrm{ToD}^{(l)}(m)\le\alpha\), then
	\begin{equation}
		\sum_{j\in\mathcal R^{(l)}(m)}R_j^{(l)}
		\le \mathcal A_l(m,\alpha).
	\end{equation}
	Moreover, whenever the feasible class \(\mathcal C_l(m,\alpha)\) is nonempty,
	\(\mathcal A_l(m,\alpha)\) is the sharp rank-based worst-case upper bound
	over \(\mathcal C_l(m,\alpha)\). If \(\mathcal C_l(m,\alpha)=\varnothing\),
	the depth-\(m\) ToD-constrained problem is infeasible and no sharpness claim
	is made.
\end{proposition}
The proof, including the tightness of this rank-based bound over the nonempty
feasible class, is given in Appendix~\ref{app:tod_sensitivity_mass}. For the
ToD-selected depth, infeasibility of all positive depths simply gives
\(m^{*(l)}(\alpha)=0\); that layer is left unpruned and contributes zero
R-Score mass and zero deletion-induced structural perturbation. For a fixed ToD
level \(\alpha\), write
\begin{equation}
	S_\alpha
	:=
	\bigcup_{l\in L}\{(l,j):j\in\mathcal R^{(l)}(m^{*(l)}(\alpha))\}. 
\end{equation}
Define
\begin{equation}
	\begin{aligned}
		A_\alpha
		:=
		\sum_{l\in L}
		\mathcal A_l\bigl(m^{*(l)}(\alpha),\alpha\bigr),\,
		B_\alpha^U
		:=
		\sum_{l\in L}
		B_l^U\bigl(m^{*(l)}(\alpha)\bigr).
	\end{aligned}
\end{equation}

\begin{corollary}[Qualitative ToD loss envelope]
	\label{cor:tod_loss_envelope}
	Suppose the ToD-selected set \(S_\alpha\) satisfies the bounded
	finite-deletion interaction condition in
	Theorem~\ref{thm:finite_deletion_envelope}. If the removal ranking satisfies
	the structural compatibility condition~\eqref{eq:structural_compatibility},
	the population ToD-selected pruning set
	satisfies
	\begin{equation}
		|\Delta\mathcal L(S_\alpha)|
		\le
		A_\alpha+\frac{\lambda}{2}\left(B_\alpha^U\right)^2.
	\end{equation}
\end{corollary}

Corollary~\ref{cor:tod_loss_envelope} should be read as a qualitative structural
loss envelope rather than as a practically computable certificate. The constants
\(b_u\), \(B_\alpha^U\), and \(\lambda\) are not estimated by the algorithm and
are not used to certify a numerical test loss after pruning. The value of the
corollary is diagnostic: it shows that the ToD-selected set is governed by two
distinct quantities, namely the R-Score mass admitted by the conflict constraint
and the structural interaction envelope associated with the removal prefixes. A
numerical certificate would require an additional procedure for estimating
\(b_u\), \(B_\alpha^U\), and \(\lambda\).

The sharpness in Proposition~\ref{prop:tod_sensitivity_mass} is conditional on
the nonempty feasible class in Lemma~\ref{lem:tod_feasibility} and concerns the
rank-based R-Score mass term. It is most
informative when the R-Score distribution has a separated high-sensitivity tail;
when R-Scores are nearly flat, no rank-based protection rule can yield a large
worst-case gain because the protection signal itself carries little tail
information.

Corollary~\ref{cor:tod_loss_envelope} gives an upper envelope, but a direct
comparison with uniform pruning requires a same-budget cost model. The
finite-deletion decomposition in Theorem~\ref{thm:finite_deletion_envelope}
shows that pruning loss contains two parts: singleton deletion sensitivity,
measured by the R-Score, and a residual interaction term controlled by
structural perturbation. This decomposition naturally suggests a local additive
comparison regime in which the marginal cost of pruning the \(k\)-th U-ranked
unit of layer \(l\) is written as
\begin{equation}
	c_{l,k}:=R_{l,k}+\gamma b_{l,k},
	\qquad \gamma\ge0.
\end{equation}
Here \(R_{l,k}\) is the R-Score of that unit and \(b_{l,k}\) is its structural
perturbation magnitude. This additive cost is not an unconditional model of
pruning loss and does not imply that ToD always dominates uniform pruning.
Rather, it identifies a favorable operating regime of ToD. In this regime, the
removal ranking proposes structurally plausible candidates, while the R-Score
excludes units with large singleton deletion sensitivity. The next result
formalizes this intuition as a sufficient exchange condition: when the
ToD-induced profile satisfies this condition, it is optimal among same-budget
profiles under the additive surrogate, and therefore has no larger surrogate
cost than same-budget uniform pruning.

\begin{proposition}[Conditional optimality under additive exchange]
	\label{prop:compatible_tod_uniform}
	For each layer \(l\), suppose the marginal costs
	\(c_{l,1},\ldots,c_{l,J^{(l)}}\) are nondecreasing along the U-ranked prefix.
	For a layer-wise pruning profile \(m=(m_l)_{l\in L}\), define
	\begin{equation}
		C(m):=\sum_{l\in L}\sum_{k=1}^{m_l}c_{l,k}.
	\end{equation}
	Let \(M\) be a target pruning budget and let \(m^{\mathrm{tod}}\) be the
	ToD-induced profile with \(\sum_l m_l^{\mathrm{tod}}=M\). Suppose that, for
	all layers \(a,b\) satisfying
	\(m_a^{\mathrm{tod}}>0\) and \(m_b^{\mathrm{tod}}<J^{(b)}\),
	\begin{equation}
		c_{a,m_a^{\mathrm{tod}}}
		\le
		c_{b,m_b^{\mathrm{tod}}+1}.
	\end{equation}
	Then
	\begin{equation}
		m^{\mathrm{tod}}\in
		\arg\min_{\sum_l m_l=M} C(m).
	\end{equation}
	If the displayed inequality is strict for every such pair \((a,b)\), then
	every other same-budget profile has strictly larger cost.
\end{proposition}

This proposition is not a score-only prediction that ToD must outperform
uniform pruning. Rather, it states conditional optimality under the additive
exchange condition: every unit admitted by the ToD profile has no larger
marginal cost than the next unit rejected in any other layer. Since uniform
pruning is one feasible same-budget profile, the strict version gives a strict
improvement over uniform whenever the uniform profile differs from
\(m^{\mathrm{tod}}\).

\paragraph*{Scope of conflict-only allocation.}
Finally, ToD should not be interpreted as an unconditional dominance theorem
over every uniform or single-score pruning strategy. Because the conflict curve
uses within-layer rankings and overlap ratios, it is invariant to layer-wise
rescalings of R-Score magnitudes. Thus, rank-based conflict control needs either
magnitude-calibrated risk information or an additive exchange condition, such as
Proposition~\ref{prop:compatible_tod_uniform}, to support dominance statements.

\begin{proposition}[Limitation of rank-only conflict control]
	\label{prop:no_free_lunch_rank_only}
	Fix any two layer-wise depth profiles \(m\) and \(\bar m\) with the same
	total budget, \(\sum_l m_l=\sum_l\bar m_l\), and suppose \(m\ne\bar m\). For
	\(m=(m_l)_{l\in L}\), write
	\begin{equation}
		S(m):=\bigcup_{l\in L}\{(l,j):j\in\mathcal R^{(l)}(m_l)\}.
	\end{equation}
	Then, while preserving all within-layer U- and R-rankings, there exist
	positive R-Score magnitudes such that the additive pruning loss
	\begin{equation}
		\Delta_{\mathrm{add}}(S):=\sum_{(l,j)\in S}R_j^{(l)}
	\end{equation}
	satisfies
	\begin{equation}
		\Delta_{\mathrm{add}}(S(m))
		>
		\Delta_{\mathrm{add}}(S(\bar m)).
	\end{equation}
	Thus conflict curves alone, viewed as rank-based objects without cross-layer
	magnitude calibration or compatibility assumptions, cannot unconditionally
	dominate uniform pruning.
\end{proposition}

This negative result does not contradict the positive ToD results above. It
shows that rank conflict alone is insufficient for unconditional dominance:
cross-layer magnitude calibration or compatibility assumptions are necessary.
This is why Corollary~\ref{cor:tod_loss_envelope} uses R-Score magnitudes and
why the uniform-comparison statement is framed as conditional optimality under
an additive exchange condition.
The proof is given in Appendix~\ref{app:no_free_lunch_rank_only}.

Detailed proofs and auxiliary discussions are collected in the appendix. They
include tie handling, U-Score consistency, sufficient conditions for
removal-score structural compatibility, a smooth-interpolation sufficient
condition for the finite-deletion interaction envelope, perturbation
interpretations of interaction terms, the exact M\"obius decomposition,
canonical ranking regimes, single-signal and compatible regimes, and the proof
of the rank-only limitation result.

\section{Experiments}
\label{sec:experiments}

\subsection{Experimental Settings}
\label{subsec:exp_settings}

\paragraph{Vision benchmarks and architectures}
We evaluate FAIR-Pruner on CIFAR-10 and CIFAR-100~\cite{krizhevsky2009learning}, SVHN~\cite{netzer2011reading}, and ImageNet-1K~\cite{deng2009imagenet}. 
The standard vision experiments cover VGG-16~\cite{simonyan2014very}, AlexNet~\cite{krizhevsky2012imagenet}, ResNet~\cite{he2016deep}, and DenseNet~\cite{huang2017densely}. 
We further include ConvNeXt~\cite{liu2022convnet} and DeiT-B~\cite{touvron2021training} to test whether ToD remains effective beyond classical backbones.

\paragraph{Mixture-of-Experts (MoE) architecture extension}
To examine whether the same allocation principle transfers beyond standard structured vision pruning, we evaluate routed-expert pruning on Qwen1.5-MoE-A2.7B-Chat~\cite{qwen15moe2024}. 
We deliberately use two complementary tasks: WiC~\cite{pilehvar2019wic}, following a generation-based word-in-context protocol, and MMLU-Pro~\cite{wang2024mmlupro}, a more challenging reasoning-oriented stress test. 
All MoE experiments are zero-shot and \emph{prune-only}: after removing routed experts, we do not fine-tune, distill, merge experts, or update the remaining parameters.

\paragraph{Score instantiations}
The U-Score should be understood as a removal-oriented within-layer ranking signal, while ToD is the mechanism that converts removal and protection signals into non-uniform layer-wise pruning depths. 
In the default vision setting, we instantiate the U-Score by the class-conditional separability measure introduced in Section~\ref{subsec:u-s}. 
For ImageNet-scale backbones such as ConvNeXt and DeiT-B, exact pairwise computation can be expensive, so we use scalable approximations described in Appendix~\ref{app:experimental_details}. 
For MoE LLMs, we instantiate the removal-oriented U-Score by soft activation~\cite{muzio2024seer}, which is more natural for routed experts than the class-conditional vision score. 
Thus, the experiments should not be read as claiming that a single Wasserstein-style score is universally optimal. 
Rather, they evaluate whether ToD can serve as a modular coordination mechanism between a removal-oriented signal and a protection-oriented signal.

\paragraph{Training and evaluation protocols}
For vision models, the unpruned baselines follow the original or commonly adopted training recipes of the corresponding architectures~\cite{simonyan2014very,krizhevsky2012imagenet,huang2017densely,he2016deep}. 
After pruning, we match the post-pruning fine-tuning schedule of the corresponding baselines whenever it is available, and we do not use auxiliary enhancement techniques unless explicitly stated in a comparison table.\footnote{All experiments are implemented in PyTorch and conducted on NVIDIA A100 and RTX 6000 Ada GPUs.} 
For MoE LLMs, the reported numbers are obtained immediately after expert removal under matched total routed-expert budgets. 
For reproducibility, we release FAIR-Pruner as an open-source, pip-installable Python package, and the source code is available on the project homepage.\footnote{\url{https://github.com/Chenqing-Lin/FAIR-Pruner}}

\subsection{Main Results on Vision Benchmarks}
\label{subsec:vision_main_results}

\begin{table*}[t]
\centering
\caption{Comparison with structured pruning methods on classical CNN benchmarks. 
MFLOPs are reported for computational comparison. 
$\Delta$ Top-1 denotes the accuracy change (in percentage points) relative to the unpruned baseline within each benchmark. 
FAIR-Pruner rows are highlighted by the method name for readability. 
Competitor numbers are taken from the cited papers unless otherwise stated.
Methods marked with $^*$ use additional techniques such as knowledge distillation, expanded search spaces, or extra training tricks.}
\label{tab:sota_all}
\setlength{\tabcolsep}{4pt}
\small
\begin{tabular}{llcccc}
\hline
Benchmark & Method & Top-1 (\%) & $\Delta$ Top-1 (pp) & MFLOPs & $\Delta$ FLOPs $\downarrow$ \\
\hline

\multirow{12}{*}{CIFAR-10 / ResNet-56}
& Baseline (Unpruned) & 93.93 & --    & 125.0 & -- \\
& Li et al.~\cite{li2017pruning}              & 93.06 & -0.87 & 90.9 & 27.3\% \\
& GAL~\cite{lin2019towards}                   & 92.98 & -0.95 & 78.3 & 37.4\% \\
& Random Pruning~\cite{li2022revisiting}      & 93.48 & -0.45 & 63.8 & 49.0\% \\
& TAS~\cite{dong2019network}                  & 92.87 & -1.06 & 63.1 & 49.5\% \\
& AMC~\cite{he2018amc}                        & 91.90 & -2.03 & 62.9 & 49.7\% \\
& DECODER-200~\cite{alwani2022decore}         & 93.26 & -0.67 & 62.6 & 50.0\% \\
& LASSO~\cite{He_2017_ICCV}                   & 91.80 & -2.13 & 62.0 & 50.4\% \\
& APS~\cite{wang2020revisiting}               & 93.42 & -0.51 & 60.3 & 51.8\% \\
& ITPruner~\cite{zheng2025information}        & 93.43 & -0.50 & 59.5 & 52.4\% \\
& MFP~\cite{he2022filter}                     & 93.56 & -0.37 & 59.3 & 52.6\% \\
& \textbf{FAIR-Pruner (Ours)}                  & \textbf{93.64} & \textbf{-0.29} & \textbf{57.8} & \textbf{53.8\%} \\
\hline

\multirow{8}{*}{CIFAR-10 / DenseNet-40}
& Baseline (Unpruned) & 94.81 & --    & 282.0 & -- \\
& GAL-0.01~\cite{lin2019towards}              & 94.29 & -0.52 & 182.9 & 35.1\% \\
& HRank~\cite{lin2020hrank}                   & 94.24 & -0.57 & 167.4 & 40.6\% \\
& Zhao et al.~\cite{zhao2019variational}      & 93.16 & -1.65 & 156.0 & 44.7\% \\
& GAL-0.05~\cite{lin2019towards}              & 93.53 & -1.28 & 128.1 & 54.6\% \\
& HRank~\cite{lin2020hrank}                   & 93.68 & -1.13 & 110.2 & 60.9\% \\
& \textbf{FAIR-Pruner (Ours)}                  & \textbf{94.38} & \textbf{-0.43} & \textbf{148.0} & 47.5\% \\
& \textbf{FAIR-Pruner (Ours)}                  & 93.65 & -1.16 & 80.2 & \textbf{71.6\%} \\
\hline

\multirow{12}{*}{ImageNet-1K / ResNet-50}
& Baseline (Unpruned) & 76.88 & --    & 4089 & -- \\
& GAL~\cite{lin2019towards}                   & 71.95 & -4.93 & 2330 & 43.0\% \\
& AutoSlim$^*$~\cite{yu2019autoslim}          & 74.90 & -1.98 & 2300 & 43.8\% \\
& DECODER-6~\cite{alwani2022decore}           & 74.58 & -2.30 & 2372 & 42.0\% \\
& MetaPruning$^*$~\cite{liu2019metapruning}   & 72.17 & -4.71 & 2260 & 44.7\% \\
& Taylor~\cite{molchanov2019importance}       & 74.50 & -2.38 & 2250 & 45.0\% \\
& Random Pruning~\cite{li2022revisiting}      & 75.13 & -1.75 & 2074 & 49.3\% \\
& HRank~\cite{lin2020hrank}                   & 74.98 & -1.90 & 2300 & 43.8\% \\
& ITPruner~\cite{zheng2025information}        & 75.28 & -1.60 & 1943 & 52.5\% \\
& MFP~\cite{he2022filter}                     & 74.86 & -2.02 & 1942 & 52.5\% \\
& \textbf{FAIR-Pruner (Ours)}                  & \textbf{75.29} & \textbf{-1.59} & \textbf{1932} & \textbf{52.8\%} \\
\hline


\end{tabular}
\end{table*}

Table~\ref{tab:sota_all} compares FAIR-Pruner with recent structured pruning methods on classical CNN benchmarks, including CIFAR-10 with ResNet-56 and DenseNet-40, and ImageNet-1K with ResNet-50. 
Across different compression regimes, FAIR-Pruner achieves a favorable accuracy--efficiency trade-off while remaining search-free. 
For SOTA comparisons, we sweep the ToD level after the scores are computed and report operating points whose FLOPs are close to the corresponding comparison block, so the comparison is made under matched or comparable computational budgets.
On CIFAR-10 with ResNet-56, it reaches 93.64\% Top-1 accuracy with 57.8M FLOPs, giving the largest FLOPs reduction in that benchmark block while maintaining competitive accuracy. 
On CIFAR-10 with DenseNet-40, FAIR-Pruner remains robust under both moderate and aggressive compression, including the high-compression setting with 80.2M FLOPs and 93.65\% Top-1 accuracy. 
On full ImageNet-1K with ResNet-50, FAIR-Pruner obtains 75.29\% Top-1 accuracy at 1932M FLOPs, matching or slightly exceeding strong baselines such as HRank and ITPruner~\cite{lin2020hrank,zheng2025information} under comparable computational budgets.

\begin{table*}[t]
\centering
\caption{Comparison on modern block-structured architectures. 
FAIR-Pruner prunes MLP hidden dimensions in ConvNeXt and attention heads/MLP hidden dimensions in DeiT-B. 
FLOPs are reported in GFLOPs and FAIR-Pruner rows are highlighted.}
\label{tab:modern_architectures}
\setlength{\tabcolsep}{3.5pt}
\small
\begin{tabular}{llcccc}
\hline
Benchmark / Model & Method 
& Top-1 (\%) & $\Delta$ Top-1 (pp) & GFLOPs & $\Delta$ FLOPs $\downarrow$ \\
\hline

\multirow{6}{*}{CIFAR-100 / ConvNeXt}
& Baseline (Unpruned) & 90.01 & --    & 8.683 & -- \\
& MetaPruning~\cite{liu2019metapruning}       & 82.61 & -7.40  & 3.275 & 62.3\% \\
& Li et al.~\cite{li2017pruning}              & 68.07 & -21.94 & 3.499 & 59.7\% \\
& Taylor~\cite{molchanov2019importance}       & 83.36 & -6.65  & 2.328 & 73.2\% \\
& FPGM~\cite{he2019filter}                    & 80.76 & -9.25  & 2.358 & 72.8\% \\
& \textbf{FAIR-Pruner (Ours)}                 & 82.96 & -7.05  & 2.391 & 72.5\% \\
\hline

\multirow{5}{*}{ImageNet-1K / DeiT-B}
& Baseline~\cite{touvron2021training}           & 81.80 & --     & 16.8 & -- \\
& WDPruning~\cite{yu2022width}                 & 80.76 & -1.04  & 9.9  & 41.1\% \\
& IA-RED~\cite{pan2021ia}        & 80.30 & -1.50  & 11.8 & 29.8\% \\
& ITPruner~\cite{zheng2025information}         & 81.20 & -0.60  & 9.9  & 41.1\% \\
& \textbf{FAIR-Pruner (Ours)}                  & 81.20 & -0.60  & 9.9  & 41.1\% \\
\hline
\end{tabular}
\end{table*}

Table~\ref{tab:modern_architectures} further evaluates FAIR-Pruner on modern block-structured architectures. 
These experiments differ from the classical CNN setting because the pruned units are no longer limited to standard convolutional channels. 
For ConvNeXt, FAIR-Pruner prunes the hidden dimensions of the feed-forward component inside modern convolutional blocks. 
For DeiT-B, FAIR-Pruner prunes attention heads together with MLP hidden dimensions. 
On CIFAR-100/ConvNeXt, FAIR-Pruner obtains 82.96\% Top-1 accuracy at 2.391 GFLOPs, remaining close to Taylor pruning and exceeding magnitude-based pruning under similar computation. 
On ImageNet-1K/DeiT-B, FAIR-Pruner reaches 81.20\% Top-1 accuracy under a 41.1\% FLOPs reduction, matching ITPruner. 
These results suggest that the ToD-based allocation mechanism is not restricted to classical convolutional channel pruning and can be adapted to block-level pruning in modern architectures. 
Implementation details for the scalable U-Score approximations used in these larger settings are provided in Appendix~\ref{app:experimental_details}.

\subsection{Architecture Extension to Routed-Expert MoE}
\label{subsec:moe_pruning}
\begin{table}[t]
\centering
\caption{WiC generation-based accuracy under routed-expert pruning. Kept experts are reported as per-layer counts with total routed experts in parentheses; for example, 30 (720) means keeping 30 of 60 routed experts per layer, i.e., 720 of 1440 experts in total. 
For FAIR, ToD induces non-uniform layer-wise budgets while matching the same total budget. 
Random is averaged over 10 seeds.}
\label{tab:wic_table9_style}
\begin{tabular}{lccc}
\toprule
Method & Kept & Pruning rate(\%) & Acc. (\%) \\
\midrule
Full model & 60 (1440) & 0  & 57.68 \\
\midrule
Random & 30 (720) & 50.0 & $42.10 \pm 17.88$ \\
Frequency & 30 (720) & 50.0  & 49.69 \\
Soft Act. & 30 (720) & 50.0  & 51.88 \\
\textbf{FAIR} & Non-unif. (720) & 50.0  & $\mathbf{56.74}$ \\
\midrule
Random & 15 (360) & 75.0  & $8.32 \pm 9.14$ \\
Frequency & 15 (360) & 75.0  & 29.15 \\
Soft Act. & 15 (360) & 75.0  & 36.05 \\
\textbf{FAIR} & Non-unif. (360) & 75.0  & $\mathbf{48.28}$ \\
\midrule
Random & 10 (240) & 83.3  & $3.35 \pm 2.54$ \\
Frequency & 10 (240) & 83.3  & 21.79 \\
Soft Act. & 10 (240) & 83.3  & 22.88 \\
\textbf{FAIR} & Non-unif. (240) & 83.3  & $\mathbf{36.83}$ \\
\bottomrule
\end{tabular}
\vspace{2pt}
\footnotesize
\end{table}

\begin{table}[t]
\centering
\caption{Overall MMLU-Pro performance under prune-only 50\% routed-expert pruning. 
Kept experts are reported as per-layer counts with total routed experts in parentheses. 
For uniform baselines, 30 (720) means keeping 30 of 60 routed experts per layer; FAIR matches the same total of 720 kept experts with a non-uniform layer-wise budget. 
Random is averaged over 5 seeds.}
\label{tab:mmlu_pro_main_keep30_strict}
\setlength{\tabcolsep}{3.5pt}
\begin{tabular}{lcccc}
\toprule
Method & Kept & Pruning rate (\%) & Acc. (\%) & Parse (\%) \\
\midrule
Full model & 60 (1440) & 0  & 18.06 & 99.68 \\
\midrule
Random & 30 (720) & 50.0  & $8.81 \pm 2$ & $73.37 \pm 20$ \\
Frequency & 30 (720) & 50.0  & 0.07 & 0.49 \\
Soft Act. & 30 (720) & 50.0  & 10.66 & 90.25 \\
\textbf{FAIR} & Non-unif. (720) & 50.0  & $\mathbf{11.53}$ & $\mathbf{92.85}$ \\
\bottomrule
\end{tabular}
\end{table}

We next test FAIR-Pruner in a routed-expert MoE architecture. 
The goal is not to claim a complete LLM compression recipe, but to examine whether the same removal--protection--allocation framework can be instantiated for expert pruning. 
This setting differs from the vision experiments in two important ways. 
First, all results are prune-only, so the numbers reflect the immediate effect of expert removal without recovery training. 
Second, the removal-oriented U-Score is instantiated by soft activation rather than the default vision U-Score, because soft activation is more directly aligned with expert routing behavior in language models.

The baselines are designed to isolate the effect of non-uniform allocation. 
Frequency and Soft Act. rank experts within each layer using routing-frequency or soft-activation information, respectively, and then keep the same number of experts in every layer. 
For FAIR-Pruner, we use soft activation as the removal-oriented expert signal and use a routing-weighted R-Score as the protection signal. 
Specifically, for an MoE layer \(l\) and routed expert \(e\), let
\(\theta_{l,e}\) be the parameter block of that expert and let
\(g_{i,t,e}^{(l)}\) be the router weight assigned to expert \(e\) at token
\(t\) of calibration sample \(i\). We use the routing-weighted Taylor
protection score
\begin{equation}
	\widehat R_{l,e}
	=
	\left|
	\frac{1}{|\mathcal T|}
	\sum_{(i,t)\in\mathcal T}
	g_{i,t,e}^{(l)}
	\left\langle
	\nabla_{\theta_{l,e}}\ell_{i,t},
	\theta_{l,e}
	\right\rangle
	\right|,
\end{equation}
where \(\mathcal T\) denotes the evaluated calibration-token set and
\(\ell_{i,t}\) is the next-token loss. This weighting distinguishes experts
that actively participate in routing from experts whose parameter sensitivity
is large but rarely used.
ToD then coordinates the soft-activation removal signal and the routing-weighted protection signal to obtain non-uniform layer-wise expert budgets under the same total number of kept routed experts. 
The comparison between Soft Act. and FAIR-Pruner measures whether ToD-based non-uniform allocation improves over a uniform soft-activation pruning rule under a matched removal signal.

Table~\ref{tab:wic_table9_style} shows that, under the matched prune-only protocol, ToD-guided expert allocation gives a clear advantage over uniform expert budgets on WiC. 
At 50\% routed-expert pruning, FAIR-Pruner keeps 720 routed experts and obtains 56.74\% WiC accuracy, close to the 57.68\% full-model accuracy and higher than uniform soft-activation pruning. 
The advantage becomes more pronounced under more aggressive budgets: at 75.0\% and 83.3\% pruning, FAIR-Pruner obtains 48.28\% and 36.83\% accuracy, respectively, while uniform soft-activation pruning drops to 36.05\% and 22.88\%. 
These results indicate that, under a matched expert budget and matched removal signal, ToD gives a more robust expert allocation than uniform expert pruning on WiC.

Table~\ref{tab:mmlu_pro_main_keep30_strict} reports the prune-only 50\% expert-pruning result on MMLU-Pro, where the uniform baselines keep 30 of 60 routed experts per MoE layer and FAIR-Pruner keeps the same total number of experts with a non-uniform layer-wise budget. 
MMLU-Pro is more demanding than WiC: it stresses broad knowledge
retrieval and reasoning under a larger answer space, and our evaluation is a
zero-shot generation-based prune-only comparison rather than a leaderboard
tuning setting. All model variants use the same prompt, decoding, parsing, and
evaluation protocol. Under this matched protocol, FAIR-Pruner improves over
uniform soft-activation pruning at the same 50\% routed-expert pruning budget,
increasing accuracy from 10.66\% to 11.53\% and improving parse rate from
90.25\% to 92.85\%.
The gain on MMLU-Pro is small and should be interpreted cautiously, especially
because the absolute zero-shot accuracy remains far below the full model. This
is expected because
reasoning-oriented tasks are more sensitive to aggressive expert removal than
the word-in-context decision in WiC. We therefore view MMLU-Pro as a stress
test of architectural extensibility: ToD gives a modest improvement over the
uniform soft-activation baseline under a difficult prune-only protocol, while
WiC shows that ToD can preserve most of the full-model accuracy when the base
task accuracy is higher before pruning.

\subsection{Layer-wise Allocation Induced by ToD}
\label{subsec:layerwise_allocation}

\begin{figure*}[t]
	\centering
	\subfloat[Representative CNN pruning profile.]{
		\includegraphics[width=0.48\linewidth]{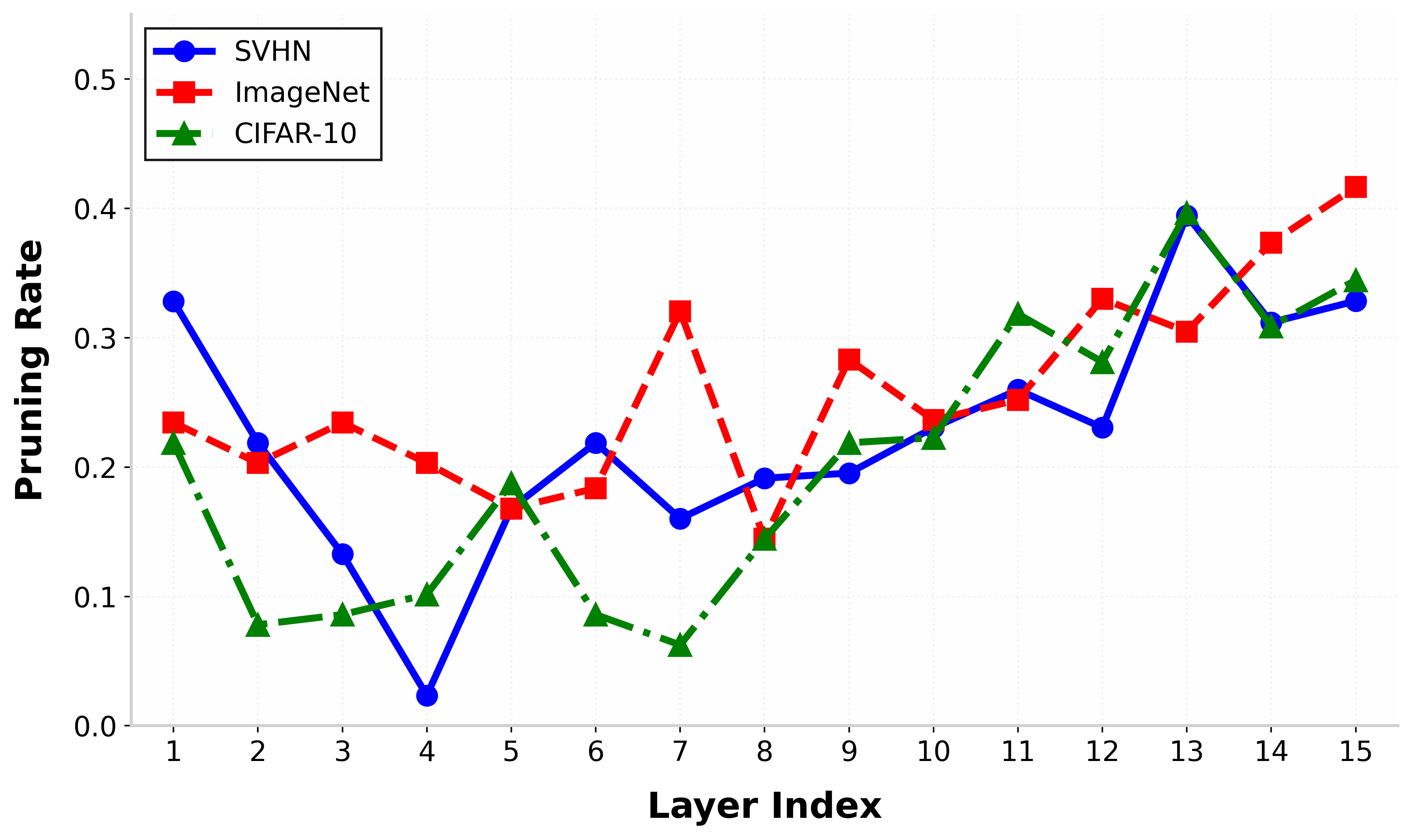}
		\label{fig:layerwise_cnn_main}
	}
	\hfill
	\subfloat[Representative MoE expert-pruning profile.]{
		\includegraphics[width=0.48\linewidth]{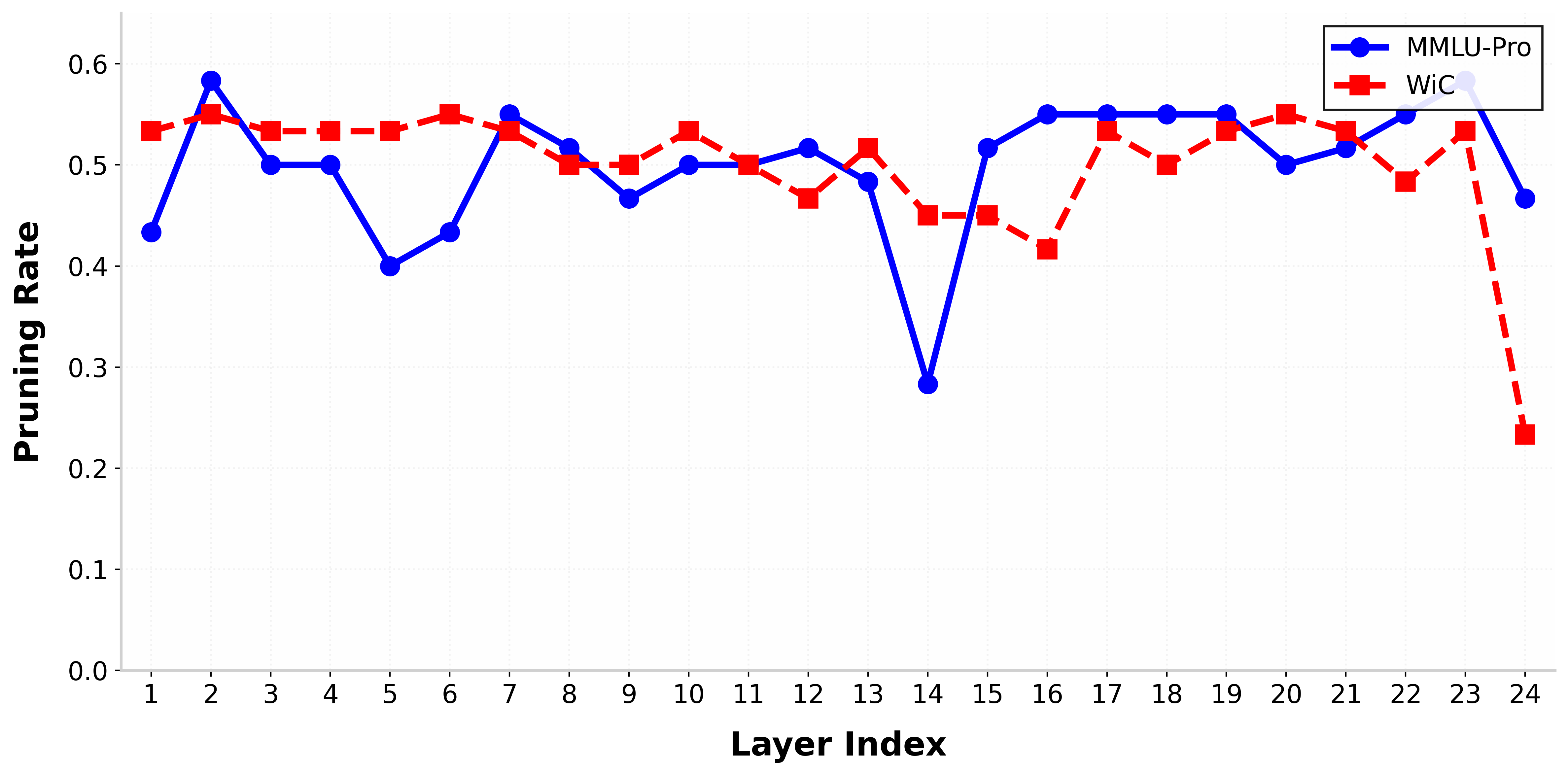}
		\label{fig:layerwise_moe_main}
	}
	\caption{Representative layer-wise pruning profiles induced by ToD. 
		In the CNN case, units correspond to structured channels or hidden widths; in the Qwen1.5-MoE case, units correspond to routed experts. 
		The profiles show that ToD induces non-uniform and architecture-dependent pruning budgets rather than imposing a fixed uniform sparsity pattern. 
		Additional profiles for AlexNet and DeiT-B are provided in Appendix~\ref{app:additional_layerwise_profiles}.}
	\label{fig:layer-wise_pr}
\end{figure*}

Figure~\ref{fig:layer-wise_pr} visualizes two representative layer-wise allocation profiles induced by ToD. 
The CNN profile shows that pruning is not distributed uniformly across layers: some layers are kept relatively conservative, while others admit more aggressive compression. 
This behavior is consistent with the fact that different layers carry different levels of representational redundancy and task sensitivity. 
The MoE profile shows a similar phenomenon at the routed-expert level: under the same total expert budget, ToD assigns different numbers of removable experts to different transformer layers instead of keeping a fixed number of experts in every layer.

These two examples serve different purposes. 
The CNN result illustrates ToD in the standard structured-pruning regime, while the MoE result shows that the same allocation principle can be transferred to routed-expert pruning. 
Together, they support the interpretation that ToD is an adaptive allocation mechanism rather than a handcrafted pruning schedule. 
More layer-wise profiles, including AlexNet and DeiT-B, are reported in Appendix~\ref{app:additional_layerwise_profiles}.

\subsection{Ablation Study and Modularity of ToD}
\label{subsec:ablation_modularity}
\begin{figure*}[t]
	\centering
	\includegraphics[width=0.95\linewidth]{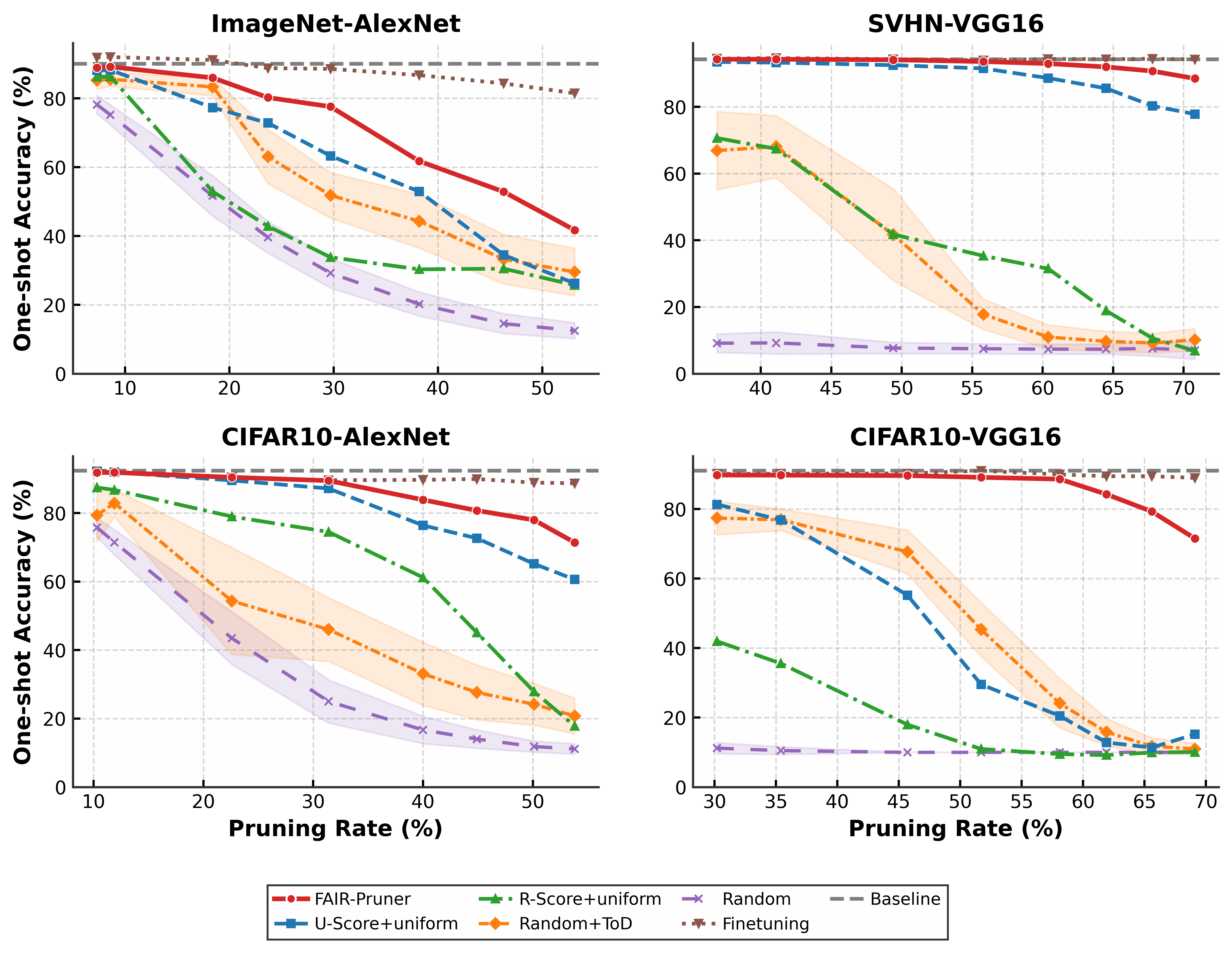}
	\caption{Ablation study on the contributions of the removal signal, protection signal, and ToD-based non-uniform layer-wise allocation to one-shot accuracy. The ToD-guided curves use the fixed ToD levels \(\alpha\in\{0.005,0.01,0.05,0.10,0.15,0.20,0.25,0.30\}\). ``U-Score+uniform'' and ``R-Score+uniform'' use the corresponding score for within-layer unit ranking under a uniform pruning ratio. ``Random'' denotes uniform random pruning. ``Random+ToD'' uses ToD to determine layer-wise pruning ratios while selecting units randomly within each layer. ``FAIR-Pruner'' combines U-Score-based within-layer ranking with ToD-guided non-uniform allocation; its fine-tuned curve reports the same pruned models after 10 epochs of fine-tuning. The shaded region shows $\pm 1$ standard deviation over 20 independent trials for the random baselines.}
	\label{fig:Ablation}
\end{figure*}

Figure~\ref{fig:Ablation} shows that the effectiveness of FAIR-Pruner does not arise only from the choice of within-layer importance metric, but also from the ToD-guided allocation of layer-wise pruning rates. The plotted ToD-guided operating points are obtained by sweeping the fixed levels \(\alpha\in\{0.005,0.01,0.05,0.10,0.15,0.20,0.25,0.30\}\). This grid is not intended to exhaust all possible ToD levels; it covers the range from conservative decisions to high-compression operating points in the tested models. We stop at \(\alpha=0.30\) because, in these sweeps, this already reaches the high-compression region; larger tolerances mainly test over-aggressive pruning with large one-shot degradation, which is less diagnostic for comparing the individual components of the pruning rule. We first compare the orange and purple curves, which isolate the contribution of layer-wise sparsity allocation. Specifically, the orange curve (Random+ToD) corresponds to the setting where ToD is used to determine the pruning ratio of each layer, after which the units to be removed are selected uniformly at random within each layer. In contrast, the purple curve corresponds to standard random pruning with a uniform per-layer pruning rate, where all layers are pruned by the same proportion and the removed units are again selected uniformly at random within each layer. The substantial gap between these two curves shows that even without using any within-layer importance ranking, the pruning rates induced by ToD already provide strong structural guidance. For example, at 35.4\% compression on CIFAR-10 with VGG-16, Random+ToD preserves 76.91\% one-shot accuracy, whereas uniform random pruning drops to 10.5\%, which is essentially chance-level.

We next compare the red and green curves to show that the benefit of ToD remains clear even when the same importance metric is used within each layer. Both FAIR-Pruner (red) and U-Score+uniform (green) adopt the same U-Score to rank units, and the only difference is whether the layer-wise pruning rates are determined by ToD or fixed uniformly across layers. Across all benchmarks in Figure~\ref{fig:Ablation}, FAIR-Pruner consistently preserves substantially higher accuracy, and the gap becomes more pronounced as sparsity increases. For instance, in the SVHN--VGG16 experiment, at a 67.8\% pruning ratio, FAIR-Pruner still achieves 90.71\% one-shot accuracy, whereas U-Score+uniform drops to 80.27\%, yielding a 10.4\% absolute gap. This suggests that the improvement is not merely due to a better within-layer saliency criterion, but is critically driven by the ToD-based pruning-rate allocation across layers.

Overall, these results suggest that ToD helps reduce over-pruning in more sensitive layers while allowing more aggressive pruning in more redundant ones. As a result, it substantially improves robustness under high compression and makes the pruning process more stable even when the same saliency metric is used.

The ablation also clarifies the roles of the two scores. 
The U-Score is more suitable for ranking removable units, while the R-Score is more suitable for defining the protected set of task-sensitive units. 
This is consistent with the score-distribution motivation in Section~\ref{sec:proposed-method}: the R-Score isolates a relatively small subset of highly influential units, whereas the U-Score gives a more discriminative ranking among less important units. 
FAIR-Pruner combines these complementary signals through ToD, which explains why it consistently outperforms Random+ToD and the uniform score-based variants.

\begin{table*}[t]
\centering
\caption{Extensibility of the ToD framework with alternative importance metrics. The external metric is used as a surrogate U-Score to generate the removal set, while the R-Score remains unchanged to define the protection set.}
\label{tab:extensibility_tod}
\begin{tabular}{llccccc}
\hline
Metric & Dataset / Model & Budget & \multicolumn{2}{c}{One-shot / Top-1 (\%)} & \multicolumn{2}{c}{Fine-tuning (\%)} \\
\cline{4-7}
 &  &  & ToD & Uniform & ToD & Uniform \\
\hline
\multicolumn{7}{c}{\textbf{L1 norm as surrogate U-Score}} \\
\hline
L1 norm & CIFAR-10 / VGG-16 & PR = 36.9\% & \textbf{81.14} & 72.08 & \textbf{90.44} & 89.48 \\
L1 norm & CIFAR-10 / VGG-16 & PR = 49.4\% & \textbf{62.24} & 25.77 & \textbf{90.07} & 88.91 \\
L1 norm & CIFAR-10 / VGG-16 & PR = 70.8\% & \textbf{13.85} & 10.02 & \textbf{88.57} & 86.30 \\
L1 norm & SVHN / VGG-16     & PR = 30.2\% & \textbf{89.64} & 76.93 & \textbf{94.23} & 93.82 \\
L1 norm & SVHN / VGG-16     & PR = 45.7\% & \textbf{84.91} & 42.34 & \textbf{94.11} & 93.84 \\
L1 norm & SVHN / VGG-16     & PR = 69.1\% & \textbf{18.32} & 10.49 & \textbf{94.00} & 93.51 \\
\hline
\multicolumn{7}{c}{\textbf{FPGM as surrogate U-Score}} \\
\hline
FPGM & ImageNet-1K / ResNet-50 & 1932M FLOPs & -- & -- & \textbf{75.51} & 74.83  \\
FPGM & ImageNet-1K / ResNet-50 & 2227M FLOPs & -- & -- & \textbf{75.76} & 75.59  \\
\hline
\end{tabular}
\end{table*}

Finally, the benefit of ToD is not tied to the default Wasserstein-based U-Score. 
Table~\ref{tab:extensibility_tod} shows that when the removal signal is replaced by L1 norm or FPGM~\cite{he2019filter}, ToD still improves over the corresponding uniform allocation under matched pruning budgets. 
For instance, on CIFAR-10/VGG-16 at PR = 49.4\%, L1+ToD reaches 90.07\% fine-tuned accuracy compared with 88.91\% for L1+Uniform; on ImageNet-1K/ResNet-50 at 1932M FLOPs, FPGM+ToD reaches 75.51\% compared with 74.83\% for FPGM+Uniform. 
Thus, ToD improves allocation even when the removal ranking is supplied by an
external pruning metric. Together with the Random+ToD and U-Score+uniform
comparisons, these results show that ToD is a modular allocation framework
rather than a by-product of the Wasserstein U-Score.

\subsection{Practical Efficiency and Hardware Footprint}
\label{subsec:efficiency_hardware}
\begin{table}[t]\caption{Inference latency of ResNet\textendash50 on a dual Intel Xeon Gold 6248R CPU server (96 threads, 3.0 GHz).}
\centering
\begin{tabular}{c c c c}
\toprule
BatchSize & Baseline(Unpruned 26M) & Our method(15M) & Speedup \\
\midrule
1 & 40.7ms & \textbf{30.4ms} & \textbf{1.34$\times$} \\
4 & 70.1ms & \textbf{49.8ms} & \textbf{1.41$\times$} \\
8 & 118.9ms & \textbf{86.7ms} & \textbf{1.37$\times$} \\
\bottomrule
\end{tabular}

\label{tab:resnet_cpu_latency_speedup}
\end{table}

FLOPs reduction does not always translate directly into deployment speedup. 
Therefore, Table~\ref{tab:resnet_cpu_latency_speedup} reports CPU inference latency on ResNet-50. 
Across all tested batch sizes, FAIR-Pruner obtains a 1.34$\times$--1.41$\times$ CPU speedup, corresponding to approximately 25\%--29\% lower wall-clock latency. 
This shows that the structured reductions induced by ToD can translate into practical acceleration in a representative CNN setting.

\begin{figure*}[t]
	\centering
	\subfloat[Peak GPU memory and WiC accuracy.]{\includegraphics[width=0.47\linewidth]{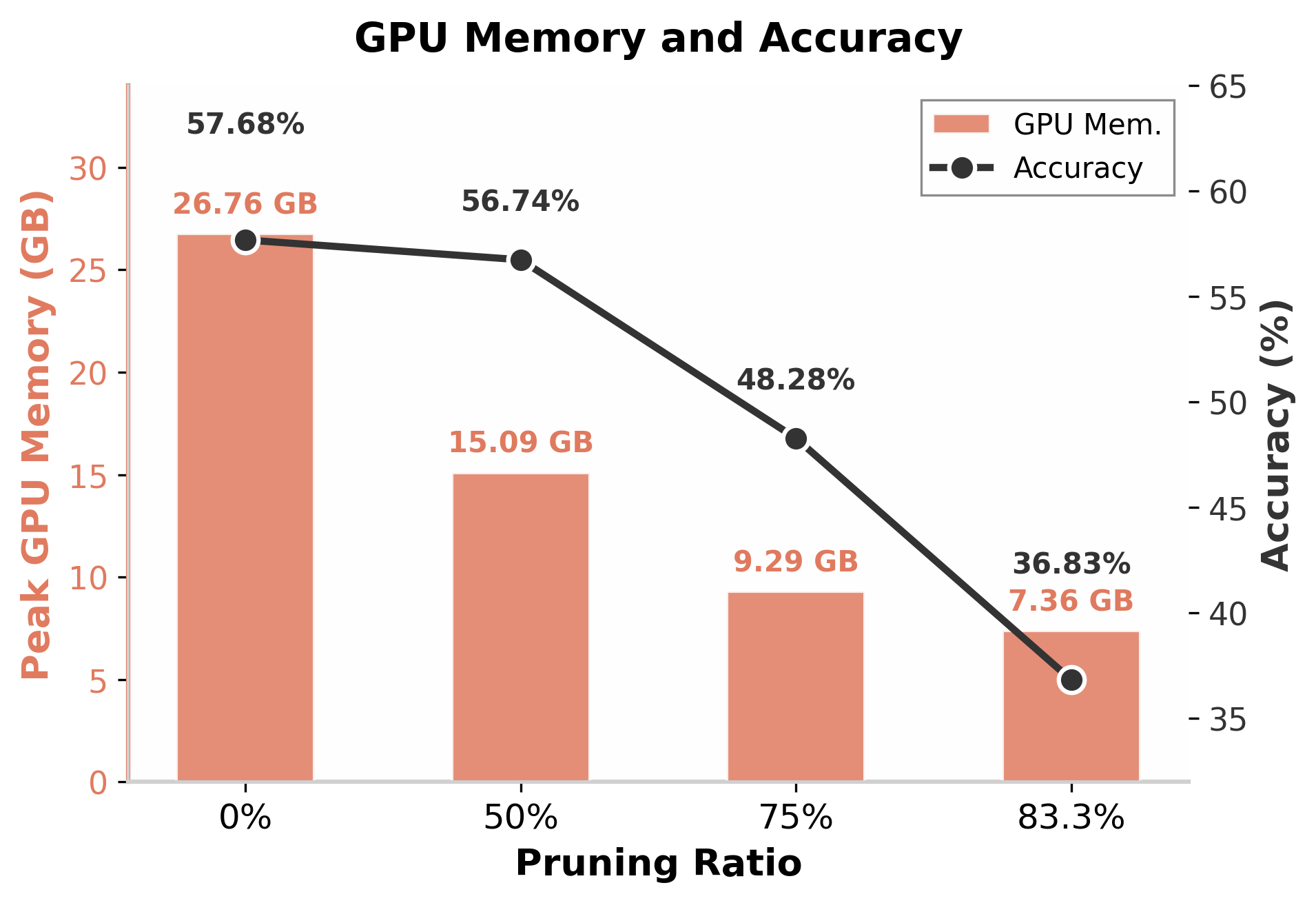}\label{fig:gpu_footprint}}
	\hfill
	\subfloat[Peak CPU RSS and WiC accuracy.]{\includegraphics[width=0.47\linewidth]{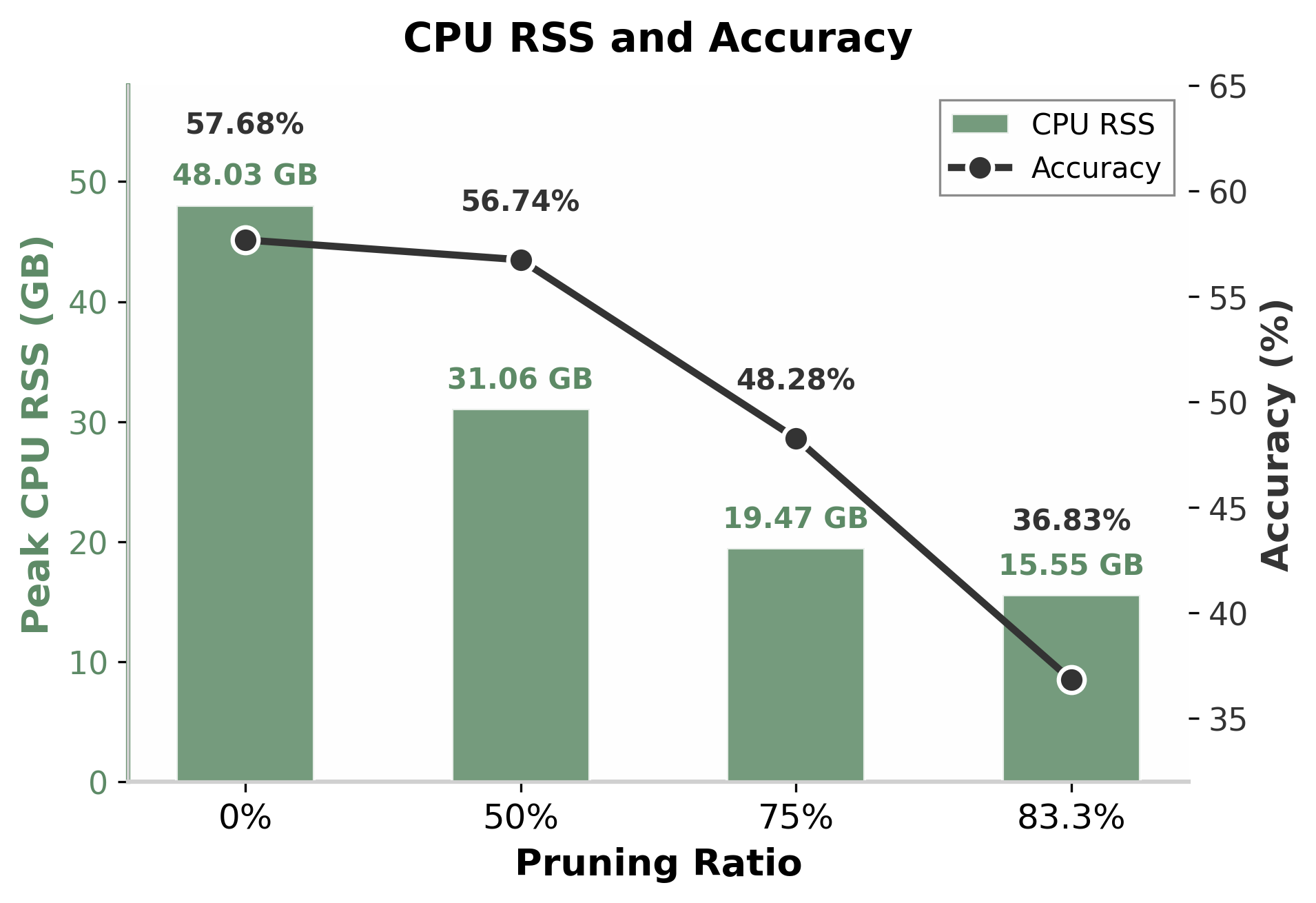}\label{fig:cpu_footprint}}
	\caption{Hardware footprint of prune-only MoE expert pruning on Qwen1.5-MoE. Bars report peak GPU memory or peak CPU resident set size during evaluation, while the line reports WiC accuracy. FAIR-Pruner substantially reduces memory usage as the routed-expert pruning ratio increases; at 50\% routed-expert pruning, it reduces peak GPU memory from 26.76GB to 15.09GB and peak CPU RSS from 48.03GB to 31.06GB, while preserving most of the full-model WiC accuracy.}
	\label{fig:gpu_cpu_footprint}
\end{figure*}

Figure~\ref{fig:gpu_cpu_footprint} complements the latency result by measuring memory footprint in the MoE setting. 
At the 50\% routed-expert pruning budget, FAIR-Pruner preserves most of the full-model WiC accuracy while reducing peak GPU memory and CPU resident memory substantially. 
At more aggressive expert-pruning budgets, memory usage continues to decrease, while accuracy degrades gradually. 
In this routed-expert generation protocol, analytical FLOPs change little because
the same number of tokens and routing operations are evaluated; the main benefit
comes from a smaller expert-parameter footprint and reduced runtime memory.
These results provide practical evidence that the structural reductions produced by FAIR-Pruner can lower both inference latency and memory footprint in representative CNN and MoE settings. 
Additional measurements of pruning-time and sample-size sensitivity are provided in Appendix~\ref{app:sample_efficiency}.

\begin{table*}[t]
	\centering
	\caption{Green inference comparison on WiC under the same wrapped MoE implementation. 
		CodeCarbon tracking is started after model loading and warm-up, so the reported CO$_2$eq corresponds only to the measured inference loop. 
		Latency is reported in ms/example. CO$_2$eq is estimated by CodeCarbon in offline mode with the same country setting. 
		Both the full-size and pruned models are evaluated through the same wrapped MoE implementation.}
	\label{tab:green_inference_wic}
	\setlength{\tabcolsep}{4pt}
	\begin{tabular}{lcccccc}
		\toprule
		Model & Pruning rate (\%) & Lat. (ms) & Lat. red. & CO$_2$eq (g) & CO$_2$/ex (mg) & CO$_2$ red. \\
		\midrule
		Full model       & --      & 951.5 & --   & 11.04 & 17.58 & --   \\
		FAIR-Pruner & 50.0 & 721.4 & 24.19\% & 8.49  & 13.52 & 23.08\% \\
		FAIR-Pruner & 75.0 & 595.8 & 37.38\% & 6.54  & 10.42 & 40.73\% \\
		FAIR-Pruner & 83.3 & 541.1 & 43.14\% & 5.87  & 9.35  & 46.80\% \\
		\bottomrule
	\end{tabular}
\end{table*}
\paragraph{Green inference analysis}
We further evaluate the inference carbon footprint using CodeCarbon. 
To avoid confounding different inference implementations, we compare the full-size and pruned models under the same wrapped MoE implementation. 
CodeCarbon tracking is started after model loading and warm-up, so the reported emissions correspond only to the measured generation loop.

As shown in Table~\ref{tab:green_inference_wic}, FAIR-Pruner yields a clear reduction in inference latency and estimated carbon emissions under the implementation-matched setting. 
At the 50\% routed-expert pruning budget, FAIR-Pruner reduces latency from 951.5ms to 721.4ms per example and lowers the estimated emissions from 11.04g to 8.49g CO$_2$eq, corresponding to a 23.08\% carbon reduction. 
More aggressive budgets further improve green inference efficiency: keep15 and keep10 reduce the estimated emissions by 40.73\% and 46.80\%, respectively.

\section{Discussion}

FAIR-Pruner suggests a modular view of adaptive structured pruning. Rather than
searching for one universal importance score, it separates the pruning decision
into three roles: a removal-oriented signal ranks structurally removable candidates,
a protection-oriented signal identifies units whose deletion may cause large
task-level loss changes, and ToD converts their conflict into layer-wise pruning
depths. This separation is useful because a unit may appear redundant from a
representation perspective but still be risky to delete, while simultaneous
deletion can introduce interactions that are not captured by singleton scores
alone.

ToD is a lightweight realization of this dual-signal principle. It does not
require architecture search, repeated retraining, or cross-layer calibration of
raw score magnitudes. Its behavior is also transparent: the selected depth is
monotone in the tolerance level, target budgets can be matched by sweeping
finitely many conflict values, and the high-\(R\)-Score mass entering the
removal set can be controlled in a rank-based sense. The same mechanism can be
paired with different removal signals, as shown by the L1, FPGM,
soft-activation, and routed-expert experiments.

The present theory supports ToD as a rank-based conflict-control rule, not as an
unconditional dominance result over all uniform or single-score methods.
Rank-based control avoids fragile cross-layer score calibration, but it cannot
encode all magnitude information relevant to global optimality. Future work may
therefore combine ToD-style conflict control with magnitude-aware risk budgets,
interaction-aware protection scores, or deployment-aware cost models. The
reusable part of FAIR-Pruner is the separation of removal ranking, sensitivity
protection, and layer-wise allocation.

\section{Conclusion}

We introduced FAIR-Pruner, a search-free framework for adaptive layer-wise structured pruning. 
Its core component, Tolerance of Difference (ToD), coordinates a removal-oriented signal and a protection-oriented signal to determine non-uniform layer-wise pruning depths without architecture search. 
As a default instantiation, FAIR-Pruner combines a Wasserstein-based U-Score with a Taylor-based R-Score, while the framework itself remains modular and can incorporate alternative removal signals.
The theoretical analysis controls the population R-Score \(R_j^{(l)}\) itself,
rather than a separate derivative surrogate: ToD bounds the high-sensitivity
mass entering the selected pruning set, and the additive-loss analysis gives a
conditional optimality statement for comparing this risk-controlled allocation
with uniform pruning under an explicit same-budget exchange condition.

Experiments on classical CNN benchmarks show that FAIR-Pruner achieves competitive accuracy--efficiency trade-offs against recent structured pruning methods. 
Additional results on ConvNeXt and DeiT-B indicate that ToD can be adapted to modern block-structured architectures, while prune-only experiments on Qwen1.5-MoE show that the same allocation principle can be instantiated for routed-expert pruning under matched expert budgets. 
Ablation and extensibility studies further confirm that the benefit of FAIR-Pruner comes not only from a particular U-Score instantiation, but also from the ToD mechanism for adaptive allocation. 
Together with the latency and memory-footprint results, these findings support FAIR-Pruner as a practical and extensible framework for adaptive structured pruning.

The main message is that adaptive structured pruning need not rely on
architecture search or globally calibrated saliency values. By measuring the
conflict between a removal ranking and a protection ranking, ToD provides a
simple and reusable mechanism for turning complementary unit-level evidence
into layer-wise pruning budgets. This framework perspective suggests future
work on stronger architecture-specific signals while retaining the same
principle of dual-signal allocation.

\section*{Acknowledgments}
This work was supported by the Laboratory for Statistical Monitoring and Intelligent Governance of Common Prosperity, the Characteristic \& Preponderant Discipline of Key Construction Universities in Zhejiang Province (Zhejiang Gongshang University--Statistics), and the Collaborative Innovation Center of Statistical Data Engineering Technology \& Application.

\bibliographystyle{IEEEtran}
\bibliography{Reference}

\clearpage
\appendices

\section{Additional Methodological Details}
\label{app:method_details}
\subsection{Target-Budget Selection of the ToD Level}
\label{app:target_budget}

In many applications, the desired compression level is specified not by a ToD
level itself, but by a target budget, such as the number of remaining
channels, parameters, FLOPs, or routed experts. We describe here how the ToD
level can be selected after the U-Scores and R-Scores have been computed.

Let $c_l>0$ denote the cost associated with removing one unit from layer $l$.
For channel pruning, $c_l$ may approximate the parameter or FLOPs reduction
induced by removing one channel; for routed-expert pruning, $c_l=1$ counts one
routed expert. The empirical removed budget induced by $\alpha$ is
\[
\widehat B(\alpha)
=
\sum_{l\in L} c_l \widehat m^{(l)}(\alpha).
\]
Given a target removed budget $B_0$, we select
\[
\widehat\alpha_{B_0}
\in
\arg\min_{\alpha\in\widehat{\mathcal A}}
\left|
\widehat B(\alpha)-B_0
\right|,
\]
where
\[
\widehat{\mathcal A}
:=
\{0,1\}
\cup
\left\{
\widehat{\mathrm{ToD}}^{(l)}(m):
l\in L,\ 1\le m\le J^{(l)}
\right\}.
\]
Because $\widehat m^{(l)}(\alpha)$ is a step function of $\alpha$, it is
sufficient to search over the finite set $\widehat{\mathcal A}$. This is not
an architecture search: the scores are computed once, and changing $\alpha$
only sweeps the already computed conflict curves.

\begin{proposition}[Stepwise budget control]
	\label{prop:stepwise_budget_control}
	For fixed empirical scores, $\widehat B(\alpha)$ is nondecreasing and
	piecewise constant in $\alpha$. Moreover, all changes of $\widehat B(\alpha)$
	occur at values in $\widehat{\mathcal A}$. Hence a budget-matched ToD model
	can be obtained by a finite sweep over $\widehat{\mathcal A}$ without
	retraining or re-estimating scores.
\end{proposition}

Because $\widehat B(\alpha)$ is monotone, the finite sweep can also be
implemented by a lightweight binary search over the sorted breakpoints in
$\widehat{\mathcal A}$.

\subsection{Empirical Behavior of the ToD Level}
\label{app:alpha_empirical_behavior}

The monotonicity result in Proposition~\ref{prop:alpha_monotone} concerns the
selected pruning depth \(\widehat m^{(l)}(\alpha)\) as a function of the
tolerance level \(\alpha\). It does not require the conflict curve
\(\widehat{\mathrm{ToD}}^{(l)}(m)\) itself to be monotone in the candidate depth
\(m\). In fact, as illustrated by the representative layer example in
Figure~\ref{fig:algorithm-overview}, \(\widehat{\mathrm{ToD}}^{(l)}(m)\) may
fluctuate with \(m\),
because adding one more unit to the removal prefix may or may not add a unit
from the protected R-Score tail. The selection rule is nevertheless monotone in
\(\alpha\), because increasing \(\alpha\) only enlarges the feasible set of
depths.

\begin{figure}[t]
	\centering
	\includegraphics[width=\linewidth]{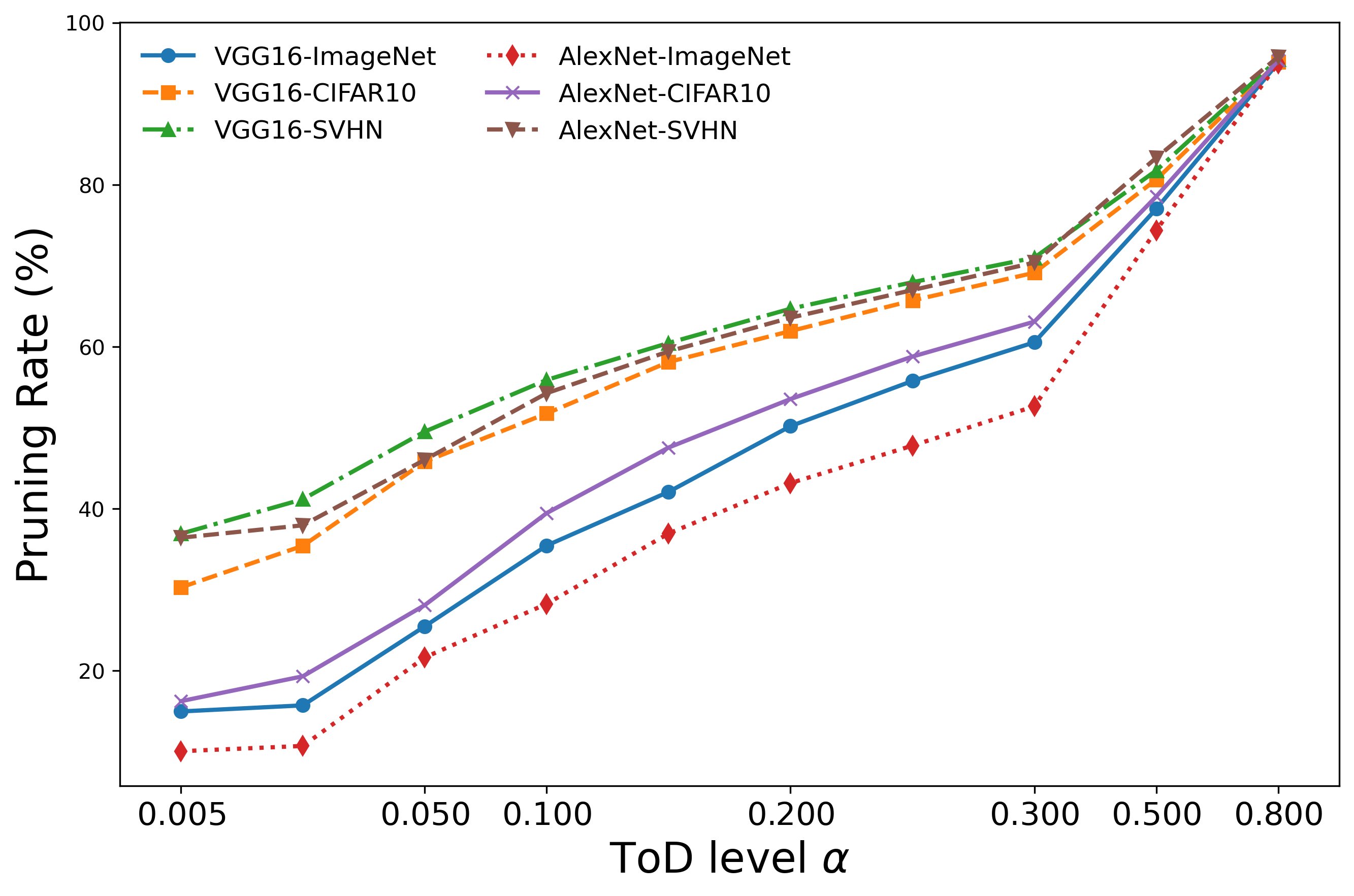}
	\caption{Achieved pruning rate as a function of the ToD level \(\alpha\).
		The selected pruning rate is monotone in \(\alpha\), consistent with
		Proposition~\ref{prop:alpha_monotone}.}
	\label{fig:alpha_pr_appendix}
\end{figure}

Figure~\ref{fig:alpha_pr_appendix} provides an empirical diagnostic of this
behavior. The achieved pruning rate increases monotonically with the ToD level
across the evaluated VGG-16 and AlexNet settings, including CIFAR-10, SVHN, and
ImageNet. The curves are stepwise in principle, but the observed operating
points show that changing \(\alpha\) gives a stable way to sweep from
conservative to aggressive pruning regimes after the scores have been computed.
This supports the use of a finite sweep or binary search over \(\alpha\) when a
target pruning budget is required. In these reported sweeps, the selected
pruning depths are positive for the operating ranges used in our experiments,
indicating that the perfect-opposition zero-depth regime from
Appendix~\ref{app:rank_regimes} does not occur in these diagnostics.

\begin{figure*}[t]
	\centering
	\subfloat[Indexed by ToD level.]{
		\includegraphics[width=0.43\textwidth]{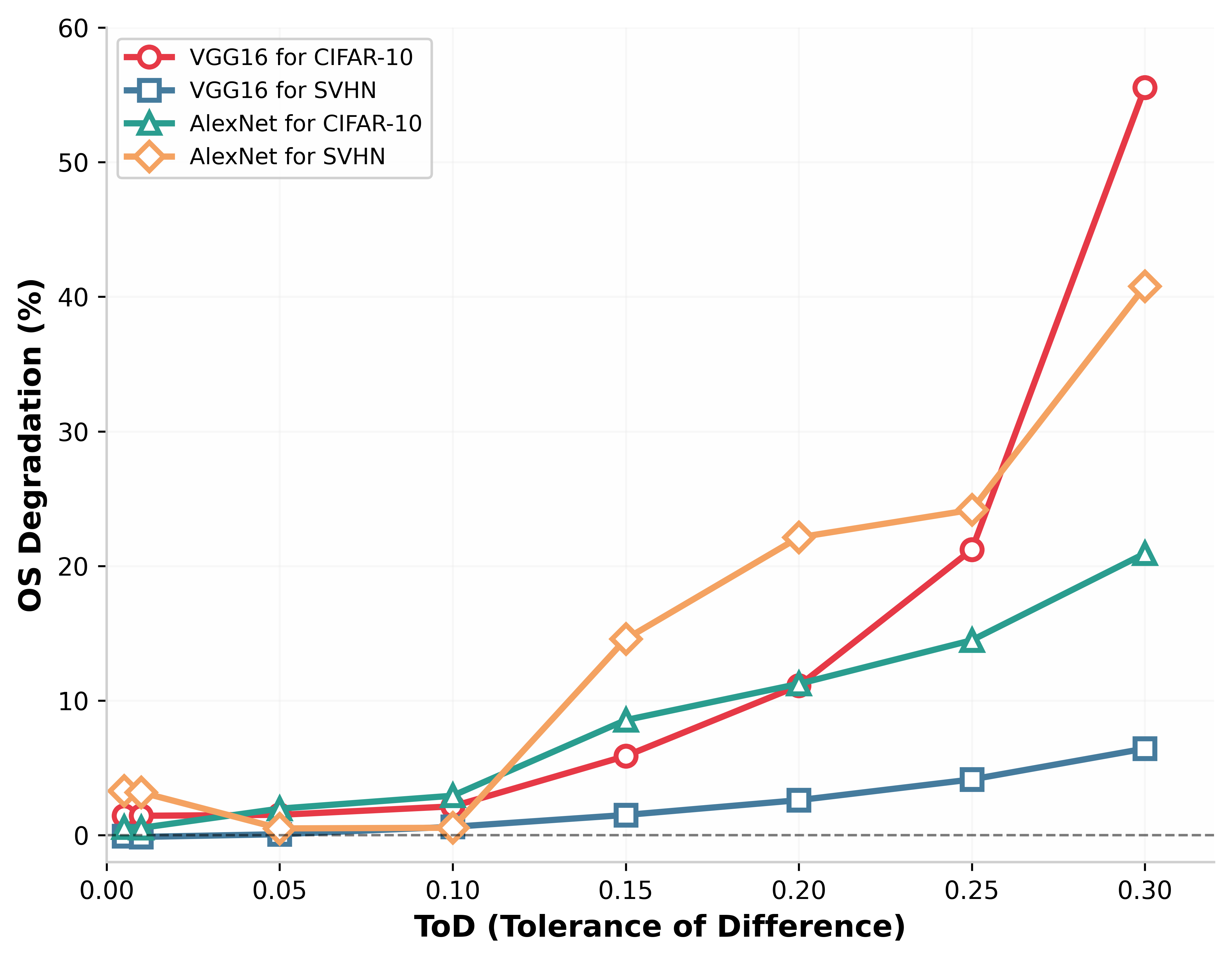}
		\label{fig:alpha_os_appendix}
	}
	\hfill
	\subfloat[Indexed by achieved pruning rate.]{
		\includegraphics[width=0.43\textwidth]{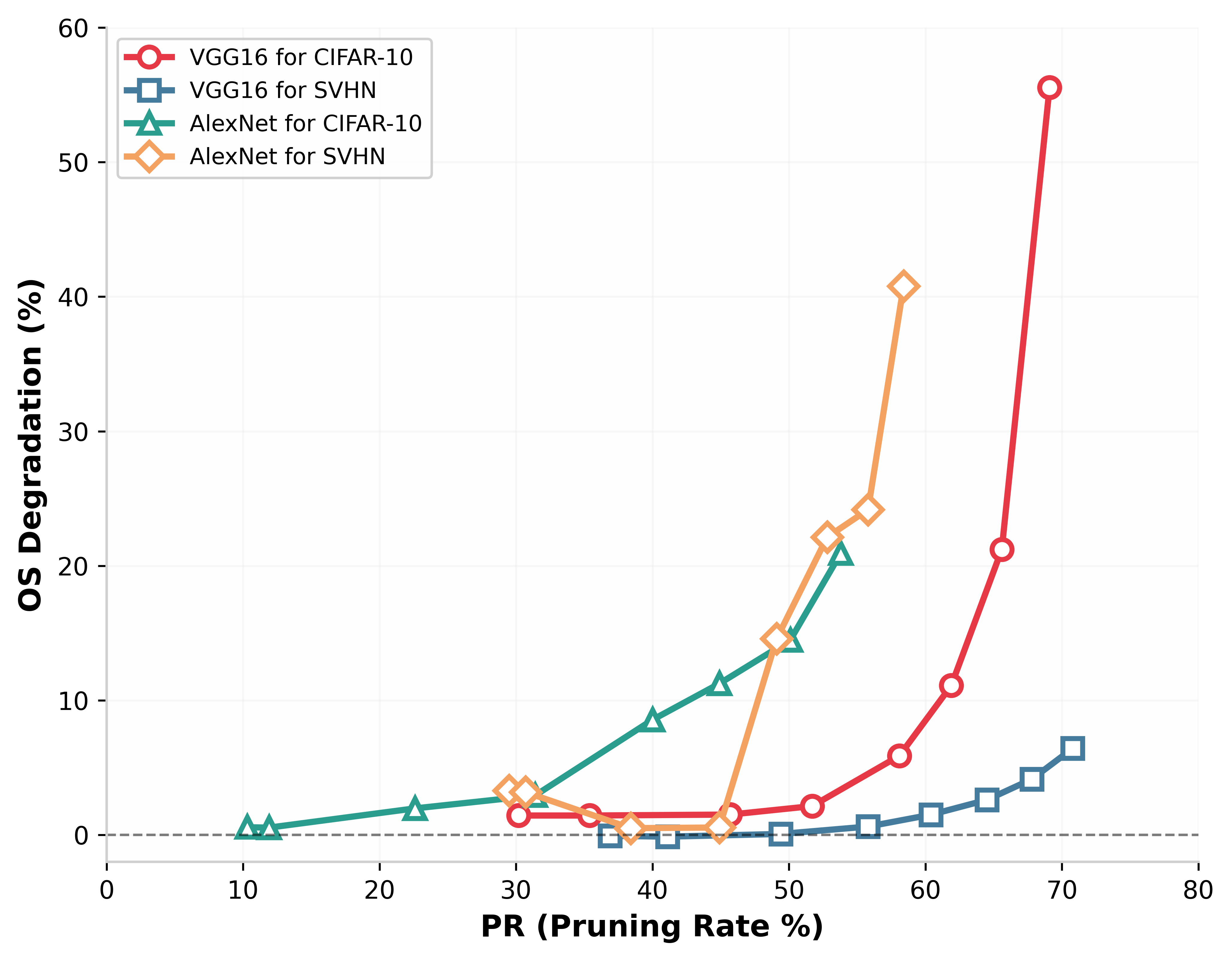}
		\label{fig:pr_os_appendix}
	}
	\caption{One-shot degradation under two indexing variables. Viewing the
		degradation as a function of the ToD level gives a more consistent
		aggressiveness scale than viewing it directly as a function of the achieved
		pruning rate.}
	\label{fig:alpha_degradation_appendix}
\end{figure*}

Figure~\ref{fig:alpha_degradation_appendix} further compares one-shot
degradation when pruning outcomes are indexed by \(\alpha\) or by the achieved
pruning rate. The ToD-level view gives
a more consistent notion of pruning aggressiveness across architectures and
datasets: small ToD levels correspond to conservative decisions, whereas larger
ToD levels admit more conflict between the removal prefix and the protected
tail and therefore lead to higher risk. By contrast, the same raw pruning rate
can be mild for one model but aggressive for another. These plots are not a
formal accuracy guarantee; they are empirical evidence that the global
tolerance level is a useful diagnostic control variable and that the selected
depths do not exhibit problematic aggregate jumps in the tested settings.

\section{Auxiliary theoretical results}
\subsection{Notation}
For a linear operator $A$ between Hilbert spaces, $\|A\|_{\mathrm{op}} := \sup_{\|x\|=1} \|Ax\|$ denotes the operator norm. For a multilinear form $T$ of order $m$, $\|T\|_{\mathrm{op}}$ is defined as the supremum of $|T(v_1,\dots,v_m)|$ over unit vectors $v_1,\dots,v_m$. The notation $D_z^m G(z)[\delta_1,\dots,\delta_m]$ represents the $m$-th order Fr\'echet derivative of $G$ at $z$ evaluated on the tuple $(\delta_1,\dots,\delta_m)$.

\subsection{Tie Handling and Near-Tie Interpretation}
\label{app:ties}
Assumption~\ref{ass:rank_separation} is used to make the population rankings
unique and to state a clean selection-consistency result. In implementation,
ties are resolved by a deterministic tie-breaking rule: units are sorted
lexicographically by \((\widehat U_j^{(l)},j)\) for the removal ranking and by
\((-\widehat R_j^{(l)},j)\) for the protection ranking. If exact population ties
occur, Theorem~\ref{thm:selection_consistency} should be interpreted as recovery
of a tied equivalence class rather than recovery of a unique index set. Near-ties
may change which unit inside a nearly indistinguishable boundary block is
selected, but they do not change the role of ToD as a conflict-control rule
between the removal prefix and the protected high-R-Score tail.

\subsection{U-Score Consistency}
\label{app:uscore_consistency}
For $l \in L$ and $j\in[J^{(l)}]$, let $F_{O^{(l)}_{j}(Z_k)}$ and $\widehat{F}_{O^{(l)}_{j},n_k}$ denote the cumulative distribution function and the empirical distribution function of $O^{(l)}_{j}(Z_k)$, respectively. Let $\widehat{O}^{(l)}_{j, n_k}$ be the random variable having distribution $\widehat{F}_{O^{(l)}_{j},n_k}$.

\begin{proposition}[U-Score consistency]
	\label{prop:uscore_consistency}
	For any $l\in L$ and $j\in[J^{(l)}]$, let
	\begin{equation}
		\widehat U_j^{(l)}
		:=
		\sup_{\substack{k_1\neq k_2 \in [K]}}
		d\bigl(\widehat{O}^{(l)}_{j, n_{k_1}}, \widehat{O}^{(l)}_{j, n_{k_2}}\bigr)
	\end{equation}
	be the empirical estimator of $U_j^{(l)}$.
	Suppose $K <\infty$ and $\mathbb E|O_j^{(l)}(Z_{k})|<\infty$ for any $k\in[K]$. Then, as $\min_{k\in[K]}n_k\to\infty$, we have
	\begin{equation}
		\widehat U_j^{(l)}\xrightarrow[]{\mathrm{a.s.}} U_j^{(l)}.
		\label{eq:uscore_as}
	\end{equation}
\end{proposition}
\begin{proof}
	Fix $l\in L$ and $j\in[J^{(l)}]$. For every ordered class pair
	$(k_1,k_2)$ with $k_1\neq k_2$, define
	\begin{equation}
		D_{k_1,k_2}^{(l,j)} 
		:= 
		d\!\biggl(O_j^{(l)}(Z_{k_1}),\,O_j^{(l)}(Z_{k_2})\biggr),
	\end{equation}
	and
	\begin{equation}
		\widehat D_{k_1,k_2}^{(l,j)} 
		:= 
		d\!\biggl(\widehat O_{j,n_{k_1}}^{(l)},\,\widehat O_{j,n_{k_2}}^{(l)}\biggr).
	\end{equation}
	Then
	\begin{equation}
		\begin{aligned}
			&U_j^{(l)}=\sup_{k_1\neq k_2} D_{k_1,k_2}^{(l,j)}, \qquad \\
			&\widehat U_j^{(l)}=\sup_{k_1\neq k_2} \widehat D_{k_1,k_2}^{(l,j)}.
		\end{aligned}
	\end{equation}
	
	Because $K<\infty$, the number of class pairs is finite. Hence it
	suffices to prove that, for each fixed $(k_1,k_2)$,
	\begin{equation}
		\widehat D_{k_1,k_2}^{(l,j)} 
		\overset{a.s.}{\longrightarrow} 
		D_{k_1,k_2}^{(l,j)}.
	\end{equation}
	
	Under $\mathbb E|O_j^{(l)}(Z_k)|<\infty$, the empirical distributions
	converge almost surely to their population distributions in
	$1$-Wasserstein distance. Therefore,
	\begin{equation}
		\begin{aligned}
			&d\!\biggl(\widehat O_{j,n_{k_1}}^{(l)},\,O_j^{(l)}(Z_{k_1})\biggr) 
			\overset{a.s.}{\longrightarrow}0, \qquad \\
			&d\!\biggl(\widehat O_{j,n_{k_2}}^{(l)},\,O_j^{(l)}(Z_{k_2})\biggr) 
			\overset{a.s.}{\longrightarrow}0.
		\end{aligned}
	\end{equation}
	By the triangle inequality of $d$,
	\begin{equation}
		\begin{aligned}
			&\quad\biggl| \widehat D_{k_1,k_2}^{(l,j)}-D_{k_1,k_2}^{(l,j)} \biggr| \\
			\le\, &\,\,
			d\!\biggl(\widehat O_{j,n_{k_1}}^{(l)},\,O_j^{(l)}(Z_{k_1})\biggr) + 
			d\!\biggl(\widehat O_{j,n_{k_2}}^{(l)},\,O_j^{(l)}(Z_{k_2})\biggr),
		\end{aligned}
	\end{equation}
	which converges almost surely to zero. Hence
	\begin{equation}
		\begin{aligned}
			\widehat D_{k_1,k_2}^{(l,j)} 
			\overset{a.s.}{\longrightarrow} 
			D_{k_1,k_2}^{(l,j)},\quad
			\text{for every fixed }(k_1,k_2).
		\end{aligned}
	\end{equation}
	
	Since the number of class pairs is finite, the supremum preserves
	almost sure convergence:
	\begin{equation}
		\begin{aligned}
			\widehat U_j^{(l)} 
			= 
			\sup_{k_1\neq k_2}\widehat D_{k_1,k_2}^{(l,j)} 
			\overset{a.s.}{\longrightarrow} 
			\sup_{k_1\neq k_2} D_{k_1,k_2}^{(l,j)} 
			= 
			U_j^{(l)}.
		\end{aligned}
	\end{equation}
	This proves the result.
\end{proof}

Based on~\eqref{eq:uscore_as}, the quantity $\widehat U_j^{(l)}$ is used as the empirical U-Score of the $j$-th unit in the $l$-th layer. 

\subsection{Sufficient Conditions for Removal-Score Structural Compatibility}
\label{app:structural_proxy_sufficient}
The structural compatibility condition~\eqref{eq:structural_compatibility} is
not a consequence of the Wasserstein U-Score definition alone. The following
proposition records a simple sufficient condition under which the U-Score prefix
controls the structural perturbation envelope used in
Corollary~\ref{cor:tod_loss_envelope}.

\begin{proposition}[A sufficient condition for structural-prefix control]
	\label{prop:structural_proxy_sufficient}
	Fix a layer \(l\in L\). Let \(U_{(1)}^{(l)}\le\cdots\le
	U_{(J^{(l)})}^{(l)}\) denote the U-Scores in nondecreasing order. Suppose
	there exist nonnegative activation-size proxies
	\(M_j^{(l)}\) and constants \(L_l,\rho_l,\eta_l\ge0\) such that, for every
	unit \(j\in[J^{(l)}]\),
	\begin{equation}
		b_{(l,j)}\le L_l M_j^{(l)}
		\label{eq:b_by_activation_proxy}
	\end{equation}
	and
	\begin{equation}
		M_j^{(l)}\le \rho_l U_j^{(l)}+\eta_l .
		\label{eq:activation_by_uscore}
	\end{equation}
	Then the structural compatibility condition~\eqref{eq:structural_compatibility}
	holds with
	\begin{equation}
		B_l^U(m)
		=
		L_l\left(
		\rho_l\sum_{r=1}^{m}U_{(r)}^{(l)}
		+
		m\eta_l
		\right),
		\qquad m=0,1,\ldots,J^{(l)}.
	\end{equation}
\end{proposition}

\begin{proof}
	Let \(\mathcal R^{(l)}(m)\) be the set of the \(m\) smallest U-Scores in
	layer \(l\). By~\eqref{eq:b_by_activation_proxy} and
	\eqref{eq:activation_by_uscore},
	\begin{equation}
		\begin{aligned}
			\sum_{j\in\mathcal R^{(l)}(m)}b_{(l,j)}
			&\le
			L_l\sum_{j\in\mathcal R^{(l)}(m)}M_j^{(l)}
			\\
			&\le
			L_l\sum_{j\in\mathcal R^{(l)}(m)}
			\left(\rho_l U_j^{(l)}+\eta_l\right)
			\\
			&=
			L_l\left(
			\rho_l\sum_{r=1}^{m}U_{(r)}^{(l)}
			+
			m\eta_l
			\right).
		\end{aligned}
	\end{equation}
	The displayed envelope is nondecreasing in \(m\) because U-Scores and
	\(\eta_l\) are nonnegative.
\end{proof}

\begin{remark}[Why the compatibility is not automatic]
	The condition above rules out a possible common-mode failure of
	class-conditional separability. For example, suppose a unit has the same
	output distribution under every class, concentrated near a large value \(a\).
	Then its Wasserstein U-Score can be zero because the class-conditional
	distributions are identical, while deleting the unit may still induce a
	large downstream perturbation proportional to \(a\) and to the corresponding
	downstream weight norm. Hence no unconditional bound from Wasserstein
	class-separability to \(b_{(l,j)}\) can hold without an additional
	activation-size or common-mode control such as~\eqref{eq:activation_by_uscore}.
\end{remark}

\subsection{Perturbation interpretation of interaction terms}
\label{app:interaction_derivative_interpretation}
\begin{proposition}[Layerwise perturbation representation of interactions]
	\label{prop:interaction_derivative_interpretation}
	Fix a prunable layer $l\in L$, and let $\mathcal U_l\subseteq \mathcal U$ denote the set of prunable units in that layer. 
	Let $h^{(l)}(X)\in\mathbb R^{d_l}$ denote the layer-$l$ representation in the original network, and assume that there exists a measurable downstream loss map
	\begin{equation}
		G_l:\mathbb R^{d_l}\times \mathcal X\times \mathcal Y \to \mathbb R
	\end{equation}
	such that
	\begin{equation}
		\mathcal{L}(\hat f(X),Y)=G_l(h^{(l)}(X);X,Y).
	\end{equation}
	For each unit $u\in \mathcal U_l$, let $\delta_u(X)\in\mathbb R^{d_l}$ denote the perturbation of the layer-$l$ representation induced by removing unit $u$, and assume that for every finite set $T\subseteq \mathcal U_l$,
	\begin{equation}
		\mathcal{L}(\hat f_{-T}(X),Y)
		=
		G_l\!\left(
		h^{(l)}(X)+\sum_{u\in T}\delta_u(X);X,Y
		\right).
	\end{equation}
	Suppose that for almost every $(X,Y)$ and every integer $m\ge2$, the map
	$z\mapsto G_l(z;X,Y)$ is $C^m$ on an open neighborhood containing
	\begin{equation}
		\left\{
		h^{(l)}(X)+\sum_{u\in T} t_u \delta_u(X) : (t_u)_{u\in T}\in [0,1]^{|T|}
		\right\}
	\end{equation}
	for every finite set $T\subseteq \mathcal U_l$ with $|T|=m$.
	
	Assume further that for every integer $m\ge 2$ and every finite set
	$T=\{u_1,\dots,u_m\}\subseteq \mathcal U_l$, there exists a nonnegative measurable random variable
	$H_m^{(l)}(X,Y)$ such that
	\begin{equation}
		\begin{aligned}
			\sup_{(t_1,\dots,t_m)\in[0,1]^m}&
			\left\|
			D_z^m G_l\!\left(
			h^{(l)}(X)+\sum_{i=1}^m t_i\delta_{u_i}(X);X,Y
			\right)
			\right\|_{\mathrm{op}}\\
			\le& H_m^{(l)}(X,Y)
		\end{aligned}
	\end{equation}
	almost surely, and
	\begin{equation}
		\mathbb E\!\left[
		H_m^{(l)}(X,Y)\prod_{i=1}^m \|\delta_{u_i}(X)\|
		\right] < \infty,
	\end{equation}
	where $\|\cdot\|$ denotes the Euclidean norm for vectors and the Frobenius norm for matrices.
	
	Then:
	
	\textup{(i)} For every pair of distinct units $\{u,v\}\subseteq \mathcal U_l$,
	\begin{equation}
		\begin{aligned}
			I_{u,v}
			=
			\mathbb E\!\biggl[
			&\int_0^1\!\!\int_0^1
			\delta_u(X)^\top
			\nabla_z^2 G_l\!\bigl(
			h^{(l)}(X)+s\delta_u(X)\\
			&+t\delta_v(X);X,Y
			\bigr)
			\delta_v(X)\,ds\,dt
			\biggr].
		\end{aligned}
	\end{equation}
	
	\textup{(ii)} More generally, for every finite set
	$T=\{u_1,\dots,u_m\}\subseteq \mathcal U_l$ with $m\ge 3$,
	\begin{equation}
		\begin{aligned}
			\kappa(T)
			&=
			\mathbb E\!\biggl[
			\int_{[0,1]^m}
			D_z^m G_l\!\left(
			h^{(l)}(X)+\sum_{i=1}^m t_i\delta_{u_i}(X);X,Y
			\right)\\
			&\hspace{3em}\times
			[\delta_{u_1}(X),\ldots,\delta_{u_m}(X)]
			\,dt_1\cdots dt_m
			\biggr],
		\end{aligned}
	\end{equation}
	
	\textup{(iii)} For every pair of distinct units $\{u,v\}\subseteq \mathcal U_l$,
	\begin{equation}
		|I_{u,v}|
		\le
		\mathbb E\!\bigl[
		H_2^{(l)}(X,Y)\|\delta_u(X)\|\|\delta_v(X)\|
		\bigr],
	\end{equation}
	and for every finite set $T\subseteq \mathcal U_l$ with $|T|=m\ge 3$,
	\begin{equation}
		|\kappa(T)|
		\le
		\mathbb E\!\left[
		H_m^{(l)}(X,Y)\prod_{u\in T}\|\delta_u(X)\|
		\right].
	\end{equation}
	Consequently, for every finite $S\subseteq \mathcal U_l$,
	\begin{equation}
		\sum_{\substack{T\subseteq S\\ |T|\ge 3}} |\kappa(T)|
		\le
		\sum_{m=3}^{|S|}
		\sum_{\substack{T\subseteq S\\|T|=m}}
		\mathbb E\!\left[
		H_m^{(l)}(X,Y)\prod_{u\in T}\|\delta_u(X)\|
		\right].
	\end{equation}
\end{proposition}
\begin{proof}
	Fix a layer $l\in L$ and suppress the superscript $(l)$ when no confusion arises.
	
	\medskip
	\noindent\textbf{Step 1: Pairwise interaction formula.}
	Let $\{u,v\}\subseteq \mathcal U_l$ be a pair of distinct units. For each realization $(x,y)$, define
	\begin{equation}
		h:=h^{(l)}(x),\qquad \delta_u:=\delta_u(x),\qquad \delta_v:=\delta_v(x),
	\end{equation}
	and introduce the function
	\begin{equation}
		\phi_{x,y}(s,t):=G_l(h+s\delta_u+t\delta_v;x,y),
		\qquad (s,t)\in[0,1]^2.
	\end{equation}
	By the assumed local $C^2$ smoothness of $G_l$ along the perturbation path, $\phi_{x,y}$ is $C^2$ on $[0,1]^2$.
	
	By the layerwise perturbation representation,
	\begin{equation}
		\begin{aligned}
			\mathcal L(\hat f(x),y)&=\phi_{x,y}(0,0),&
			\mathcal L(\hat f_{-\{u\}}(x),y)&=\phi_{x,y}(1,0),\\
			\mathcal L(\hat f_{-\{v\}}(x),y)&=\phi_{x,y}(0,1),&
			\mathcal L(\hat f_{-\{u,v\}}(x),y)&=\phi_{x,y}(1,1).
		\end{aligned}
	\end{equation}
	Hence
	\begin{equation}
		I_{u,v}
		=
		\mathbb E\!\left[
		\begin{aligned}
			&\phi_{X,Y}(1,1)-\phi_{X,Y}(1,0)\\
			&\quad-\phi_{X,Y}(0,1)+\phi_{X,Y}(0,0)
		\end{aligned}
		\right].
	\end{equation}
	
	Since $\phi_{x,y}$ is $C^2$, the two-dimensional fundamental theorem of calculus gives
	\begin{equation}
		\begin{aligned}
			&\phi_{x,y}(1,1)-\phi_{x,y}(1,0)-\phi_{x,y}(0,1)+\phi_{x,y}(0,0)\\
			=&
			\int_0^1\!\!\int_0^1
			\frac{\partial^2 \phi_{x,y}}{\partial s\,\partial t}(s,t)\,ds\,dt.
		\end{aligned}
	\end{equation}
	By the chain rule,
	\begin{equation}
		\frac{\partial^2 \phi_{x,y}}{\partial s\,\partial t}(s,t)
		=
		\delta_u^\top
		\nabla_z^2 G_l(h+s\delta_u+t\delta_v;x,y)
		\delta_v.
	\end{equation}
	
	Moreover, by the operator norm inequality,
	\begin{equation}
		\begin{aligned}
			&\left|
			\delta_u^\top
			\nabla_z^2 G_l(h+s\delta_u+t\delta_v;x,y)
			\delta_v
			\right|\\
			\le\,&
			\left\|
			\nabla_z^2 G_l(h+s\delta_u+t\delta_v;x,y)
			\right\|_{\mathrm{op}}
			\|\delta_u\|\,\|\delta_v\|.
		\end{aligned}
	\end{equation}
	By the envelope assumption with $m=2$,
	\begin{equation}
		\left\|
		\nabla_z^2 G_l(h+s\delta_u+t\delta_v;x,y)
		\right\|_{\mathrm{op}}
		\le H_2^{(l)}(x,y),
	\end{equation}
	so that
	\begin{equation}
		\left|
		\frac{\partial^2 \phi_{x,y}}{\partial s\,\partial t}(s,t)
		\right|
		\le
		H_2^{(l)}(x,y)\|\delta_u\|\,\|\delta_v\|.
	\end{equation}
	Restoring random variables, we obtain
	\begin{equation}
		\left|
		\frac{\partial^2 \phi_{X,Y}}{\partial s\,\partial t}(s,t)
		\right|
		\le
		H_2^{(l)}(X,Y)\|\delta_u(X)\|\,\|\delta_v(X)\|.
	\end{equation}
	The right-hand side is integrable by assumption. Therefore, Fubini's theorem applies, and we may take expectation after integrating over $(s,t)\in[0,1]^2$. Substituting the chain-rule identity yields
	\begin{equation}
		\begin{aligned}
			I_{u,v}
			=
			\mathbb E\!\biggl[
			&\int_0^1\!\!\int_0^1
			\delta_u(X)^\top
			\nabla_z^2 G_l\!\bigl(
			h^{(l)}(X)+s\delta_u(X)\\
			&+t\delta_v(X);X,Y
			\bigr)
			\delta_v(X)\,ds\,dt
			\biggr].
		\end{aligned}
	\end{equation}
	
	\medskip
	\noindent\textbf{Step 2: Higher-order interaction formula.}
	Now let $T=\{u_1,\dots,u_m\}\subseteq \mathcal U_l$ with $m\ge 3$. For each realization $(x,y)$, define
	\begin{equation}
		h:=h^{(l)}(x),\qquad \delta_i:=\delta_{u_i}(x),\quad i=1,\dots,m,
	\end{equation}
	and consider
	\begin{equation}
		\begin{aligned}
			\psi_{x,y}(t_1,\dots,t_m)
			:=
			G_l\!\left(
			h+\sum_{i=1}^m t_i\delta_i;x,y
			\right),
		\end{aligned}
	\end{equation}
	where $(t_1,\dots,t_m)\in[0,1]^m$. By assumption, $\psi_{x,y}$ is $C^m$ on $[0,1]^m$.
	
	For every subset $B\subseteq[m]$, let
	\begin{equation}
		T_B:=\{u_i:i\in B\},\qquad
		1_B:=(1\{1\in B\},\dots,1\{m\in B\}).
	\end{equation}
	By the layerwise perturbation representation,
	\begin{equation}
		\Delta \mathcal L(T_B)
		=
		\mathbb E\!\left[\psi_{X,Y}(1_B)-\psi_{X,Y}(0)\right].
	\end{equation}
	Therefore,
	\begin{equation}
		\begin{aligned}
			\kappa(T)
			&=
			\sum_{B\subseteq[m]}(-1)^{m-|B|}\Delta \mathcal L(T_B)\\
			&=
			\mathbb E\!\left[
			\sum_{B\subseteq[m]}(-1)^{m-|B|}\psi_{X,Y}(1_B)
			\right],
		\end{aligned}
	\end{equation}
	because
	\begin{equation}
		\sum_{B\subseteq[m]}(-1)^{m-|B|}=0.
	\end{equation}
	
	By repeated application of the one-dimensional fundamental theorem of calculus,
	\begin{equation}
		\begin{aligned}
			&\sum_{B\subseteq[m]}(-1)^{m-|B|}\psi_{x,y}(1_B)\\
			=&
			\int_{[0,1]^m}
			\frac{\partial^m \psi_{x,y}}{\partial t_1\cdots\partial t_m}(t_1,\dots,t_m)
			\,dt_1\cdots dt_m.
		\end{aligned}
	\end{equation}
	Again by the chain rule,
	\begin{equation}
		\begin{aligned}
			\frac{\partial^m \psi_{x,y}}{\partial t_1\cdots\partial t_m}(t_1,\dots,t_m)
			&=
			D_z^m G_l\!\left(
			h+\sum_{i=1}^m t_i\delta_i;x,y
			\right)\\
			&\quad
			[\delta_1,\dots,\delta_m].
		\end{aligned}
	\end{equation}
	
	By multilinearity and the operator norm inequality,
	\begin{equation}
		\begin{aligned}
			&\left|
			D_z^m G_l\!\left(
			h+\sum_{i=1}^m t_i\delta_i;x,y
			\right)
			[\delta_1,\dots,\delta_m]
			\right|\\
			\le\,&
			\left\|
			D_z^m G_l\!\left(
			h+\sum_{i=1}^m t_i\delta_i;x,y
			\right)
			\right\|_{\mathrm{op}}
			\prod_{i=1}^m \|\delta_i\|.
		\end{aligned}
	\end{equation}
	By the envelope assumption,
	\begin{equation}
		\left\|
		D_z^m G_l\!\left(
		h+\sum_{i=1}^m t_i\delta_i;x,y
		\right)
		\right\|_{\mathrm{op}}
		\le H_m^{(l)}(x,y),
	\end{equation}
	so that
	\begin{equation}
		\left|
		\frac{\partial^m \psi_{x,y}}{\partial t_1\cdots\partial t_m}(t_1,\dots,t_m)
		\right|
		\le
		H_m^{(l)}(x,y)\prod_{i=1}^m \|\delta_i\|.
	\end{equation}
	Restoring random variables, we obtain
	\begin{equation}
		\left|
		\frac{\partial^m \psi_{X,Y}}{\partial t_1\cdots\partial t_m}(t_1,\dots,t_m)
		\right|
		\le
		H_m^{(l)}(X,Y)\prod_{i=1}^m \|\delta_{u_i}(X)\|.
	\end{equation}
	The right-hand side is integrable by assumption. Therefore, Fubini's theorem applies, and we may exchange expectation and integration over $[0,1]^m$. Substituting the chain-rule identity yields
	\begin{equation}
		\begin{aligned}
			\kappa(T)
			=
			\mathbb E\!\biggl[
			&\int_{[0,1]^m}
			D_z^m G_l\!\biggl(
			h^{(l)}(X)+\sum_{i=1}^m t_i\delta_{u_i}(X);X,Y
			\biggr)\\
			&\times[\delta_{u_1}(X),\dots,\delta_{u_m}(X)]
			\,dt_1\cdots dt_m
			\biggr].
		\end{aligned}
	\end{equation}
	
	\medskip
	\noindent\textbf{Step 3: Bounds under derivative envelopes.}
	The bound for the pairwise interaction term follows directly from Step 1 and the envelope inequality:
	\begin{equation}
		|I_{u,v}|
		\le
		\mathbb E\!\bigl[
		H_2^{(l)}(X,Y)\|\delta_u(X)\|\|\delta_v(X)\|
		\bigr].
	\end{equation}
	Similarly, for every finite set $T\subseteq \mathcal U_l$ with $|T|=m\ge 3$, Step 2 yields
	\begin{equation}
		|\kappa(T)|
		\le
		\mathbb E\!\left[
		H_m^{(l)}(X,Y)\prod_{u\in T}\|\delta_u(X)\|
		\right].
	\end{equation}
	Finally, summing over all subsets $T\subseteq S$ with $|T|\ge 3$ gives
	\begin{equation}
		\begin{aligned}
			\sum_{\substack{T\subseteq S\\|T|\ge 3}}|\kappa(T)|
			\le
			\sum_{m=3}^{|S|}
			\sum_{\substack{T\subseteq S\\|T|=m}}
			\mathbb E\!\left[
			H_m^{(l)}(X,Y)\prod_{u\in T}\|\delta_u(X)\|
			\right].
		\end{aligned}
	\end{equation}
	This completes the proof.
\end{proof}
\begin{remark}[Translation into perturbation magnitudes]
	Proposition~\ref{prop:interaction_derivative_interpretation} shows that the pairwise interaction term is a curvature-weighted product of two induced representation perturbations, while the higher-order remainder is governed by higher-order multilinear interactions of these perturbations. 
	In particular, when pruning a unit amounts to zeroing out its coordinate at layer $l$, we have
	\begin{equation}
		\delta_u(X)=-h_u^{(l)}(X)e_u,
		\qquad
		\|\delta_u(X)\|=|h_u^{(l)}(X)|.
	\end{equation}
	Thus, activation-magnitude-type quantities enter the non-additive terms through the perturbation magnitudes. 
	This should not be interpreted as saying that activation magnitude alone determines pruning risk; rather, it explains one concrete route through which representation-level quantities contribute to the interaction terms beyond the single-unit population R-Score.
\end{remark}

\subsection{Canonical ranking regimes}
\label{app:rank_regimes}
Fix a layer $l\in L$ and write $J:=J^{(l)}$.
Assume that all population U-Scores and R-Scores in this layer are
distinct. Let $\pi_l(1),\ldots,\pi_l(J)$ be the ordering induced by
the U-Scores:
\begin{equation}
	U^{(l)}_{\pi_l(1)}<U^{(l)}_{\pi_l(2)}<\cdots<U^{(l)}_{\pi_l(J)}.
\end{equation}
For each $r\in[J]$, define
\begin{equation}
	\tau_l(r):=\operatorname{rank}_{R^{(l)}}(\pi_l(r)),
\end{equation}
where $\operatorname{rank}_{R^{(l)}}(u)\in[J]$ denotes the rank of
unit $u$ among the population R-Scores in increasing order, so that
rank $J$ corresponds to the largest R-Score.

\begin{proposition}[Rank representation and characteristic regimes]
	\label{prop:rank_regimes}
	For every $m\in[J]$,
	\begin{equation}
		\mathrm{ToD}^{(l)}(m)
		=
		\frac{1}{m}
		\sum_{r=1}^{m}
		\mathbf 1\!\left\{
		\tau_l(r)\ge J-m+1
		\right\}.
	\end{equation}
	Moreover, the following regimes hold.
	
	\textup{(i) Perfect alignment.}
	If the U- and R-Scores induce the same ordering,
	\begin{equation}
		U^{(l)}_{\pi_l(1)}<\cdots<U^{(l)}_{\pi_l(J)},
		\qquad
		R_{\pi_l(1)}^{(l)}<\cdots<R_{\pi_l(J)}^{(l)},
	\end{equation}
	then
	\begin{equation}
		\mathrm{ToD}^{(l)}(m)
		=
		\begin{cases}
			0, & 1\le m\le J/2,\\[4pt]
			2-\dfrac{J}{m}, & J/2<m\le J.
		\end{cases}
	\end{equation}
	Hence
	\begin{equation}
		m^{*(l)}(\alpha)
		=
		\left\lfloor
		\frac{J^{(l)}}{2-\alpha}
		\right\rfloor,\alpha \in [0,1).
	\end{equation}
	
	\textup{(ii) Perfect opposition.}
	If the U- and R-Scores induce opposite orderings,
	\begin{equation}
		U^{(l)}_{\pi_l(1)}<\cdots<U^{(l)}_{\pi_l(J)},
		\qquad
		R_{\pi_l(1)}^{(l)}>\cdots>R_{\pi_l(J)}^{(l)},
	\end{equation}
	then
	\begin{equation}
		\mathrm{ToD}^{(l)}(m)=1,
		\qquad \forall m\in[J],
	\end{equation}
	and therefore
	\begin{equation}
		m^{*(l)}(\alpha)=0,
		\qquad \forall \alpha<1.
	\end{equation}
	This is a zero-depth protection behavior rather than an ambiguity of the
	rule: if the units that look most removable are exactly those identified as
	most sensitive, ToD leaves the layer unpruned for any strict tolerance level.
	
	\textup{(iii) Random-baseline regime.}
	If, conditional on the population R-ranking, the population
	U-ranking is a uniformly random permutation of the $J$ units, then
	for every fixed $m\in[J]$,
	\begin{equation}
		H_l(m):=
		\left|
		\mathcal R^{(l)}(m)\cap \mathcal P^{(l)}(m)
		\right|
		\sim
		\mathrm{Hypergeometric}(J,m,m),
	\end{equation}
	so that
	\begin{equation}
		\mathbb E[\mathrm{ToD}^{(l)}(m)] = \frac{m}{J}.
	\end{equation}
\end{proposition}
\begin{proof}[Proof of Proposition~\ref{prop:rank_regimes}]
	Fix a layer $l\in L$ and write $J:=J^{(l)}$.
	
	By definition,
	\begin{equation}
		\mathcal R^{(l)}(m)=\{\pi_l(1),\dots,\pi_l(m)\}.
	\end{equation}
	Moreover,
	\begin{equation}
		\begin{aligned}
			\mathcal P^{(l)}(m) 
			&= 
			\biggl\{ u:\operatorname{rank}_{R^{(l)}}(u)\ge J-m+1 \biggr\}.
		\end{aligned}
	\end{equation}
	Therefore,
	\begin{equation}
		\begin{aligned}
			&u=\pi_l(r)\in \mathcal R^{(l)}(m)\cap \mathcal P^{(l)}(m) \\
			\quad\Longleftrightarrow\quad &
			1\le r\le m 
			\ \text{and}\ 
			\tau_l(r)\ge J-m+1.
		\end{aligned}
	\end{equation}
	Hence
	\begin{equation}
		\begin{aligned}
			\biggl| \mathcal R^{(l)}(m)\cap \mathcal P^{(l)}(m) \biggr| 
			&= 
			\sum_{r=1}^{m}\mathbf 1\!\biggl\{\tau_l(r)\ge J-m+1\biggr\},
		\end{aligned}
	\end{equation}
	and division by $m$ yields
	\begin{equation}
		\begin{aligned}
			\mathrm{ToD}^{(l)}(m) 
			&= 
			\frac{1}{m} 
			\sum_{r=1}^{m}\mathbf 1\!\biggl\{\tau_l(r)\ge J-m+1\biggr\}.
		\end{aligned}
	\end{equation}
	
	We now prove the three characteristic regimes.
	
	\medskip
	\noindent
	\textbf{(i) Perfect alignment.}
	If the U- and R-Scores induce the same ordering, then
	\begin{equation}
		\begin{aligned}
			\tau_l(r)=r, 
			\qquad r\in[J].
		\end{aligned}
	\end{equation}
	Therefore,
	\begin{equation}
		\begin{aligned}
			\mathrm{ToD}^{(l)}(m) 
			&= 
			\frac{1}{m}\sum_{r=1}^{m} 
			\mathbf 1\!\biggl\{r\ge J-m+1\biggr\}.
		\end{aligned}
	\end{equation}
	If $m\le J/2$, then $J-m+1>m$, so no index $r\in\{1,\dots,m\}$ can
	satisfy $r\ge J-m+1$. Hence
	\begin{equation}
		\mathrm{ToD}^{(l)}(m)=0.
	\end{equation}
	If $m>J/2$, then the qualifying indices are
	\begin{equation}
		r=J-m+1,\dots,m,
	\end{equation}
	whose number is
	\begin{equation}
		m-(J-m+1)+1=2m-J.
	\end{equation}
	Hence
	\begin{equation}
		\mathrm{ToD}^{(l)}(m)=\frac{2m-J}{m}=2-\frac{J}{m}.
	\end{equation}
	
	To determine the induced pruning depth, note that for $m\le J/2$ the
	constraint $\mathrm{ToD}^{(l)}(m)\le \alpha$ is always satisfied.
	For $m>J/2$, the constraint becomes
	\begin{equation}
		\begin{aligned}
			2-\frac{J}{m}\le \alpha 
			\quad\Longleftrightarrow\quad 
			m\le \frac{J}{2-\alpha}.
		\end{aligned}
	\end{equation}
	Taking the largest feasible integer gives
	\begin{equation}
		\begin{aligned}
			m^{*(l)}(\alpha) 
			&= 
			\biggl\lfloor \frac{J^{(l)}}{2-\alpha}\biggr\rfloor.
		\end{aligned}
	\end{equation}
	
	\medskip
	\noindent
	\textbf{(ii) Perfect opposition.}
	If the U- and R-Scores induce opposite orderings, then
	\begin{equation}
		\begin{aligned}
			\tau_l(r)=J-r+1, 
			\qquad r\in[J].
		\end{aligned}
	\end{equation}
	Hence, for every $r\le m$,
	\begin{equation}
		\tau_l(r)=J-r+1\ge J-m+1.
	\end{equation}
	So every one of the first $m$ U-ranked units belongs to the protected
	top-$m$ R-tail, which implies
	\begin{equation}
		\begin{aligned}
			\mathcal R^{(l)}(m)=\mathcal P^{(l)}(m) 
			\qquad\text{and}\qquad 
			\mathrm{ToD}^{(l)}(m)=1
		\end{aligned}
	\end{equation}
	for all $m\in[J]$. Therefore, if $\alpha<1$, no positive $m$ is
	feasible, and thus
	\begin{equation}
		m^{*(l)}(\alpha)=0.
	\end{equation}
	
	\medskip
	\noindent
	\textbf{(iii) Random-baseline regime.}
	Assume that, conditional on the population R-ranking, the population
	U-ranking is a uniformly random permutation of the $J$ units. Then
	the removal set $\mathcal R^{(l)}(m)$ is a uniformly random subset of
	size $m$, while the protection set $\mathcal P^{(l)}(m)$ is a fixed
	subset of size $m$. Therefore the overlap
	\begin{equation}
		\begin{aligned}
			H_l(m):= 
			\biggl|\mathcal R^{(l)}(m)\cap \mathcal P^{(l)}(m)\biggr|
		\end{aligned}
	\end{equation}
	has the Hypergeometric distribution with population size $J$,
	number of marked items $m$, and sample size $m$:
	\begin{equation}
		H_l(m)\sim \mathrm{Hypergeometric}(J,m,m).
	\end{equation}
	Its expectation is
	\begin{equation}
		\mathbb E[H_l(m)] = \frac{m\cdot m}{J}=\frac{m^2}{J}.
	\end{equation}
	Hence
	\begin{equation}
		\begin{aligned}
			\mathbb E[\mathrm{ToD}^{(l)}(m)] 
			&= 
			\frac{1}{m}\mathbb E[H_l(m)] 
			&= 
			\frac{m}{J}.
		\end{aligned}
	\end{equation}
	This proves the stated baseline behavior.
\end{proof}

\subsection{Single-Signal and Compatible Regimes}
\label{app:single_signal_regimes}
The additive loss model in Proposition~\ref{prop:compatible_tod_uniform} also
clarifies when a single signal may or may not be sufficient. If \(\gamma=0\), or
if the structural term is negligible in the local pruning region, the model
reduces to R-Score control. This is a useful degenerate case, but it is not
expected to describe aggressive structured pruning, where simultaneous deletions
usually interact. If all relevant R-Scores are nearly equal, the protection
signal carries little tail information and a U-only rule may be sufficient; this
is exactly the case where R-Score protection has little to add. If the U- and
R-rankings are fully compatible, ToD can reduce to a single ranking or to a
near-uniform profile. This may occur locally in some layers, but is unlikely to
hold across heterogeneous networks. The main regime targeted by FAIR-Pruner is
the complementary one: R-Scores contain a protected high-sensitivity tail,
U-Scores provide a smooth removal ordering, and the two rankings are compatible
with the marginal cost condition in Proposition~\ref{prop:compatible_tod_uniform}.

\subsection{Conflict-curve ordering}
\label{app:conflict_curve_ordering}
\begin{proposition}[Conflict-curve ordering and layer-wise pruning ratios]
	\label{prop:conflict_curve_ordering}
	For each layer $l\in L$, define the right-endpoint interpolated
	conflict curve
	\begin{equation}
		\widetilde C_l(0):=0,
		\qquad
		\widetilde C_l(x)
		:=
		\mathrm{ToD}^{(l)}(\lceil xJ^{(l)}\rceil),
		\quad x\in(0,1].
	\end{equation}
	If two layers $l_1,l_2\in L$ satisfy
	\begin{equation}
		\widetilde C_{l_1}(x)\le \widetilde C_{l_2}(x),
		\qquad \forall x\in[0,1],
	\end{equation}
	then for every $\alpha\in(0,1)$,
	\begin{equation}
		\frac{m^{*(l_1)}(\alpha)}{J^{(l_1)}}
		\ge
		\frac{m^{*(l_2)}(\alpha)}{J^{(l_2)}}.
	\end{equation}
	Hence, under a shared ToD level $\alpha$, the layer whose U- and
	R-rankings are more compatible receives a larger pruning ratio.
\end{proposition}
\begin{proof}
	Fix $\alpha\in(0,1)$. For each layer $l\in L$, define
	\begin{equation}
		\begin{aligned}
			A_l(\alpha) 
			&:= 
			\{x\in[0,1]:\widetilde C_l(x)\le \alpha\}.
		\end{aligned}
	\end{equation}
	
	We first show that
	\begin{equation}
		\sup A_l(\alpha)=\frac{m^{*(l)}(\alpha)}{J^{(l)}}.
	\end{equation}
	Write $J:=J^{(l)}$ and $m^*:=m^{*(l)}(\alpha)$. By definition,
	\begin{equation}
		m^*=\max\{m\in\{0,1,\dots,J\}:\mathrm{ToD}^{(l)}(m)\le \alpha\}.
	\end{equation}
	Recall that
	\begin{equation}
		\begin{aligned}
			\widetilde C_l(0)=0, \qquad 
			\widetilde C_l(x)=\mathrm{ToD}^{(l)}(\lceil xJ\rceil), 
			\quad x\in(0,1].
		\end{aligned}
	\end{equation}
	For any $x\in(0,1]$, let $m=\lceil xJ\rceil$. Then
	\begin{equation}
		\begin{aligned}
			x\in\Bigl(\frac{m-1}{J},\,\frac{m}{J}\Bigr] 
			\quad\Longleftrightarrow\quad 
			\lceil xJ\rceil=m.
		\end{aligned}
	\end{equation}
	Hence
	\begin{equation}
		A_l(\alpha)
		=
		\{0\}\cup
		\bigcup_{\substack{m\in\{1,\dots,J\}: \mathrm{ToD}^{(l)}(m)\le \alpha}}
		\Bigl(\frac{m-1}{J},\,\frac{m}{J}\Bigr].
	\end{equation}
	The feasible indices need not form an initial segment because the ToD curve
	is not assumed to be monotone in $m$. Nevertheless, by the definition of
	$m^*$ as the largest feasible index, the largest right endpoint in the above
	union is \(m^*/J\). Therefore,
	\begin{equation}
		\sup A_l(\alpha)=\frac{m^{*(l)}(\alpha)}{J^{(l)}}.
	\end{equation}
	
	Now suppose that, for two layers $l_1,l_2\in L$,
	\begin{equation}
		\begin{aligned}
			\widetilde C_{l_1}(x)\le \widetilde C_{l_2}(x), 
			\qquad \forall x\in[0,1].
		\end{aligned}
	\end{equation}
	Then for every $x\in A_{l_2}(\alpha)$, we have
	\begin{equation}
		\widetilde C_{l_2}(x)\le \alpha,
	\end{equation}
	and hence
	\begin{equation}
		\widetilde C_{l_1}(x)\le \widetilde C_{l_2}(x)\le \alpha.
	\end{equation}
	So $x\in A_{l_1}(\alpha)$. Therefore,
	\begin{equation}
		A_{l_2}(\alpha)\subseteq A_{l_1}(\alpha).
	\end{equation}
	Taking suprema on both sides gives
	\begin{equation}
		\sup A_{l_2}(\alpha)\le \sup A_{l_1}(\alpha).
	\end{equation}
	Using the identity proved above, we conclude that
	\begin{equation}
		\begin{aligned}
			\frac{m^{*(l_1)}(\alpha)}{J^{(l_1)}} 
			&= 
			\sup A_{l_1}(\alpha) 
			&\ge 
			\sup A_{l_2}(\alpha) 
			&= 
			\frac{m^{*(l_2)}(\alpha)}{J^{(l_2)}}.
		\end{aligned}
	\end{equation}
	This proves the result.
\end{proof}

\section{Proofs of the main text results}
\label{app:proofs}

This appendix gives detailed proofs of the main theoretical results:
Proposition~\ref{prop:stepwise_budget_control}, Proposition~\ref{prop:alpha_monotone},
Theorem~\ref{thm:selection_consistency}, Proposition~\ref{prop:no_free_lunch_rank_only},
Proposition~\ref{prop:structural_proxy_sufficient},
Theorem~\ref{thm:finite_deletion_envelope},
Theorem~\ref{thm:absolute_variation_bound}, Proposition~\ref{prop:tod_sensitivity_mass},
Corollary~\ref{cor:tod_loss_envelope},
and Proposition~\ref{prop:compatible_tod_uniform}.
In the proofs only, when \(u=(l,j)\), we write
\(R_u:=R_j^{(l)}=|\Delta\mathcal L(\{u\})|\) for brevity.
\subsection{Proof of Proposition~\ref{prop:stepwise_budget_control}}

\begin{proof}
	Fix the empirical scores. For each layer $l$, the empirical conflict values
	\begin{equation}
		\widehat{\mathrm{ToD}}^{(l)}(m),\qquad m=0,1,\ldots,J^{(l)},
	\end{equation}
	are fixed finite numbers. By definition,
	\begin{equation}
		\widehat m^{(l)}(\alpha)
		=
		\max\left\{m:\widehat{\mathrm{ToD}}^{(l)}(m)\le\alpha\right\}.
	\end{equation}
	If $0<\alpha_1\le\alpha_2<1$, then
	\begin{equation}
		\{m:\widehat{\mathrm{ToD}}^{(l)}(m)\le\alpha_1\}
		\subseteq
		\{m:\widehat{\mathrm{ToD}}^{(l)}(m)\le\alpha_2\}.
	\end{equation}
	Taking maxima gives
	\begin{equation}
		\widehat m^{(l)}(\alpha_1)\le \widehat m^{(l)}(\alpha_2).
	\end{equation}
	Therefore
	\begin{equation}
		\widehat B(\alpha)=\sum_{l\in L}c_l\widehat m^{(l)}(\alpha)
	\end{equation}
	is nondecreasing because each $c_l>0$.
	
	Moreover, each $\widehat m^{(l)}(\alpha)$ takes values in the finite set
	$\{0,1,\ldots,J^{(l)}\}$ and can change only when $\alpha$ reaches one of the
	finite values $\widehat{\mathrm{ToD}}^{(l)}(m)$. Hence $\widehat m^{(l)}(\alpha)$
	is piecewise constant, and so is $\widehat B(\alpha)$. All possible changes of
	$\widehat B(\alpha)$ are contained in
	\begin{equation}
		\widehat{\mathcal A}
		=
		\{0,1\}\cup
		\{\widehat{\mathrm{ToD}}^{(l)}(m):l\in L,\ 1\le m\le J^{(l)}\}.
	\end{equation}
	Thus a budget-matched ToD model can be obtained by evaluating the finite set
	$\widehat{\mathcal A}$, without recomputing scores or retraining the network.
\end{proof}

\subsection{Proof of Proposition~\ref{prop:alpha_monotone}}
\label{app:proof_prop2}

\begin{proof}[Proof of Proposition~\ref{prop:alpha_monotone}]
	Fix a layer $l\in L$ and let $0<\alpha_1\le \alpha_2<1$. Then
	\begin{equation}
		\begin{aligned}
			&\biggl\{ 
			m\in\{0,1,\dots,J^{(l)}\}: 
			\widehat{\mathrm{ToD}}^{(l)}(m)\le \alpha_1 
			\biggr\} \\
			\subseteq &
			\biggl\{ 
			m\in\{0,1,\dots,J^{(l)}\}: 
			\widehat{\mathrm{ToD}}^{(l)}(m)\le \alpha_2 
			\biggr\}.
		\end{aligned}
	\end{equation}
	Taking maxima on both sides gives
	\begin{equation}
		\widehat m^{(l)}(\alpha_1)\le \widehat m^{(l)}(\alpha_2).
	\end{equation}
	Hence $\widehat m^{(l)}(\alpha)$ is nondecreasing in $\alpha$.
	The same monotonicity then holds for the overall pruning ratio
	obtained by aggregating the layer-wise pruning depths.
\end{proof}


\subsection{Preparatory Lemmas for Theorem~\ref{thm:selection_consistency}}
\label{app:lemmas_thm1}

For the following lemmas, fix one layer $l\in L$ and suppress the
superscript $(l)$ whenever no confusion arises. Since $J^{(l)}<\infty$,
all rankings and set comparisons below are over a finite index set.

\begin{lemma}[Rank recovery]
	\label{lem:rank_recovery}
	Under Assumptions~\ref{ass:u_score_consistency}--\ref{ass:rank_separation}, the empirical U-ranking
	eventually coincides almost surely with the population U-ranking,
	and the empirical R-ranking eventually coincides almost surely with
	the population R-ranking.
\end{lemma}

\begin{proof}
	By Assumption~\ref{ass:u_score_consistency},
	\begin{equation}
		\begin{aligned}
			\max_{j\in[J^{(l)}]} |\widehat U_j-U_j| 
			\overset{a.s.}{\longrightarrow}0.
		\end{aligned}
	\end{equation}
	By Assumption~\ref{ass:rank_separation},
	\begin{equation}
		\Delta_U:=\min_{i\neq j}|U_i-U_j|>0.
	\end{equation}
	Hence, on a probability-one event, for all sufficiently large $n$,
	\begin{equation}
		\max_j |\widehat U_j-U_j|<\Delta_U/4.
	\end{equation}
	
	Fix any pair $i\neq j$ with $U_i<U_j$. Then
	\begin{equation}
		\begin{aligned}
			\widehat U_i-\widehat U_j 
			&= 
			(\widehat U_i-U_i)+(U_i-U_j)+(U_j-\widehat U_j) \\
			&\le 
			|\widehat U_i-U_i|-(U_j-U_i)+|\widehat U_j-U_j|.
		\end{aligned}
	\end{equation}
	Since $U_j-U_i\ge \Delta_U$, we obtain
	\begin{equation}
		\begin{aligned}
			\widehat U_i-\widehat U_j 
			&< 
			\Delta_U/4-\Delta_U+\Delta_U/4 \\
			&= 
			-\Delta_U/2<0.
		\end{aligned}
	\end{equation}
	Thus $\widehat U_i<\widehat U_j$ eventually almost surely for every
	$i\neq j$, and the empirical U-ranking eventually coincides with the
	population U-ranking.
	
	Exactly the same argument applies to the R-Scores using
	Assumption~\ref{ass:protection_consistency} and
	\begin{equation}
		\Delta_R:=\min_{i\neq j}|R_i-R_j|>0.
	\end{equation}
	This proves the claim.
\end{proof}

\begin{lemma}[Set recovery]
	\label{lem:set_recovery}
	Under the assumptions of Lemma~\ref{lem:rank_recovery}, for every
	fixed $m\in\{0,1,\dots,J^{(l)}\}$,
	\begin{equation}
		\begin{aligned}
			&\widehat{\mathcal R}^{(l)}(m)=\mathcal R^{(l)}(m), \qquad \\
			&\widehat{\mathcal P}^{(l)}(m)=\mathcal P^{(l)}(m),
		\end{aligned}
	\end{equation}
	eventually almost surely. Consequently,
	\begin{equation}
		\widehat{\mathrm{ToD}}^{(l)}(m)=\mathrm{ToD}^{(l)}(m)
	\end{equation}
	eventually almost surely for every fixed $m$.
\end{lemma}

\begin{proof}
	Once the empirical U-ranking coincides with the population U-ranking,
	the set of the $m$ smallest empirical U-Scores is exactly the set of
	the $m$ smallest population U-Scores. Likewise, once the empirical
	R-ranking coincides with the population R-ranking, the set of the $m$
	largest empirical R-Scores is exactly the set of the $m$ largest
	population R-Scores. Assumption~\ref{ass:rank_separation} excludes ties, so both sets are
	uniquely defined. Therefore,
	\begin{equation}
		\begin{aligned}
			&\widehat{\mathcal R}^{(l)}(m)=\mathcal R^{(l)}(m), \qquad \\
			&\widehat{\mathcal P}^{(l)}(m)=\mathcal P^{(l)}(m),
		\end{aligned}
	\end{equation}
	eventually almost surely. The equality of the ToD values follows
	immediately from the equality of the sets.
\end{proof}

\begin{lemma}[Recovery of the pruning depth]
	\label{lem:depth_recovery}
	For any fixed $\alpha\in(0,1)$,
	\begin{equation}
		\widehat m^{(l)}(\alpha)=m^{*(l)}(\alpha)
	\end{equation}
	eventually almost surely.
\end{lemma}

\begin{proof}
	By Lemma~\ref{lem:set_recovery}, for every fixed
	$m\in\{0,1,\dots,J^{(l)}\}$,
	\begin{equation}
		\widehat{\mathrm{ToD}}^{(l)}(m)=\mathrm{ToD}^{(l)}(m)
	\end{equation}
	eventually almost surely. Since $\{0,1,\dots,J^{(l)}\}$ is finite,
	there exists a probability-one event on which all these equalities
	hold simultaneously for all sufficiently large $n$. Hence,
	eventually almost surely,
	\begin{equation}
		\begin{aligned}
			\biggl\{ m:\widehat{\mathrm{ToD}}^{(l)}(m)\le \alpha \biggr\}
			= 
			\biggl\{ m:\mathrm{ToD}^{(l)}(m)\le \alpha \biggr\}.
		\end{aligned}
	\end{equation}
	Taking maxima yields
	\begin{equation}
		\widehat m^{(l)}(\alpha)=m^{*(l)}(\alpha)
	\end{equation}
	eventually almost surely.
\end{proof}


\subsection{Proof of Theorem~\ref{thm:selection_consistency}}
\label{app:proof_thm1}

\begin{proof}[Proof of Theorem~\ref{thm:selection_consistency}]
	Fix a layer $l\in L$. By
	Lemmas~\ref{lem:rank_recovery}--\ref{lem:depth_recovery},
	\begin{equation}
		\begin{aligned}
			\widehat{\mathcal R}^{(l)}(\widehat m^{(l)}(\alpha)) 
			= 
			\mathcal R^{(l)}(m^{*(l)}(\alpha))
		\end{aligned}
	\end{equation}
	eventually almost surely. Define the layer-wise error event
	\begin{equation}
		\begin{aligned}
			E_n^{(l)}(\alpha) 
			:= 
			\biggl\{ 
			\widehat{\mathcal R}^{(l)}(\widehat m^{(l)}(\alpha)) 
			\neq 
			\mathcal R^{(l)}(m^{*(l)}(\alpha)) 
			\biggr\}.
		\end{aligned}
	\end{equation}
	Then $\mathbf 1_{E_n^{(l)}(\alpha)}\to 0$ almost surely. Since
	\begin{equation}
		0\le \mathbf 1_{E_n^{(l)}(\alpha)}\le 1,
	\end{equation}
	the dominated convergence theorem gives
	\begin{equation}
		\begin{aligned}
			\Pr\!\big(E_n^{(l)}(\alpha)\big) 
			= 
			\mathbb E\!\biggl[\mathbf 1_{E_n^{(l)}(\alpha)}\biggr] 
			\to 0.
		\end{aligned}
	\end{equation}
	This proves the layer-wise statement.
	
	If $|L|<\infty$, define the overall error event
	\begin{equation}
		\begin{aligned}
			E_n(\alpha) 
			:= 
			\biggl\{ 
			\exists\, l\in L: 
			\widehat{\mathcal R}^{(l)}(\widehat m^{(l)}(\alpha)) 
			\neq 
			\mathcal R^{(l)}(m^{*(l)}(\alpha)) 
			\biggr\}.
		\end{aligned}
	\end{equation}
	By the union bound,
	\begin{equation}
		\begin{aligned}
			\Pr\!\big(E_n(\alpha)\big) 
			&\le 
			\sum_{l\in L}\Pr\!\big(E_n^{(l)}(\alpha)\big)\to 0.
		\end{aligned}
	\end{equation}
	This proves the result.
\end{proof}


\subsection{Proof of Proposition~\ref{prop:no_free_lunch_rank_only}}
\label{app:no_free_lunch_rank_only}

\begin{proof}
	Let
	\begin{equation}
		L_+:=\{l\in L:m_l>\bar m_l\},
		\qquad
		L_-:=\{l\in L:m_l<\bar m_l\}.
	\end{equation}
	Because \(m\ne\bar m\) and \(\sum_l m_l=\sum_l\bar m_l\), both sets are
	nonempty. For each layer \(l\), choose any positive baseline scores
	\(q_j^{(l)}\) whose ordering is consistent with the prescribed within-layer
	R-ranking. We construct R-Scores of the form
	\begin{equation}
		R_j^{(l)}=a_lq_j^{(l)},\qquad a_l>0.
	\end{equation}
	Layer-wise multiplication by \(a_l\) preserves every within-layer R-ranking,
	and it does not affect the within-layer U-ranking used to form
	\(\mathcal R^{(l)}(m)\).
	
	Since the removal sets are prefixes of the same within-layer U-ranking, for
	\(l\in L_+\) we have
	\begin{equation}
		\mathcal R^{(l)}(\bar m_l)\subsetneq \mathcal R^{(l)}(m_l),
	\end{equation}
	whereas for \(l\in L_-\),
	\begin{equation}
		\mathcal R^{(l)}(m_l)\subsetneq \mathcal R^{(l)}(\bar m_l).
	\end{equation}
	Set
	\begin{equation}
		\begin{aligned}
			d_l^+
			&:=
			\sum_{j\in\mathcal R^{(l)}(m_l)\setminus\mathcal R^{(l)}(\bar m_l)}
			q_j^{(l)},\qquad l\in L_+,\\
			d_l^-
			&:=
			\sum_{j\in\mathcal R^{(l)}(\bar m_l)\setminus\mathcal R^{(l)}(m_l)}
			q_j^{(l)},\qquad l\in L_-.
		\end{aligned}
	\end{equation}
	Therefore
	\begin{equation}
		\begin{aligned}
			&\Delta_{\mathrm{add}}(S(m))-\Delta_{\mathrm{add}}(S(\bar m))\\
			&\quad =
			\sum_{l\in L_+}a_ld_l^+
			-
			\sum_{l\in L_-}a_ld_l^-.
		\end{aligned}
	\end{equation}
	The quantities \(d_l^+\) and \(d_l^-\) are strictly positive. Taking
	\(a_l=M\) for \(l\in L_+\), \(a_l=\varepsilon\) for \(l\in L_-\), and any
	fixed positive value for the remaining layers, the displayed difference is
	positive for sufficiently large \(M\) and sufficiently small \(\varepsilon\).
	Thus the claimed domination can be reversed while all within-layer rankings
	are kept fixed.
\end{proof}

\subsection{Proof of Theorem~\ref{thm:finite_deletion_envelope}}
\label{app:finite_deletion_envelope}

\begin{proposition}[Smooth interpolation implies bounded deletion interaction]
	\label{prop:smooth_implies_deletion_interaction}
	Fix a finite pruning set \(S\subseteq\mathcal U\). Suppose there exists an
	auxiliary masked interpolation \(z\mapsto\hat f_{z;S}\),
	\(z\in[0,1]^{|S|}\), such that
	\begin{equation}
		\hat f_{\mathbf 1;S}=\hat f,
		\qquad
		\hat f_{\mathbf 1-h_A;S}=\hat f_{-A},
		\qquad A\subseteq S,
	\end{equation}
	where \(h_A\) is the indicator vector of \(A\) in the coordinate system of
	\(S\). Let
	\begin{equation}
		\Phi_S(z):=\mathbb E[\mathcal L(\hat f_{z;S}(X),Y)].
	\end{equation}
	If \(\Phi_S\) is twice differentiable and, for all distinct \(u,v\in S\),
	\begin{equation}
		\sup_{z\in[0,1]^{|S|}}
		\left|\partial_{z_u z_v}^2\Phi_S(z)\right|
		\le
		\lambda b_ub_v,
	\end{equation}
	then the bounded finite-deletion interaction condition in
	Theorem~\ref{thm:finite_deletion_envelope} holds.
\end{proposition}

\begin{proof}
	Fix distinct \(u,v\in S\) and \(T\subseteq S\setminus\{u,v\}\). Define
	\begin{equation}
		\psi(s,t):=\Phi_S(\mathbf 1-h_T-se_u-te_v),
		\qquad (s,t)\in[0,1]^2.
	\end{equation}
	The vertex identities imply that
	\begin{equation}
		\begin{aligned}
			&\Delta\mathcal L(T\cup\{u,v\})
			-\Delta\mathcal L(T\cup\{u\})\\
			&\qquad
			-\Delta\mathcal L(T\cup\{v\})
			+\Delta\mathcal L(T)\\
			=&\,
			\psi(1,1)-\psi(1,0)\\
			&-\psi(0,1)+\psi(0,0).
		\end{aligned}
	\end{equation}
	By the two-dimensional fundamental theorem of calculus,
	\begin{equation}
		\begin{aligned}
			&\psi(1,1)-\psi(1,0)-\psi(0,1)+\psi(0,0)\\
			&\qquad =
			\int_0^1\!\!\int_0^1 \partial_{st}^2\psi(s,t)\,ds\,dt.
		\end{aligned}
	\end{equation}
	Since \(\partial_{st}^2\psi(s,t)=
	\partial_{z_u z_v}^2\Phi_S(\mathbf 1-h_T-se_u-te_v)\), the mixed-derivative
	bound gives the desired finite-deletion interaction bound.
\end{proof}

\begin{proof}
	Let \(S=\{u_1,\ldots,u_s\}\) be any ordering of the finite pruning set, and set
	\(S_0=\varnothing\) and \(S_t=\{u_1,\ldots,u_t\}\). Define
	\begin{equation}
		\Gamma(S):=\Delta\mathcal L(S)-\sum_{u\in S}\Delta\mathcal L(\{u\}).
	\end{equation}
	For \(t=1,\ldots,s\), write
	\begin{equation}
		M_t
		:=
		\Delta\mathcal L(S_t)
		-
		\Delta\mathcal L(S_{t-1})
	\end{equation}
	for the marginal risk change of deleting \(u_t\) after \(S_{t-1}\) has already
	been deleted, and
	\begin{equation}
		D_t
		:=
		\Delta\mathcal L(\{u_t\})
	\end{equation}
	for the single-unit deletion change. Then
	\begin{equation}
		\Gamma(S)=\sum_{t=1}^s(M_t-D_t).
	\end{equation}
	Fix an arbitrary ordering \(S_{t-1}=\{v_1,\ldots,v_{t-1}\}\) and set
	\(T_q=\{v_1,\ldots,v_q\}\). Then
	\begin{equation}
		\begin{aligned}
			M_t-D_t
			=&\,
			\bigl[\Delta\mathcal L(T_{t-1}\cup\{u_t\})
			-\Delta\mathcal L(T_{t-1})\bigr]\\
			&-
			\bigl[\Delta\mathcal L(\{u_t\})-\Delta\mathcal L(\varnothing)\bigr]\\
			=&
			\sum_{q=1}^{t-1}
			\Bigl[
			\Delta\mathcal L(T_q\cup\{u_t\})
			-\Delta\mathcal L(T_q)\\
			&\hspace{4em}
			-\Delta\mathcal L(T_{q-1}\cup\{u_t\})
			+\Delta\mathcal L(T_{q-1})
			\Bigr].
		\end{aligned}
	\end{equation}
	Each summand is a second-order finite deletion difference for the pair
	\(\{u_t,v_q\}\) under background \(T_{q-1}\). The theorem condition therefore
	gives
	\begin{equation}
		|M_t-D_t|
		\le
		\lambda b_{u_t}\sum_{v\in S_{t-1}}b_v.
	\end{equation}
	Summing over \(t\) gives
	\begin{equation}
		|\Gamma(S)|
		\le
		\lambda\sum_{t=1}^s b_{u_t}\sum_{v\in S_{t-1}}b_v
		=
		\lambda\sum_{\{u,v\}\subseteq S}b_ub_v.
	\end{equation}
	Finally,
	\begin{equation}
		|\Gamma(S)|
		\le
		\lambda\sum_{\{u,v\}\subseteq S}b_ub_v
		\le
		\frac{\lambda}{2}\left(\sum_{u\in S}b_u\right)^2.
	\end{equation}
	This proves the interaction bound. Also,
	\begin{equation}
		\Delta\mathcal L(S)
		=
		\sum_{u\in S}\Delta\mathcal L(\{u\})+\Gamma(S).
	\end{equation}
	Taking absolute values and applying the triangle inequality gives
	\begin{equation}
		\begin{aligned}
			|\Delta\mathcal L(S)|
			&\le
			\sum_{u\in S}|\Delta\mathcal L(\{u\})|+|\Gamma(S)|\\
			&=
			\sum_{u\in S}R_u+|\Gamma(S)|.
		\end{aligned}
	\end{equation}
	Combining this with the interaction bound yields
	\begin{equation}
		|\Delta\mathcal L(S)|
		\le
		\sum_{u\in S}R_u
		+
		\frac{\lambda}{2}
		\left(\sum_{u\in S}b_u\right)^2.
	\end{equation}
\end{proof}


\begin{theorem}[M\"obius decomposition of simultaneous pruning loss]
	\label{thm:absolute_variation_bound}
	For every unordered pair of distinct units \(\{u,v\}\subseteq\mathcal U\), define
	\begin{equation}
		I_{u,v}
		:=
		\Delta\mathcal L(\{u,v\})
		-
		\Delta\mathcal L(\{u\})
		-
		\Delta\mathcal L(\{v\}).
	\end{equation}
	For every finite \(T\subseteq\mathcal U\) with \(|T|\ge 3\), define
	\begin{equation}
		\kappa(T):=\sum_{B\subseteq T}(-1)^{|T|-|B|}\Delta\mathcal L(B).
	\end{equation}
	Then, for every finite \(S\subseteq\mathcal U\),
	\begin{equation}
		\Delta\mathcal L(S)
		=\sum_{u\in S}\Delta\mathcal L(\{u\})
		+\sum_{\{u,v\}\subseteq S}I_{u,v}
		+\sum_{\substack{T\subseteq S\\ |T|\ge 3}}\kappa(T),
	\end{equation}
	and hence
	\begin{equation}
		|\Delta\mathcal L(S)|
		\le \sum_{u\in S} R_u
		+\sum_{\{u,v\}\subseteq S}|I_{u,v}|
		+\sum_{\substack{T\subseteq S\\ |T|\ge 3}}|\kappa(T)|.
	\end{equation}
\end{theorem}

\subsection{Proof of Theorem~\ref{thm:absolute_variation_bound}}
\label{app:proof_thm2}
\label{app:mobius_decomposition}

\begin{proof}[Proof of Theorem~\ref{thm:absolute_variation_bound}]
	Define the set function
	\begin{equation}
		\begin{aligned}
			F(S):=\Delta \mathcal L(S), 
			\qquad S\subseteq \mathcal U,
		\end{aligned}
	\end{equation}
	with the convention $F(\varnothing)=0$.
	
	For every nonempty finite set $T\subseteq \mathcal U$, define
	\begin{equation}
		\begin{aligned}
			\mu(T):= 
			\sum_{B\subseteq T}(-1)^{|T|-|B|}F(B).
		\end{aligned}
	\end{equation}
	This is the M\"obius transform of the set function $F$ on the Boolean
	lattice of subsets of $\mathcal U$.
	
	We first show that, for every finite pruning set $S\subseteq\mathcal U$,
	\begin{equation}
		F(S)=\sum_{\varnothing\neq T\subseteq S}\mu(T).
	\end{equation}
	Indeed, substituting the definition of $\mu(T)$ gives
	\begin{equation}
		\begin{aligned}
			\sum_{\varnothing\neq T\subseteq S}\mu(T) 
			= 
			\sum_{\varnothing\neq T\subseteq S} 
			\sum_{B\subseteq T}(-1)^{|T|-|B|}F(B).
		\end{aligned}
	\end{equation}
	Because all sums are finite, we may interchange the order of
	summation:
	\begin{equation}
		\begin{aligned}
			\sum_{\varnothing\neq T\subseteq S}\mu(T) 
			&= 
			\sum_{B\subseteq S} F(B) 
			\sum_{\substack{T:\,B\subseteq T\subseteq S\\ T\neq \varnothing}} 
			(-1)^{|T|-|B|}.
		\end{aligned}
	\end{equation}
	
	Now fix $B\subseteq S$.
	If $B=S$, then the inner sum contains only the term $T=S$, hence it
	equals $1$.
	If $\varnothing\ne B\subsetneq S$, write $T=B\cup C$ with
	$C\subseteq S\setminus B$. Then
	\begin{equation}
		\begin{aligned}
			\sum_{T:\,B\subseteq T\subseteq S} 
			(-1)^{|T|-|B|} 
			&= 
			\sum_{C\subseteq S\setminus B}(-1)^{|C|} \\
			&= 
			(1-1)^{|S\setminus B|} \\
			&= 0.
		\end{aligned}
	\end{equation}
	The remaining case $B=\varnothing$ contributes nothing because
	$F(\varnothing)=0$.
	Therefore, every coefficient vanishes except the one corresponding to
	$B=S$, and we obtain
	\begin{equation}
		\sum_{\varnothing\neq T\subseteq S}\mu(T)=F(S)=\Delta\mathcal L(S).
	\end{equation}
	
	We next identify the terms of orders one and two.
	For a singleton $\{u\}$,
	\begin{equation}
		\begin{aligned}
			\mu(\{u\}) 
			&= 
			\sum_{B\subseteq \{u\}} (-1)^{1-|B|}F(B) \\
			&= 
			F(\{u\})-F(\varnothing) 
			= 
			\Delta\mathcal L(\{u\}).
		\end{aligned}
	\end{equation}
	For a pair of distinct units $\{u,v\}$,
	\begin{equation}
		\begin{aligned}
			\mu(\{u,v\}) 
			&= 
			\sum_{B\subseteq \{u,v\}} (-1)^{2-|B|}F(B) \\
			&= 
			F(\{u,v\})-F(\{u\})-F(\{v\})+F(\varnothing),
		\end{aligned}
	\end{equation}
	that is,
	\begin{equation}
		\begin{aligned}
			\mu(\{u,v\}) 
			&= 
			\Delta \mathcal L(\{u,v\})-\Delta \mathcal L(\{u\})-\Delta \mathcal L(\{v\}) \\
			&= 
			I_{u,v}.
		\end{aligned}
	\end{equation}
	For every set $T$ with $|T|\ge 3$, the coefficient $\mu(T)$ is
	exactly the quantity $\kappa(T)$ defined in the theorem statement.
	
	Substituting these identities into the M\"obius inversion formula
	yields
	\begin{equation}
		\begin{aligned}
			\Delta \mathcal L(S) 
			&= 
			\sum_{u\in S}\Delta \mathcal L(\{u\}) + 
			\sum_{\{u,v\}\subseteq S} I_{u,v} + 
			\sum_{\substack{T\subseteq S\\ |T|\ge 3}} \kappa(T),
		\end{aligned}
	\end{equation}
	which proves the exact decomposition.
	
	Finally, taking absolute values on both sides and applying the
	triangle inequality, we obtain
	\begin{equation}
		\begin{aligned}
			|\Delta \mathcal L(S)| 
			&\le 
			\sum_{u\in S} |\Delta \mathcal L(\{u\})| + 
			\sum_{\{u,v\}\subseteq S} |I_{u,v}| + 
			\sum_{\substack{T\subseteq S\\ |T|\ge 3}} |\kappa(T)|.
		\end{aligned}
	\end{equation}
	By the definition
	\begin{equation}
		\begin{aligned}
			R_u:=|\Delta \mathcal L(\{u\})|, \qquad 
		\end{aligned}
	\end{equation}
	this is exactly
	\begin{equation}
		\begin{aligned}
			|\Delta \mathcal L(S)| 
			&\le 
			\sum_{u\in S} R_u + 
			\sum_{\{u,v\}\subseteq S} |I_{u,v}| + 
			\sum_{\substack{T\subseteq S\\ |T|\ge 3}} |\kappa(T)|.
		\end{aligned}
	\end{equation}
	This completes the proof.
\end{proof}


\subsection{Proof of Lemma~\ref{lem:tod_feasibility}}
\label{app:tod_feasibility}

\begin{proof}
	Let \(J:=J^{(l)}\) and \(P:=\mathcal P^{(l)}(m)\), so that \(|P|=m\). For any
	set \(S\subseteq[J]\) with \(|S|=m\),
	\begin{equation}
		|S\cap P|
		=
		|S|+|P|-|S\cup P|
		\ge 2m-J.
	\end{equation}
	Since an overlap is nonnegative, every size-\(m\) set satisfies
	\begin{equation}
		|S\cap P|\ge \max\{0,2m-J\}.
	\end{equation}
	This lower bound is attainable by choosing as many elements of \(S\) as possible
	outside \(P\). Hence the minimum possible overlap between a size-\(m\) removal
	set and the protected set is \(\max\{0,2m-J\}\). Therefore a set satisfying
	\(|S\cap P|\le\lfloor\alpha m\rfloor\) exists if and only if
	\begin{equation}
		\max\{0,2m-J\}\le\lfloor\alpha m\rfloor .
	\end{equation}
	Ignoring integer rounding gives the equivalent scale condition
	\(m\le J/(2-\alpha)\).
\end{proof}

\subsection{Proof of Proposition~\ref{prop:tod_sensitivity_mass}}
\label{app:tod_sensitivity_mass}

\begin{proof}
	If $m=0$, then \(\mathcal R^{(l)}(0)=\varnothing\),
	\(\mathcal C_l(0,\alpha)=\{\varnothing\}\), and
	\(\mathcal A_l(0,\alpha)=0\), so both the bound and the sharpness statement
	are immediate.
	Now suppose $m\ge1$. Let
	\begin{equation}
		\begin{aligned}
			H_l(m)&:=\mathcal R^{(l)}(m)\cap\mathcal P^{(l)}(m),\\
			h&:=|H_l(m)|,\qquad c:=\lfloor\alpha m\rfloor.
		\end{aligned}
	\end{equation}
	The condition $\mathrm{ToD}^{(l)}(m)\le\alpha$ gives $h/m\le\alpha$. Since
	$h$ is an integer,
	\begin{equation}
		h\le \lfloor\alpha m\rfloor=c.
	\end{equation}
	The units in $H_l(m)$ belong to the protected top-$m$ R-Score tail. Therefore,
	\begin{equation}
		\sum_{j\in H_l(m)}R_j^{(l)}
		\le
		\sum_{r=1}^h R_{[r]}^{(l)}.
	\end{equation}
	Every unit in $\mathcal R^{(l)}(m)\setminus\mathcal P^{(l)}(m)$ lies outside
	the protected top-$m$ tail. Among all such outside units, the largest
	\(m-h\) R-Scores are
	\(R_{[m+1]}^{(l)},\ldots,R_{[2m-h]}^{(l)}\). Since this set
	has cardinality $m-h$,
	\begin{equation}
		\sum_{j\in\mathcal R^{(l)}(m)\setminus\mathcal P^{(l)}(m)}R_j^{(l)}
		\le
		\sum_{r=m+1}^{2m-h}R_{[r]}^{(l)}.
	\end{equation}
	Combining the two parts,
	\begin{equation}
		\sum_{j\in\mathcal R^{(l)}(m)}R_j^{(l)}
		\le
		\sum_{r=1}^h R_{[r]}^{(l)}
		+
		\sum_{r=m+1}^{2m-h}R_{[r]}^{(l)}.
	\end{equation}
	For \(h<c\), increasing \(h\) by one replaces the outside-tail term
	\(R_{[2m-h]}^{(l)}\) by the protected-tail term
	\(R_{[h+1]}^{(l)}\). Since
	\(h+1\le m<2m-h\), the sorted order gives
	\begin{equation}
		R_{[h+1]}^{(l)}
		\ge
		R_{[2m-h]}^{(l)}.
	\end{equation}
	Thus the right-hand side is nondecreasing in \(h\) on \(0\le h\le c\), and is
	maximized at \(h=c\). Therefore
	\begin{equation}
		\sum_{j\in\mathcal R^{(l)}(m)}R_j^{(l)}
		\le
		\sum_{r=1}^{c}R_{[r]}^{(l)}
		+
		\sum_{r=m+1}^{2m-c}R_{[r]}^{(l)}
		=
		\mathcal A_l(m,\alpha).
	\end{equation}
	
	It remains to prove sharpness over the nonempty feasible class
	\(\mathcal C_l(m,\alpha)\). By Lemma~\ref{lem:tod_feasibility}, this class is
	nonempty exactly when
	\(\max\{0,2m-J^{(l)}\}\le\lfloor\alpha m\rfloor\). When this condition
	fails, the depth-\(m\) ToD-constrained problem is infeasible and no
	sharpness claim is asserted. Otherwise, consider the set consisting of the
	\(c\) units with the largest R-Scores in the
	protected top-\(m\) tail together with the \(m-c\) units with the largest
	R-Scores outside that protected tail. Its overlap with
	\(\mathcal P^{(l)}(m)\) is exactly \(c\), so its ToD value is
	\(c/m\le\alpha\), and its total R-Score mass is exactly
	\(\mathcal A_l(m,\alpha)\). Hence no smaller bound that depends
	only on the sorted R-Scores and the ToD constraint can hold
	uniformly over all admissible size-\(m\) removal sets. This proves the
	proposition.
\end{proof}

\begin{remark}[Interpretation of the sharp R-Score budget]
	The sharpness statement in Proposition~\ref{prop:tod_sensitivity_mass} is a
	rank-based one over the nonempty feasible class. For example, if
	\(R_{[1]}^{(l)}=\cdots=R_{[m]}^{(l)}=1\) and all remaining R-Scores are zero,
	the unconstrained depth-\(m\) worst-case R-Score mass is \(m\), whereas the
	ToD-constrained sharp budget is \(\lfloor\alpha m\rfloor\). If the R-Scores
	are nearly flat, no rank-based protection rule can produce a large worst-case
	gain, which is exactly the regime in which the protection signal carries
	little high-sensitivity-tail information.
\end{remark}

\subsection{Proof of Corollary~\ref{cor:tod_loss_envelope}}
\label{app:tod_loss_control}

\begin{proof}
	Let
	\begin{equation}
		m_l:=m^{*(l)}(\alpha),
		\qquad
		S_\alpha=
		\bigcup_{l\in L}\{(l,j):j\in\mathcal R^{(l)}(m_l)\}.
	\end{equation}
	By Theorem~\ref{thm:finite_deletion_envelope},
	\begin{equation}
		|\Delta\mathcal L(S_\alpha)|
		\le
		\sum_{u\in S_\alpha}R_u
		+
		\frac{\lambda}{2}
		\left(\sum_{u\in S_\alpha}b_u\right)^2.
	\end{equation}
	For the R-Score term, Proposition~\ref{prop:tod_sensitivity_mass}
	gives
	\begin{equation}
		\sum_{u\in S_\alpha}R_u
		=
		\sum_{l\in L}\sum_{j\in\mathcal R^{(l)}(m_l)}R_j^{(l)}
		\le
		\sum_{l\in L}\mathcal A_l(m_l,\alpha)
		=
		A_\alpha.
	\end{equation}
	For the structural term, the compatibility condition
	\eqref{eq:structural_compatibility} gives
	\begin{equation}
		\sum_{u\in S_\alpha}b_u
		=
		\sum_{l\in L}\sum_{j\in\mathcal R^{(l)}(m_l)}b_{(l,j)}
		\le
		\sum_{l\in L}B_l^U(m_l)
		=
		B_\alpha^U.
	\end{equation}
	Combining these two estimates yields
	\begin{equation}
		|\Delta\mathcal L(S_\alpha)|
		\le
		A_\alpha+\frac{\lambda}{2}\left(B_\alpha^U\right)^2.
	\end{equation}
	This proves the corollary.
\end{proof}

\subsection{Proof of Proposition~\ref{prop:compatible_tod_uniform}}
\label{app:compatible_tod_uniform}

\begin{proof}
	Let \(m\) be any feasible profile with \(\sum_l m_l=M\). If
	\(m=m^{\mathrm{tod}}\), there is nothing to prove. Otherwise, since the two
	profiles have the same total budget, there exist layers \(a,b\) such that
	\begin{equation}
		m_a<m_a^{\mathrm{tod}},
		\qquad
		m_b>m_b^{\mathrm{tod}}.
	\end{equation}
	Move one pruning unit from layer \(b\) to layer \(a\), and call the resulting
	profile \(m'\). The change in cost is
	\begin{equation}
		C(m')-C(m)=c_{a,m_a+1}-c_{b,m_b}.
	\end{equation}
	Because the marginal costs are nondecreasing along each U-ranked prefix,
	\begin{equation}
		c_{a,m_a+1}
		\le
		c_{a,m_a^{\mathrm{tod}}},
		\qquad
		c_{b,m_b}
		\ge
		c_{b,m_b^{\mathrm{tod}}+1}.
	\end{equation}
	The compatibility condition gives
	\begin{equation}
		c_{a,m_a^{\mathrm{tod}}}
		\le
		c_{b,m_b^{\mathrm{tod}}+1},
	\end{equation}
	and hence \(C(m')\le C(m)\). Repeating this exchange finitely many times
	transforms \(m\) into \(m^{\mathrm{tod}}\) without increasing \(C\). Therefore
	\(C(m^{\mathrm{tod}})\le C(m)\) for every feasible \(m\). If all compatibility
	inequalities are strict, each nontrivial exchange is strict, so every
	\(m\ne m^{\mathrm{tod}}\) has strictly larger cost.
\end{proof}

\section{Sensitivity to the Order of the Taylor Approximation in the R-Score}
\label{app:taylor_order}

In the main method, the Reconstruction Score (R-Score) is implemented using a first-order Taylor approximation of the loss change induced by removing a unit. A natural question is whether a higher-order approximation would materially alter the pruning decisions or improve the final pruning performance. This question is particularly relevant because previous studies have pointed out that first-order Taylor saliency may be inaccurate for some layers, especially deeper ones, where units with small first-order scores can still induce non-negligible loss increases when removed~\cite{molchanov2019importance}.

To examine this issue, we conduct a sensitivity analysis comparing first-order and second-order Taylor approximations within the R-Score. Our goal is not to replace the main pipeline, but to evaluate whether the order of the Taylor approximation substantially changes the resulting pruning decisions and the final model performance.

\subsection{Experimental setup}

We perform this analysis on CIFAR-10 using a CIFAR-style VGG16-BN backbone trained from scratch. The model reaches a test accuracy of \textbf{92.26\%} before pruning. To make the comparison as controlled as possible, the pre-pruning model is kept fixed throughout the experiment, and only the approximation order used in the R-Score is changed.

For the local sensitivity analysis, we focus on the deepest convolutional layer (\texttt{features.40}) and compare:
\begin{enumerate}
	\item the first-order Taylor approximation,
	\item the second-order Taylor approximation,
	\item the exact loss increase caused by removing a single channel.
\end{enumerate}
The calibration subset contains 256 examples. For each pruning budget, we compare the removal sets selected by the first-order and second-order approximations, report their Jaccard similarity, and measure the one-shot test accuracy drop after removing the selected channels.

For a more global check, we further construct an \emph{all-layer FAIR-like} pruning pipeline in which:
(1) the U-Score is fixed as a class-conditional 1D Wasserstein score computed from GAP activations, (2) the R-Score is instantiated either by the first-order or the second-order Taylor approximation, (3) all convolutional layers are pruned under the same ToD level.

We then compare the resulting global pruning ratios and the performance of the pruned models across a range of ToD levels. Finally, under two representative settings, namely $\alpha=0.10$ and $\alpha=0.30$, we fine-tune the first-order and second-order pruned models for three epochs and record their test accuracy after each epoch.

\subsection{Results on a representative deep layer}

Table~\ref{tab:taylor_order_deep_layer} reports the results for the deepest convolutional layer. Several observations can be made.

First, the removal sets selected by first-order and second-order Taylor approximations are not identical, especially under smaller pruning budgets. For example, when pruning 10\% of the channels in the deepest layer, the Jaccard similarity between the two removal sets is only 0.7895, indicating a noticeable difference in local channel ranking.

Second, despite this difference in ranking, the resulting one-shot performance gap is very small. Across all tested budgets, both approximations lead to only minor accuracy drops, and the second-order approximation is only marginally better. For example, at the 20\% pruning budget, the one-shot accuracy drop is 0.30 percentage points for the first-order approximation and 0.20 percentage points for the second-order approximation.

These results suggest that higher-order curvature information does alter the local saliency ordering of channels, but its impact on final single-layer pruning performance is limited in this setting.

\begin{table}
	\centering
	\caption{Sensitivity of the deepest convolutional layer (\texttt{features.40}) to the order of the Taylor approximation in the R-Score. ``Drop'' denotes the absolute decrease in test accuracy after one-shot channel removal.}
	\label{tab:taylor_order_deep_layer}
	\setlength{\tabcolsep}{7pt}
	\begin{tabular}{cccc}
		\toprule
		Pruning rate & Jaccard(1st vs.\ 2nd) & 1st-order drop & 2nd-order drop \\
		\midrule
		10\% & 0.7895 & 0.12\% & \textbf{0.09}\% \\
		20\% & 0.8545 & 0.30\% & \textbf{0.20}\% \\
		30\% & 0.9371 & 0.51\% & \textbf{0.47}\% \\
		\bottomrule
	\end{tabular}
\end{table}

\subsection{All-layer pruning under different ToD levels}

We next investigate whether the approximation order has a larger effect when all convolutional layers are pruned jointly. Table~\ref{tab:taylor_order_global_multi_alpha} summarizes the all-layer FAIR-like results across different ToD levels.

Several trends are worth noting. First, the first-order and second-order approximations can lead to non-negligible differences in the selected removal sets, as reflected by the average Jaccard similarity across layers. This discrepancy is most visible in the relatively conservative-to-moderate pruning regime. For example, at $\alpha=0.10$, the average removal-set Jaccard is only 0.7503.

Second, the approximation order can substantially affect the one-shot performance in the global pruning setting, especially at smaller ToD levels. At $\alpha=0.00$, the first-order variant loses 64.24 percentage points of test accuracy, whereas the second-order variant loses 24.93 percentage points. A similarly large gap is observed at $\alpha=0.10$, where the second-order approximation yields a considerably better one-shot starting point.

Third, as the ToD level increases and pruning becomes more aggressive, the difference between the two approximations gradually diminishes. Beyond $\alpha \geq 0.20$, both variants collapse to similarly poor one-shot accuracy, suggesting that the aggressiveness of the global pruning decision dominates the effect of the Taylor approximation order in this regime.

Overall, these results indicate that the order of the Taylor approximation matters more in the \emph{global one-shot pruning stage} than in the single-layer local analysis. In particular, the second-order approximation may provide a noticeably better initialization when the pruning decision is taken jointly across all layers.

\begin{table*}
	\centering
	\caption{All-layer FAIR-like pruning under different ToD levels. ``Pruning rate'' denotes the proportion of trainable parameters removed, i.e., \(1-\text{keep ratio}\). ``Drop'' denotes the absolute decrease in test accuracy relative to the original unpruned model.}
	\label{tab:taylor_order_global_multi_alpha}
	\setlength{\tabcolsep}{6pt}
	\begin{tabular}{ccccccc}
		\toprule
		ToD level $\alpha$ & Avg.\ Jaccard & 1st pruning rate & 2nd pruning rate & 1st drop & 2nd drop \\
		\midrule
		0.00 & 0.8265 & 8.39\% & 8.56\% & 64.24\% & 24.93\% \\
		0.05 & 0.7996 & 12.64\% & 12.61\% & 68.90\% & 27.74\% \\
		0.10 & 0.7503 & 23.40\% & 22.57\% & 75.65\% & 43.31\% \\
		0.15 & 0.8410 & 30.35\% & 29.79\% & 80.55\% & 80.39\% \\
		0.20 & 0.8913 & 38.16\% & 37.59\% & 82.26\% & 82.06\% \\
		0.25 & 0.8819 & 45.16\% & 45.94\% & 82.26\% & 82.26\% \\
		0.30 & 0.8841 & 51.66\% & 53.57\% & 82.26\% & 82.26\% \\
		\bottomrule
	\end{tabular}
\end{table*}

\subsection{Short fine-tuning under representative ToD levels}

To further assess the practical impact of the Taylor approximation order, we fine-tune the globally pruned models for three epochs under two representative ToD levels: a moderate setting ($\alpha=0.10$) and a more aggressive setting ($\alpha=0.30$). The results are reported in Table~\ref{tab:taylor_order_finetune}.

At $\alpha=0.10$, although the second-order approximation provides a much better one-shot starting point than the first-order approximation, this gap shrinks rapidly after short fine-tuning. After three epochs, the two variants become nearly indistinguishable, reaching 87.60\% and 87.64\% test accuracy, respectively.

At $\alpha=0.30$, the pruning is substantially more aggressive, with keep ratios of only 48.34\% and 46.43\% for the first-order and second-order variants. In this regime, both methods start from heavily degraded one-shot models, but the second-order approximation recovers somewhat better after fine-tuning. After three epochs, the second-order variant reaches 86.91\%, compared with 85.16\% for the first-order variant.

Taken together, these results suggest that higher-order information can matter more under aggressive global pruning, especially in terms of the quality of the pruned model before or during early fine-tuning. Nevertheless, under moderate pruning, the final difference after short fine-tuning remains very small.

\begin{table}
	\centering
	\caption{Three-epoch fine-tuning after all-layer FAIR-like pruning under two representative ToD levels. ``Keep ratio'' denotes the ratio of trainable parameters after pruning to those before pruning.}
	\label{tab:taylor_order_finetune}
	\setlength{\tabcolsep}{6pt}
	\begin{tabular}{lccccc}
		\toprule
		Setting & Method & pruning rate & Epoch 1 & Epoch 2 & Epoch 3 \\
		\midrule
		\multirow{2}{*}{$\alpha=0.10$}
		& 1st-order & 23.40\% & 86.27\% & 87.40\% & 87.60\% \\
		& 2nd-order & 22.57\% & 87.83\% & 87.30\% & 87.64\% \\
		\midrule
		\multirow{2}{*}{$\alpha=0.30$}
		& 1st-order & 51.66\% & 83.78\% & 84.85\% & 85.16\% \\
		& 2nd-order & 53.57\% & 84.48\% & 84.46\% & 86.91\% \\
		\bottomrule
	\end{tabular}
\end{table}

\subsection{Discussion on the order of the Taylor expansion of R-Score}

Overall, the above results support a nuanced conclusion. The second-order Taylor approximation does affect local saliency ranking, but its impact on single-layer pruning performance is limited. In contrast, in the all-layer pruning setting, the approximation order can substantially influence the one-shot behavior, especially under conservative-to-moderate ToD levels, and may also affect the short-term fine-tuning trajectory under more aggressive pruning.

At the same time, the experiments do not suggest that the second-order approximation is uniformly necessary. Under moderate pruning, the final performance gap after short fine-tuning is very small, indicating that the first-order Taylor approximation remains a practically reasonable and computationally efficient default choice for the R-Score. Nevertheless, the stronger performance of the second-order approximation in some aggressive settings suggests that incorporating higher-order curvature information may still be a promising direction for future work, especially when one-shot performance or highly compressed models are of interest.

\subsection{R-Score magnitude profiles across layers}

\begin{figure*}[t]
	\centering
	\subfloat[\texttt{features.3}.]{
		\includegraphics[width=0.235\linewidth]{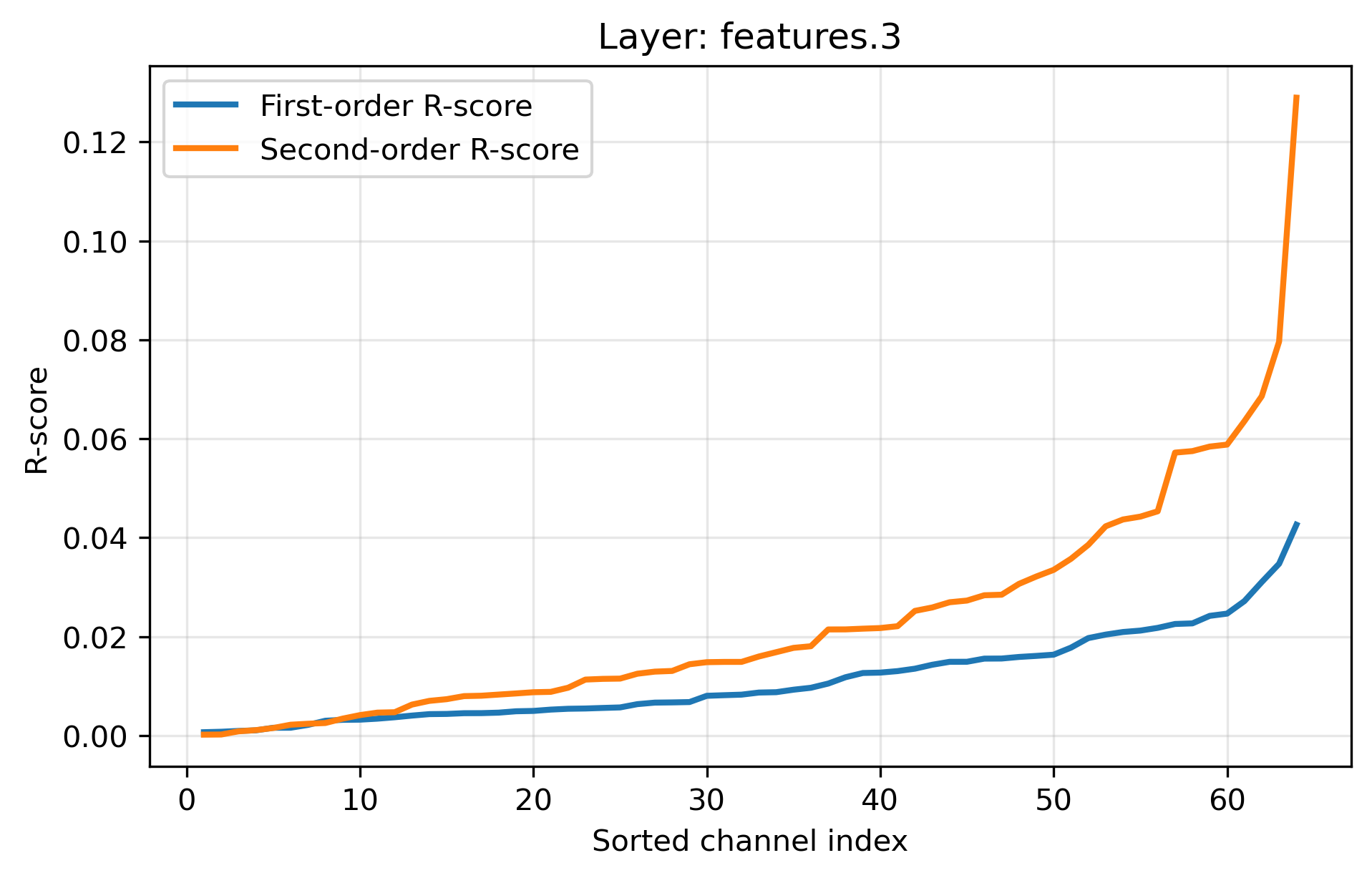}
	}
	\hfill
	\subfloat[\texttt{features.7}.]{
		\includegraphics[width=0.235\linewidth]{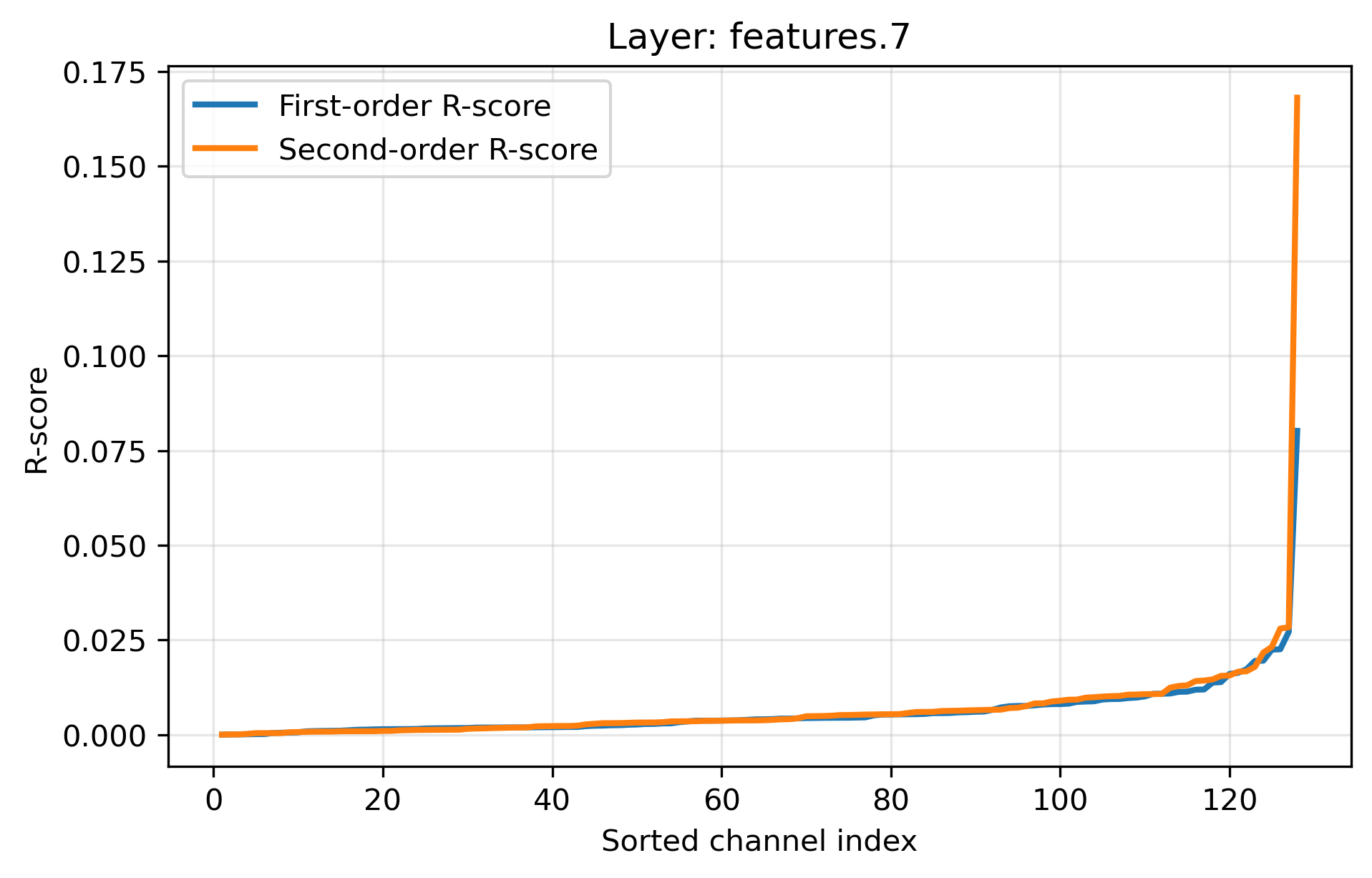}
	}
	\hfill
	\subfloat[\texttt{features.10}.]{
		\includegraphics[width=0.235\linewidth]{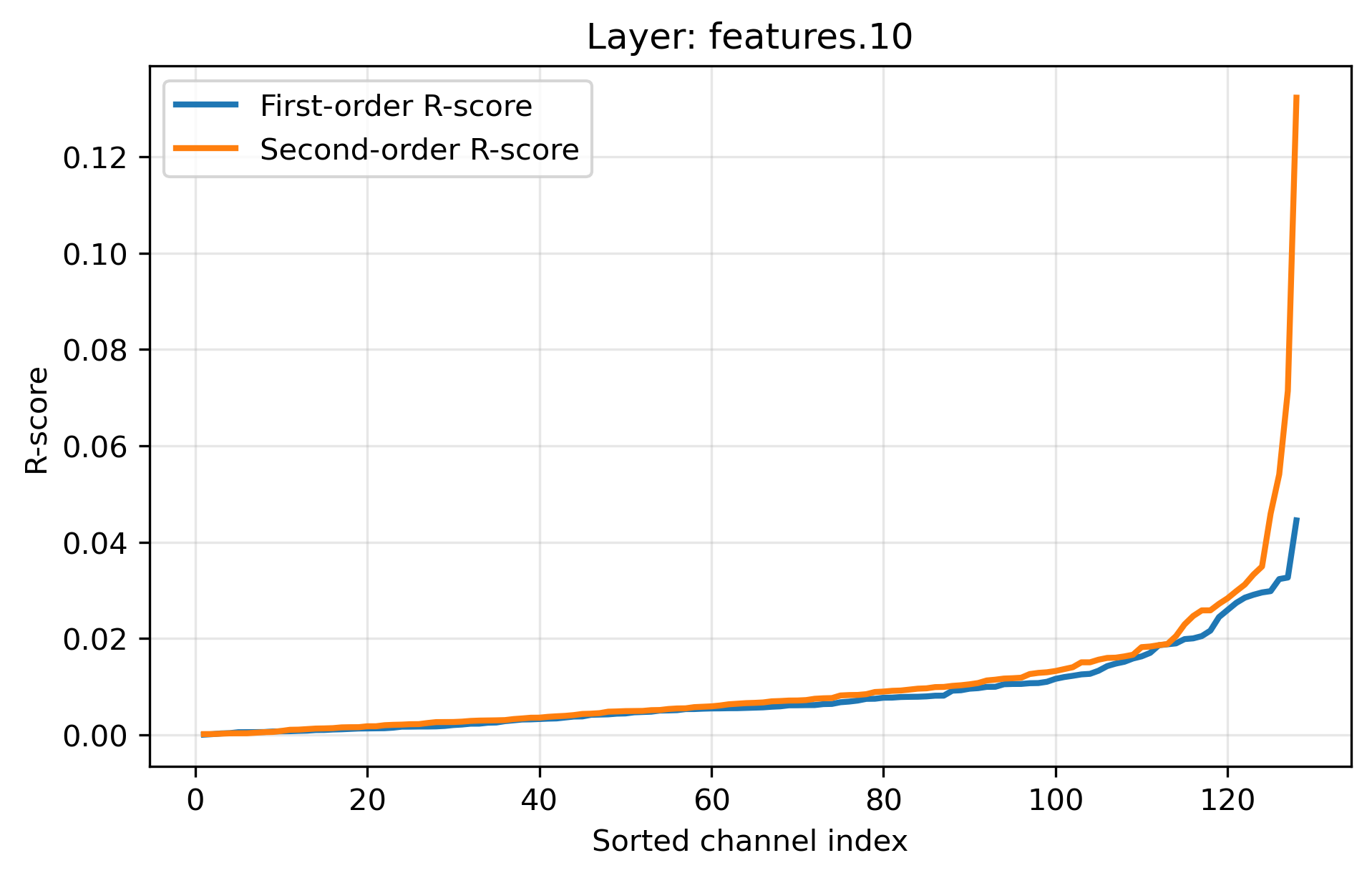}
	}
	\hfill
	\subfloat[\texttt{features.14}.]{
		\includegraphics[width=0.235\linewidth]{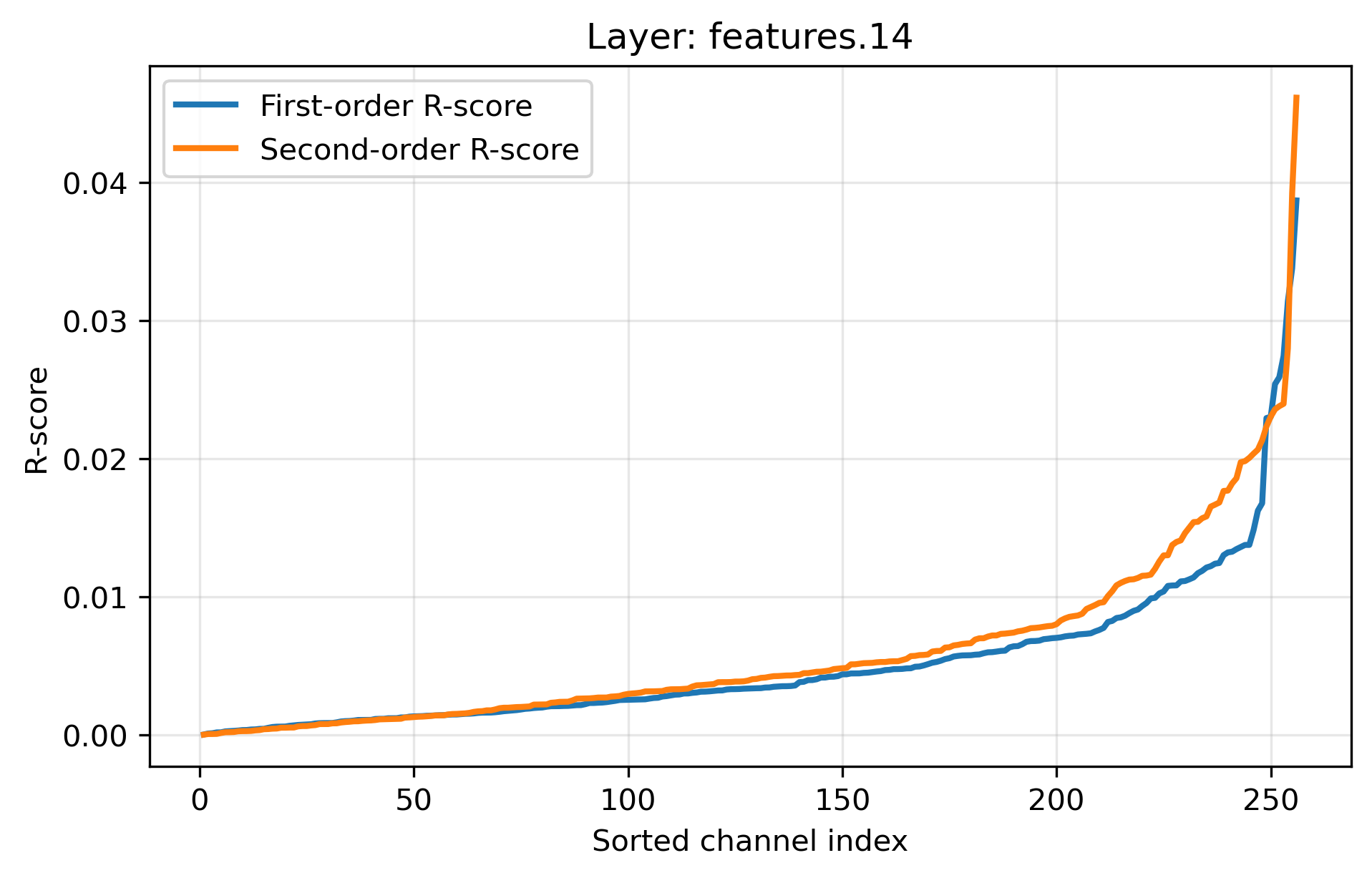}
	}
	\\
	\subfloat[\texttt{features.17}.]{
		\includegraphics[width=0.235\linewidth]{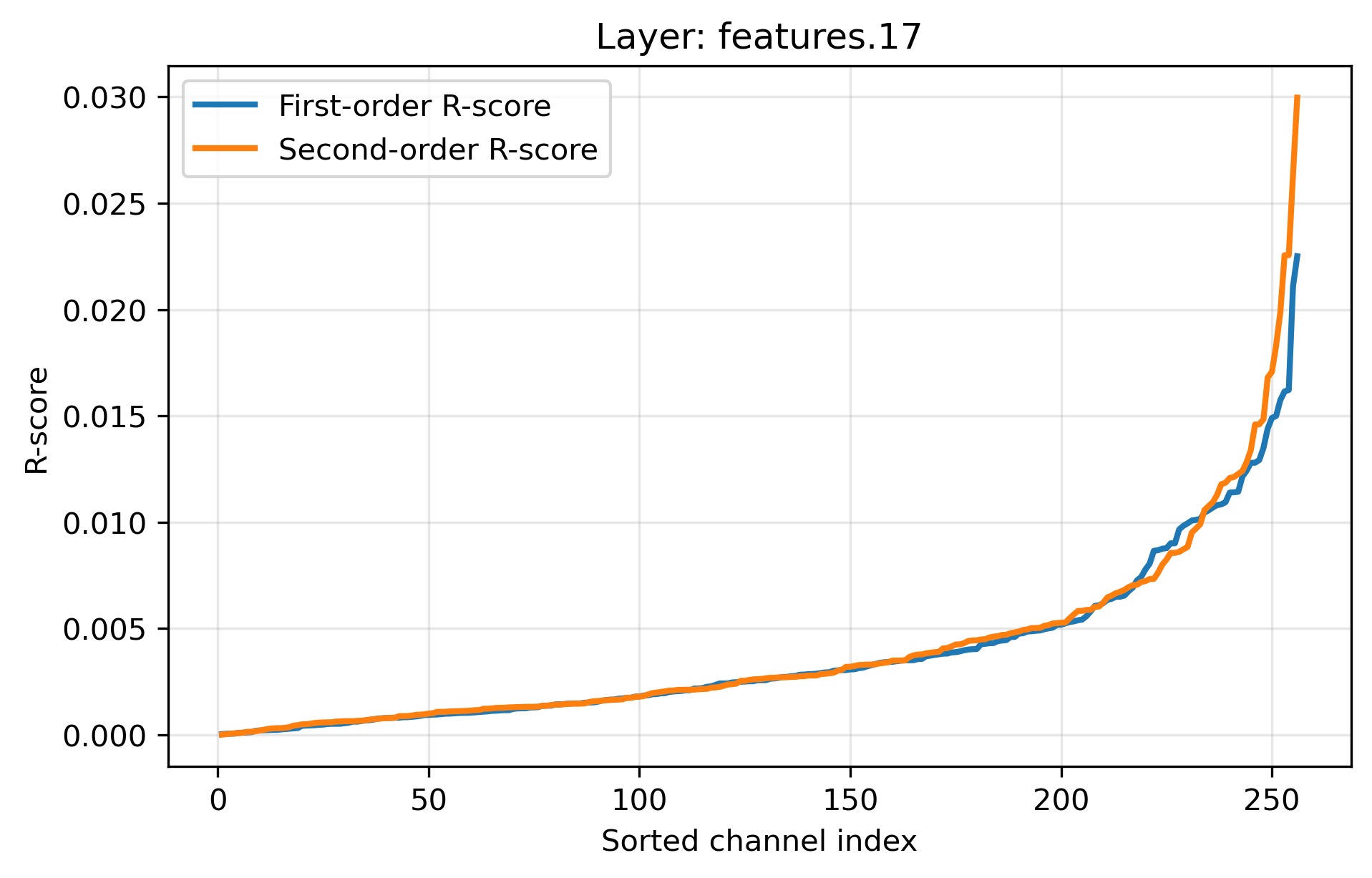}
	}
	\hfill
	\subfloat[\texttt{features.20}.]{
		\includegraphics[width=0.235\linewidth]{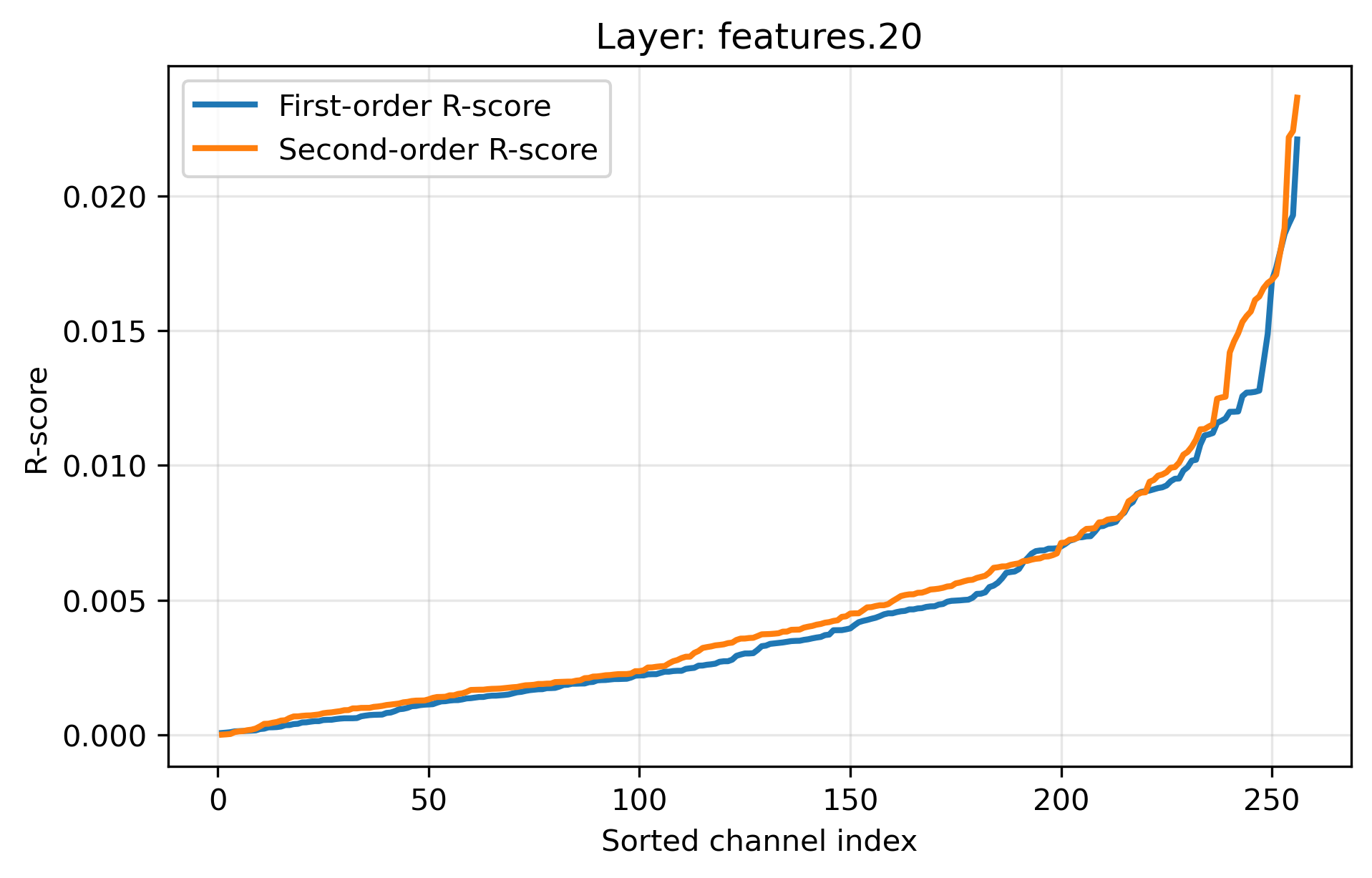}
	}
	\hfill
	\subfloat[\texttt{features.24}.]{
		\includegraphics[width=0.235\linewidth]{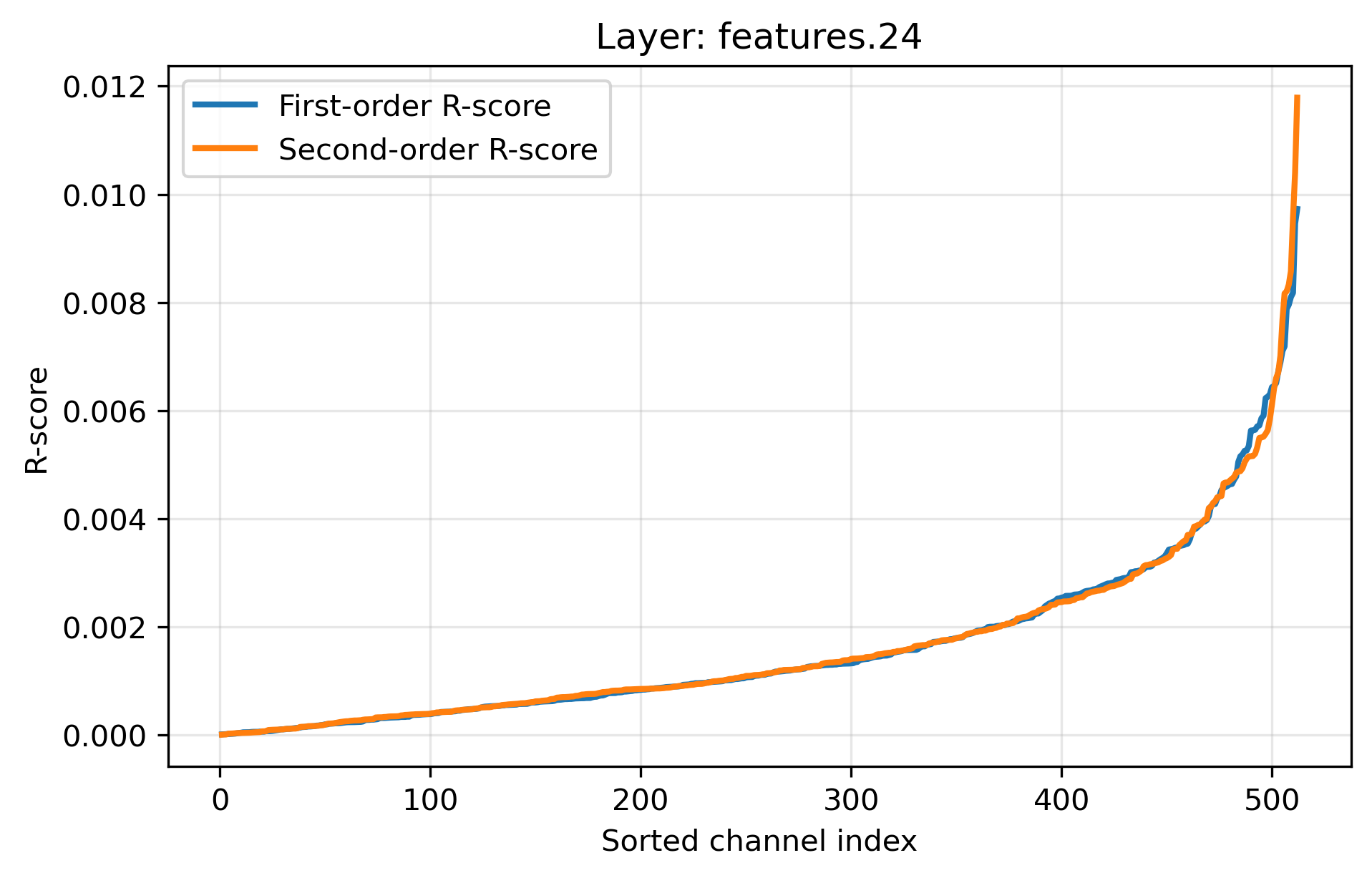}
	}
	\hfill
	\subfloat[\texttt{features.27}.]{
		\includegraphics[width=0.235\linewidth]{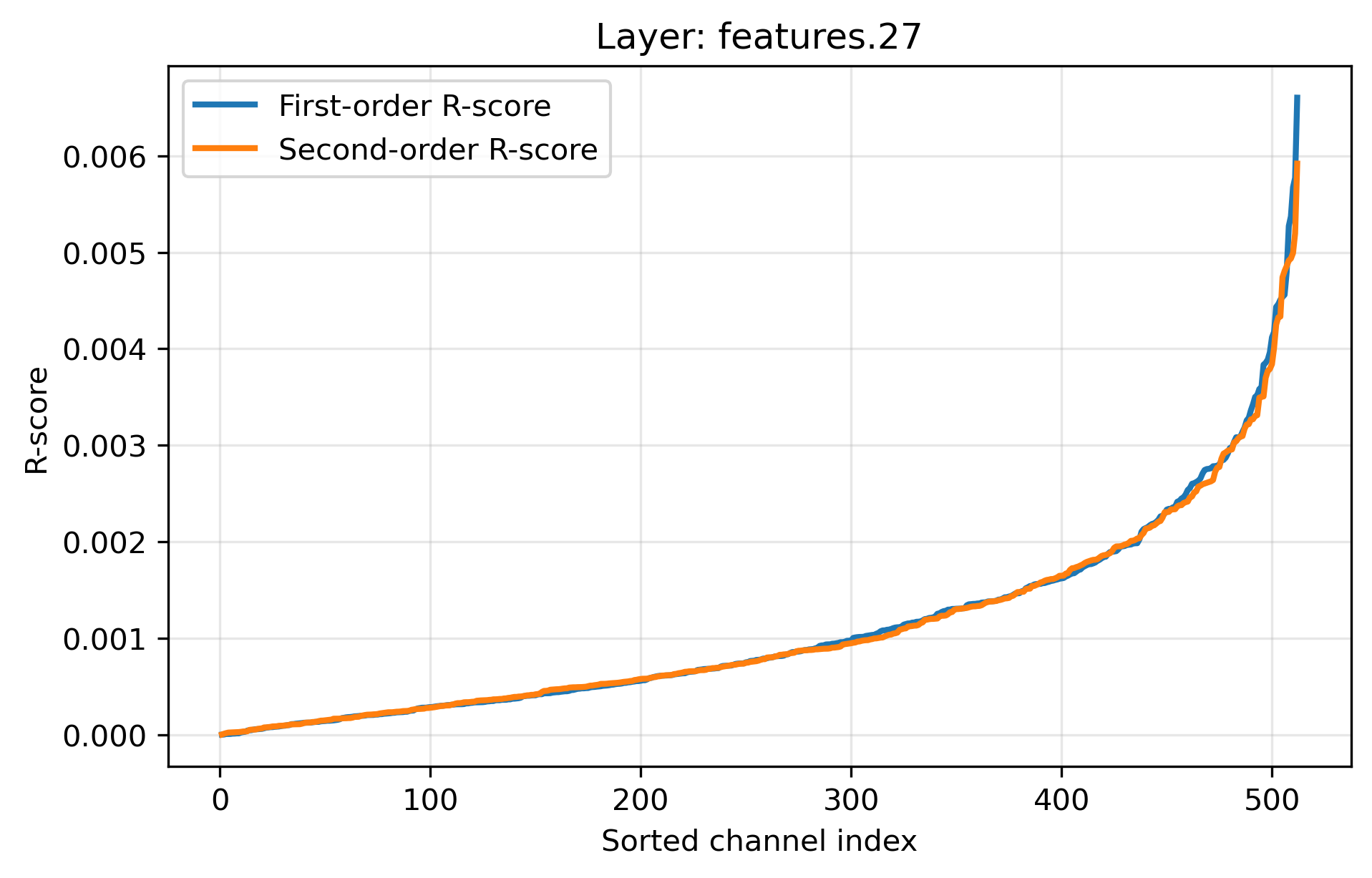}
	}
	\\
	\subfloat[\texttt{features.30}.]{
		\includegraphics[width=0.235\linewidth]{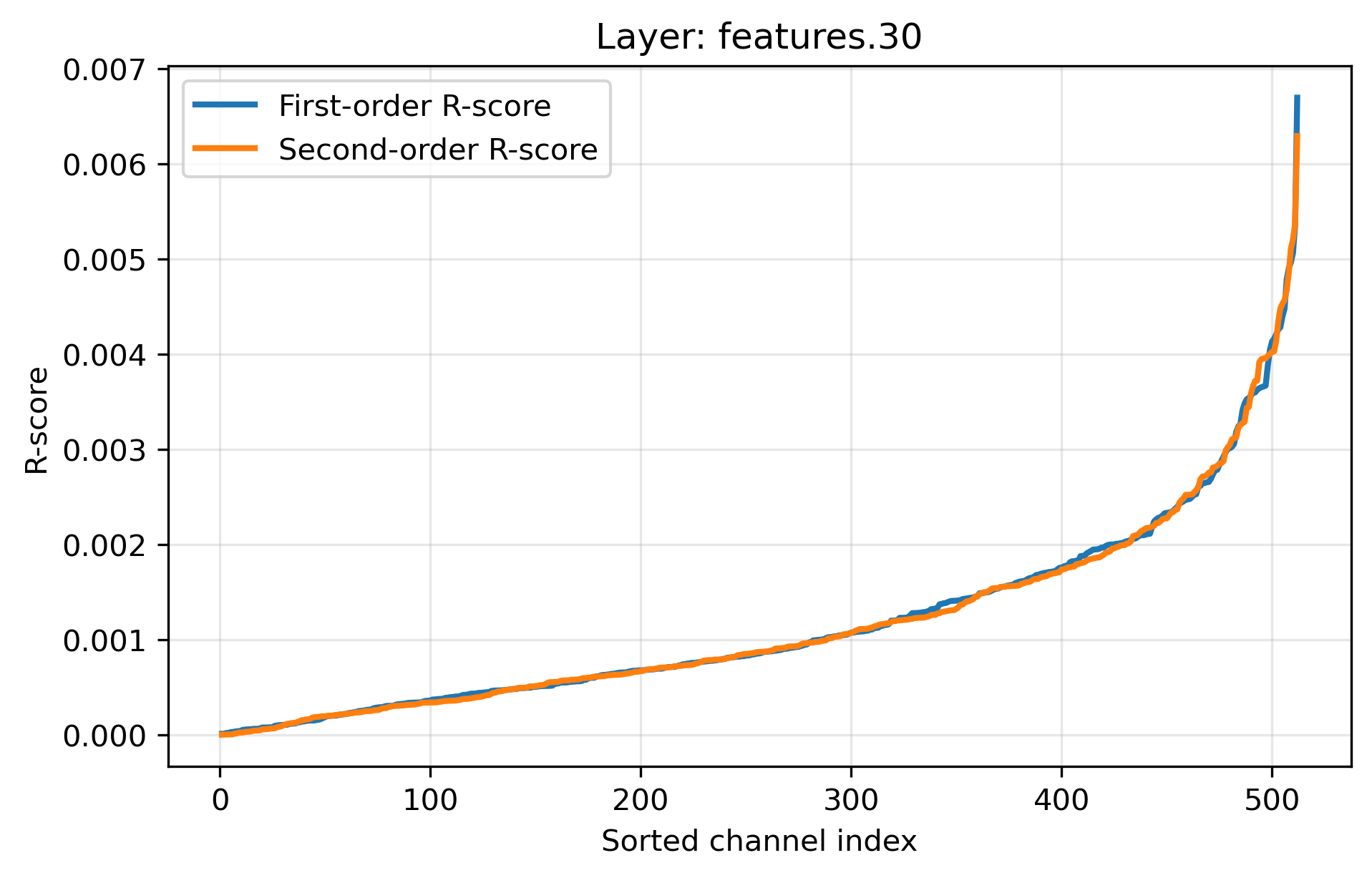}
	}
	\hfill
	\subfloat[\texttt{features.34}.]{
		\includegraphics[width=0.235\linewidth]{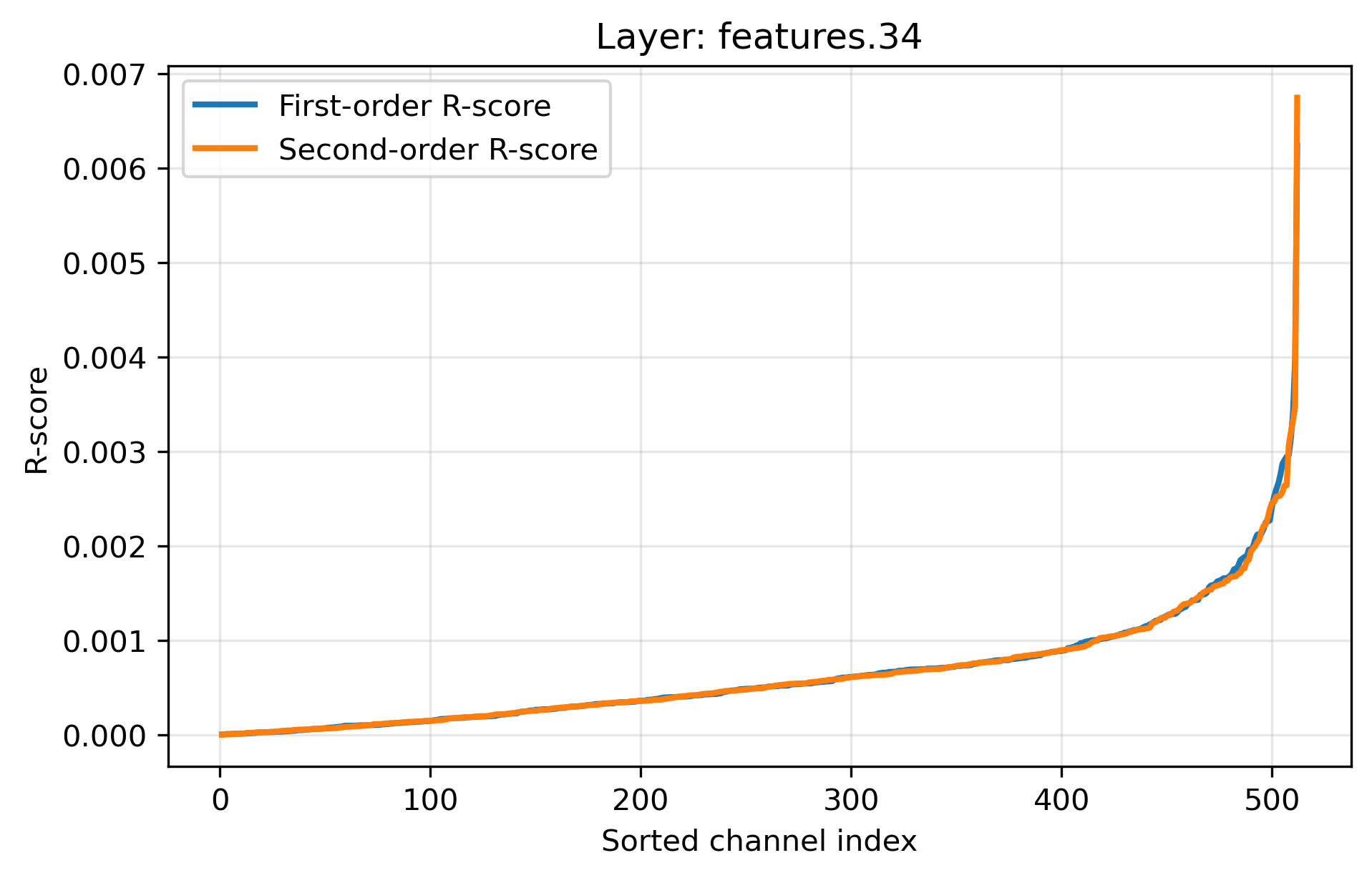}
	}
	\hfill
	\subfloat[\texttt{features.37}.]{
		\includegraphics[width=0.235\linewidth]{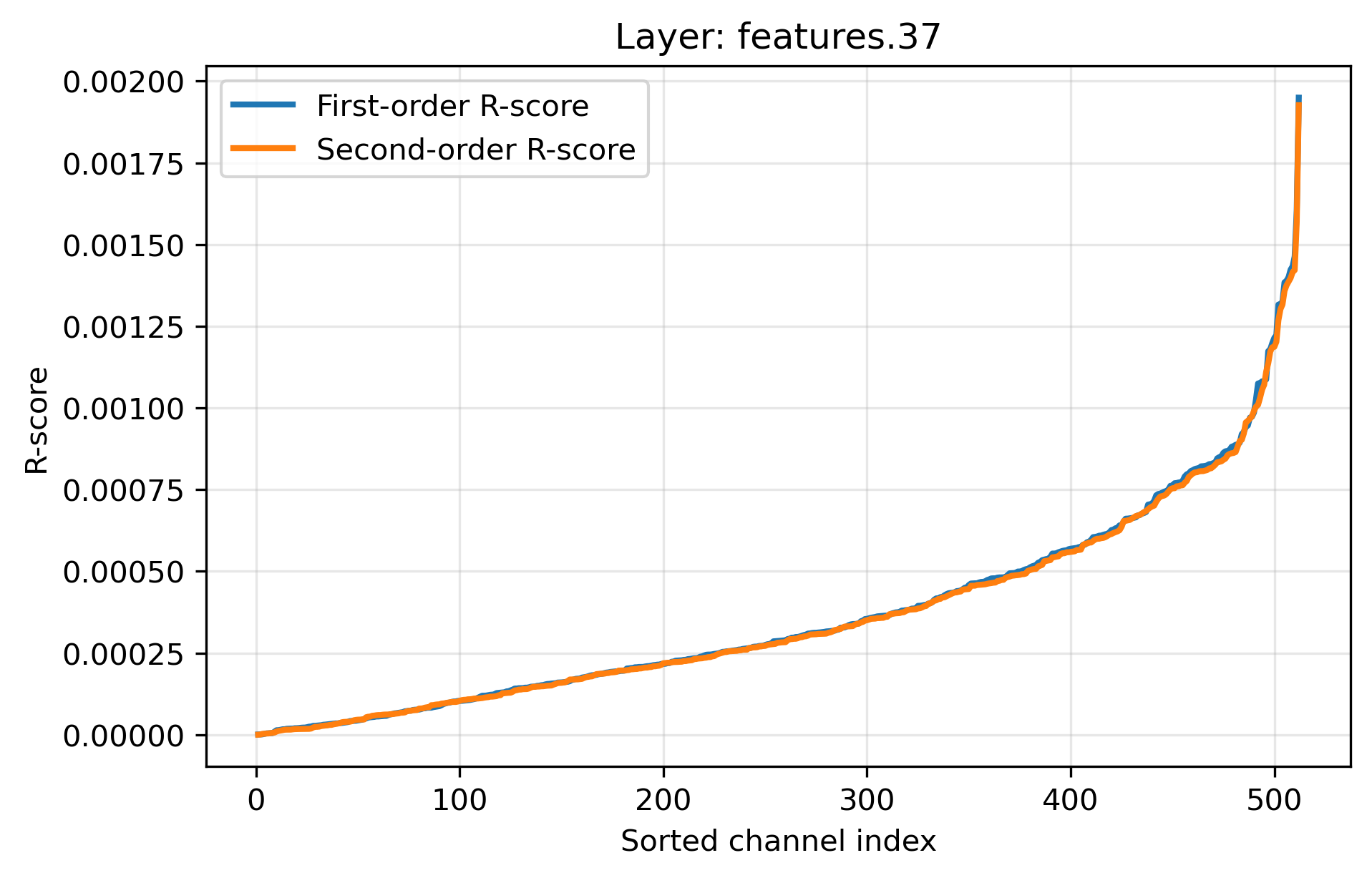}
	}
	\hfill
	\subfloat[\texttt{features.40}.]{
		\includegraphics[width=0.235\linewidth]{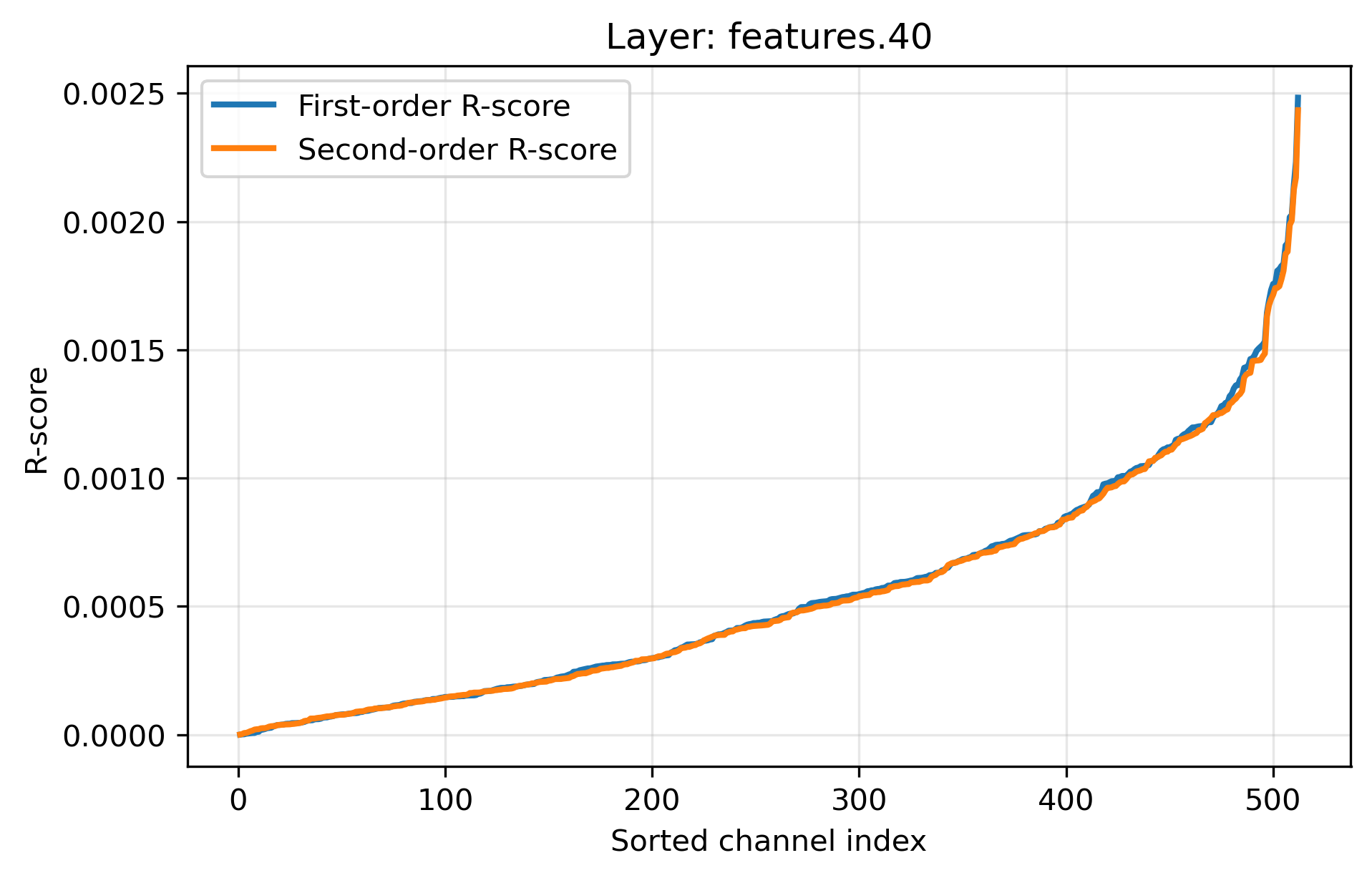}
	}
	\caption{Sorted channel-wise R-Score magnitude profiles across representative VGG16-BN layers. The first-order and second-order Taylor R-Scores generally show similar long-tailed trends, especially in the high-sensitivity tail, although their absolute scales can differ in shallow layers.}
	\label{fig:rscore_profiles_all}
\end{figure*}

The preceding tables compare first-order and second-order Taylor R-Scores mainly
through rank agreement and pruning performance. Figure~\ref{fig:rscore_profiles_all}
provides a complementary visualization of the sorted channel-wise R-Score
magnitudes across representative VGG16-BN layers. The two approximations are
not identical in absolute scale, especially in shallow layers, but they show
similar long-tailed magnitude profiles across most layers. This supports the
interpretation that the first-order R-Score captures the dominant magnitude
trend needed to identify the protected high-sensitivity tail, while the
second-order approximation can still alter local scale and the behavior of
aggressive global pruning.

\section{Additional Experimental Details}
\label{app:experimental_details}

\subsection{Detailed Experimental Settings for Fair SOTA Comparison}
\label{app:detailed_sota_settings}

To ensure a fair comparison with prior structured pruning methods, we use benchmark settings that are standard in the pruning literature and keep the evaluation protocol aligned with the compared methods whenever possible. Most experiments are conducted on CIFAR-10, which contains 50{,}000 training images and 10{,}000 test images from 10 classes with resolution \(32 \times 32\). We also evaluate on ImageNet-1K, which contains 1.28 million training images and 50{,}000 validation images from 1{,}000 classes, with input resolution \(224 \times 224\).

Our SOTA comparisons involve ResNet-family models and DenseNet. Unless otherwise specified, post-pruning fine-tuning is performed using SGD with momentum 0.9, batch size 256, initial learning rate 0.005, and cosine learning rate decay. On CIFAR-10, pruned models are fine-tuned for 300 epochs. On ImageNet-1K, we use the standard fine-tuning schedule adopted for the corresponding ResNet-family baselines. Standard data augmentation is used, including random crop and horizontal flip on CIFAR-10, and random resized crop and horizontal flip on ImageNet-1K, followed by normalization.

All FAIR-Pruner rows are produced by our implementation. Competitor rows are taken from the cited papers unless stated otherwise. The ConvNeXt block is the exception: we reproduce the competing criteria in the same ConvNeXt/CIFAR-100 setting and prune MLP hidden dimensions. For DeiT-B, our row prunes attention heads and MLP hidden dimensions, whereas the cited baselines keep their original ViT compression protocols, such as width/depth pruning or token reduction.

For pruning-score estimation, we use the entire CIFAR-10 training set as the pruning dataset. On ImageNet-1K, we use a randomly sampled 2\% subset of the training set as the pruning dataset to reduce computational overhead. This pruning dataset is used to compute the U-Score and R-Score in FAIR-Pruner. Unless otherwise stated, FAIR-Pruner does not use auxiliary enhancements such as knowledge distillation, teacher supervision, expanded search spaces, or other extra training tricks. When competing methods use such additional techniques, this is explicitly indicated in the comparison tables. Reported FLOPs are computed under the same input resolution and counting convention as the compared methods whenever possible.

\subsection{Scalable U-Score Approximations for Large-scale Backbones}
\label{app:scalable_uscore}

The default vision U-Score uses class-conditional separability to rank removable units. 
A direct implementation over all class pairs can become expensive on ImageNet-scale datasets, especially when the number of classes is large. 
For large-scale backbones, we therefore use scalable approximations that preserve the same role of the U-Score as a removal-oriented ranking signal while reducing the number of pairwise comparisons.

For ImageNet-1K/ResNet-50, we first extract penultimate-layer features from the publicly available unpruned baseline and compute the mean feature vector for each of the 1000 classes. 
We then cluster these class means into a smaller number of parent groups using $K$-means and compute the U-Score with respect to the parent-group labels. 
With 20 parent groups, the number of class-pair comparisons is reduced from $\binom{1000}{2}$ to $\binom{20}{2}$, which lowers the pairwise-comparison cost to roughly $1/2600$ of the original computation. 
For ConvNeXt and DeiT-B, we use the corresponding block-level or hidden-width statistics to obtain a scalable removal ranking, while keeping the ToD mechanism and R-Score protection rule unchanged.

\paragraph*{Grouped U-Score for ImageNet-scale backbones}
Let \(\mathcal G(Y)\in[G]\) denote the parent-group label obtained by clustering
class prototypes, and write \(Z_{\mathcal G=a}\) for an input drawn
conditionally on \(\mathcal G(Y)=a\). The grouped population U-Score is
\begin{equation}
	U_{j,\mathcal G}^{(l)}
	=
	\sup_{\substack{a,b\in[G]\\a\ne b}}
	d\!\left(
	O_j^{(l)}(Z_{\mathcal G=a}),
	O_j^{(l)}(Z_{\mathcal G=b})
	\right).
\end{equation}
This grouped score is not an unbiased estimator of the original 1000-class
U-Score. It is a scalable removal-oriented ranking signal that preserves the
role required by ToD: it provides a within-layer ordering of candidate
removable units. The consistency argument in Proposition~\ref{prop:uscore_consistency}
applies to this grouped population score with \(K\) replaced by \(G\), provided
the grouping rule is fixed after class-prototype construction.

These approximations are used only to make the removal-oriented U-Score scalable in large models. 
They do not change the definition of ToD, which continues to coordinate a removal-oriented ranking and a protection-oriented ranking to determine the layer-wise pruning depth.

\subsection{Resolution Stability of Parent-class U-Score Approximation}
\label{app:parent_uscore_stability}

To validate the parent-class approximation used above, we conduct a controlled
check on CIFAR-100, where the full fine-label U-Score can still be computed.
We compare unit rankings obtained from the full set of 100 fine classes with
rankings obtained from \(K\) parent classes constructed by \(K\)-means clustering
of class prototypes. The agreement is measured by Spearman rank correlation
within each block and unit type. The purpose of this experiment is not to claim
that the grouped U-Score is an unbiased estimate of the full fine-class U-Score,
but to check whether it preserves the removal ranking needed by ToD.

Table~\ref{tab:cifar100_uscore_meanK} shows that moderate values of \(K\) already
preserve the full-class ranking reasonably well while reducing the pairwise
class-comparison cost. For example, \(K=50\) gives an approximate \(4\times\)
reduction in class-pair cost and achieves mean Spearman correlations of 0.871
for attention heads and 0.892 for MLP units. Table~\ref{tab:cifar100_uscore_layerwiseK}
further shows that early blocks are nearly invariant to the class resolution,
whereas mid and late blocks are more sensitive. This suggests choosing \(K\)
based on the desired speed--stability trade-off, with special attention to late
layers. Finally, Table~\ref{tab:cifar100_official_vs_kmeans20} compares the
learned \(K=20\) parent classes with the official CIFAR-100 coarse hierarchy.
The two parent partitions give comparable rank agreement with the full
fine-class U-Score, supporting the use of learned parent classes when a native
semantic hierarchy is unavailable.

\begin{table}[t]
	\caption{CIFAR-100 parent-class approximation. Spearman rank correlation between U-Score rankings computed using all 100 fine classes and those computed using \(K\) \(K\)-means parent classes. The class-pair computation cost scales approximately as \(O(K^2)\), so the cost ratio is \((K/100)^2\).}
	\label{tab:cifar100_uscore_meanK}
	\centering
	\small
	\setlength{\tabcolsep}{3pt}
	\begin{tabular}{ccccc}
		\toprule
		\multirow{2}{*}{\(K\)} & \multirow{2}{*}{Cost ratio} & \multirow{2}{*}{Speedup} & \multicolumn{2}{c}{Mean Spearman vs.\ full} \\
		\cmidrule(lr){4-5}
		& & & Head U-Score & MLP U-Score \\
		\midrule
		10 & 0.01 & 100.00 & 0.798 & 0.730 \\
		20 & 0.04 & 25.00  & 0.746 & 0.784 \\
		30 & 0.09 & 11.11  & 0.799 & 0.838 \\
		50 & 0.25 & 4.00   & 0.871 & 0.892 \\
		80 & 0.64 & 1.56   & 0.938 & 0.953 \\
		\bottomrule
	\end{tabular}
	
\end{table}

\begin{table*}[t]
	\caption{Layer-wise resolution stability on CIFAR-100. Spearman rank correlation between full-class U-Score rankings and \(K\)-parent-class U-Score rankings, averaged over early blocks (0--2), middle blocks (3--7), and late blocks (8--11).}
	\label{tab:cifar100_uscore_layerwiseK}
	\centering
	\small
	\setlength{\tabcolsep}{6pt}
	\begin{tabular}{ccccccc}
		\toprule
		\multirow{2}{*}{\(K\)} & \multicolumn{3}{c}{Head Spearman} & \multicolumn{3}{c}{MLP Spearman} \\
		\cmidrule(lr){2-4}\cmidrule(lr){5-7}
		& Early & Mid & Late & Early & Mid & Late \\
		\midrule
		10 & 0.937 & 0.808 & 0.680 & 0.887 & 0.781 & 0.548 \\
		20 & 0.979 & 0.720 & 0.605 & 0.959 & 0.831 & 0.594 \\
		30 & 0.984 & 0.787 & 0.675 & 0.975 & 0.885 & 0.678 \\
		50 & 0.981 & 0.866 & 0.795 & 0.984 & 0.924 & 0.784 \\
		80 & 0.977 & 0.916 & 0.935 & 0.992 & 0.966 & 0.909 \\
		\bottomrule
	\end{tabular}
\end{table*}

\begin{table*}[t]
	\caption{CIFAR-100 official hierarchy versus learned parent classes. Entries report Spearman rank correlation against the full 100-class U-Score rankings on representative blocks. The learned \(K=20\) parent partition behaves similarly to the dataset's native coarse grouping as a U-Score ranking surrogate.}
	\label{tab:cifar100_official_vs_kmeans20}
	\centering
	\small
	\setlength{\tabcolsep}{8pt}
	\begin{tabular}{ccccc}
		\toprule
		\multirow{2}{*}{Block} & \multicolumn{2}{c}{Official coarse classes} & \multicolumn{2}{c}{\(K\)-means parents (\(K=20\))} \\
		\cmidrule(lr){2-3}\cmidrule(lr){4-5}
		& Head & MLP & Head & MLP \\
		\midrule
		0  & 0.972 & 0.972 & 0.958 & 0.975 \\
		5  & 0.902 & 0.808 & 0.811 & 0.806 \\
		10 & 0.636 & 0.600 & 0.483 & 0.578 \\
		\bottomrule
	\end{tabular}
\end{table*}

\subsection{Additional Layer-wise Allocation Profiles}
\label{app:additional_layerwise_profiles}

\begin{figure*}[t]
	\centering
	\subfloat[AlexNet profile.]{
		\includegraphics[width=0.48\linewidth]{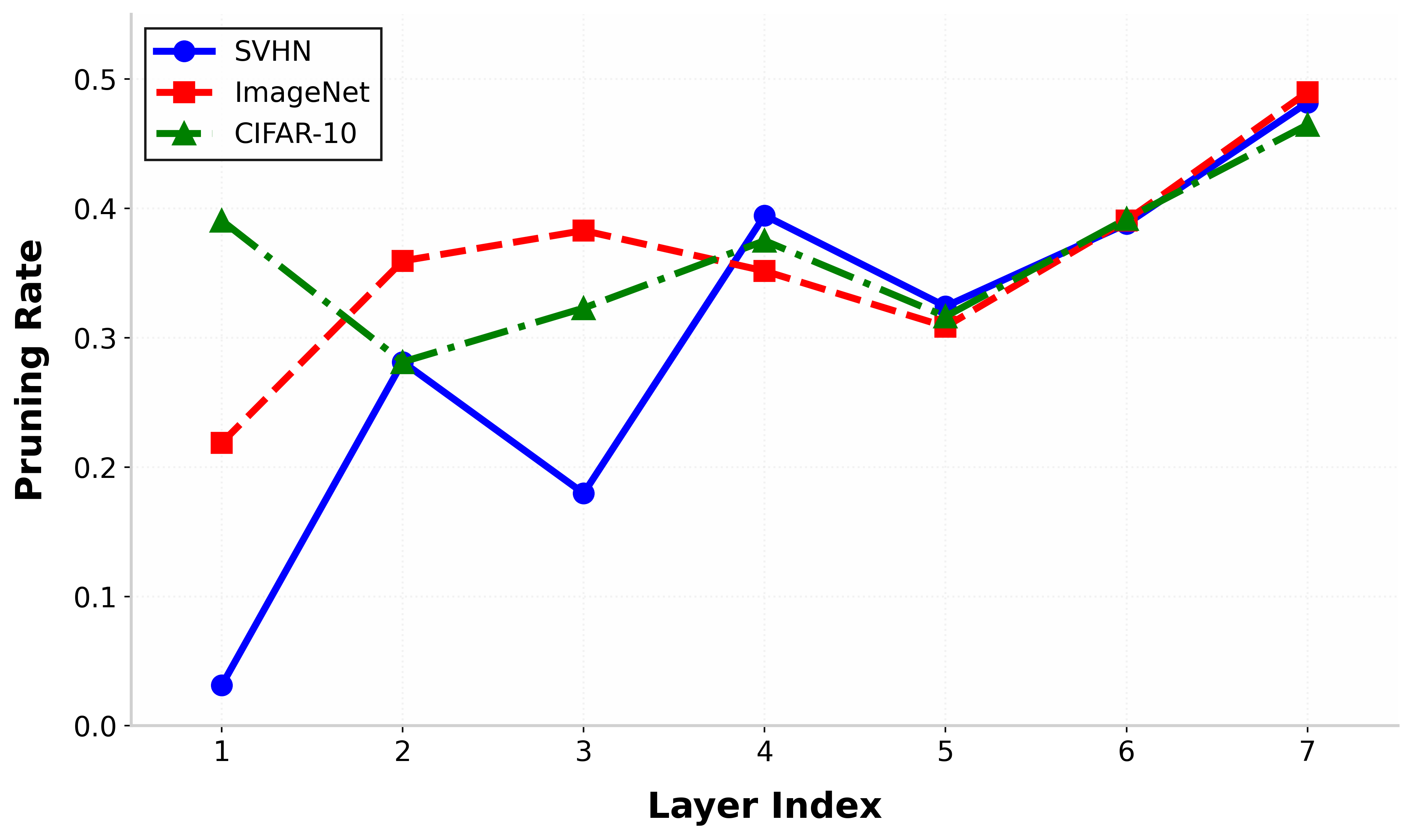}
		\label{fig:layerwise_alexnet_appendix}
	}
	\hfill
	\subfloat[DeiT-B profile.]{
		\includegraphics[width=0.48\linewidth]{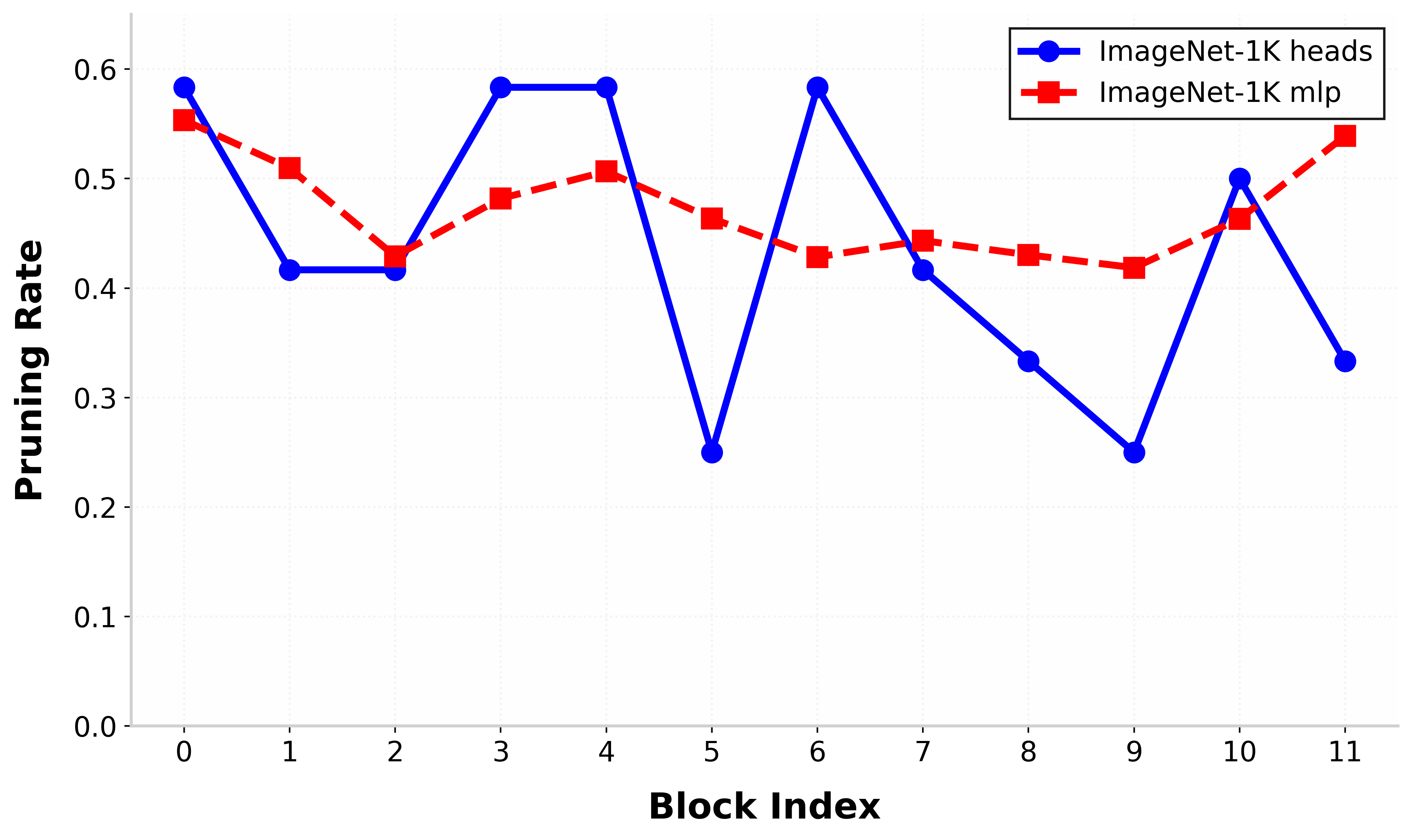}
		\label{fig:layerwise_deit_appendix}
	}
	\caption{Additional layer-wise pruning profiles induced by ToD. 
		These results complement Figure~\ref{fig:layer-wise_pr} and show that the non-uniform allocation behavior is also observed on AlexNet and DeiT-B.}
	\label{fig:additional_layerwise_profiles}
\end{figure*}

Figure~\ref{fig:additional_layerwise_profiles} reports additional layer-wise pruning profiles for AlexNet and
DeiT-B, complementing the representative CNN and MoE profiles in Figure~\ref{fig:layer-wise_pr}.

\section{Detailed Results for MoE LLM Pruning}
\label{app:llm_details}
\begin{table*}[t]
\centering
\caption{Per-category accuracy (\%) on MMLU-Pro under the prune-only 50\% routed-expert pruning setting. 
For score-based pruning methods, calibration is performed only on the official validation split. 
FAIR-Pruner uses soft activation as the removal-oriented expert signal and ToD to induce non-uniform layer-wise expert budgets. 
Random results are reported as mean $\pm$ standard deviation over 5 seeds.}
\label{tab:mmlu_pro_subsets_keep30_strict}
\footnotesize
\setlength{\tabcolsep}{3.6pt}

\begin{tabular}{lccccccc}
\toprule
Method
& Biology
& Business
& Chemistry
& Comp.\ Sci.
& Economics
& Engineering
& Health \\
\midrule
Full Model
& 34.17 & 15.08 & 10.95 & 20.49 & 28.32 & 10.84 & 19.93 \\
Soft Act. (50\%)
& 10.32 & 10.39 & 8.48 & 13.17 & 13.27 & 9.18 & 11.25 \\
FAIR-Pruner (50\%)
& 12.83 & 11.15 & 8.66 & 16.34 & 13.74 & 9.49 & 12.35 \\
Random (50\%)
& 9.93$\pm$3.03
& 8.67$\pm$2.73
& 8.13$\pm$2.65
& 9.80$\pm$2.57
& 10.59$\pm$1.52
& 7.47$\pm$2.66
& 9.39$\pm$2.37 \\
Frequency (50\%)
& 0.28 & 0.00 & 0.00 & 0.24 & 0.00 & 0.10 & 0.00 \\
\bottomrule
\end{tabular}

\vspace{1.2mm}

\begin{tabular}{lccccccc}
\toprule
Method
& History
& Law
& Math
& Other
& Philosophy
& Physics
& Psychology \\
\midrule
Full Model
& 20.47 & 14.80 & 11.92 & 19.59 & 19.04 & 12.09 & 32.46 \\
Soft Act. (50\%)
& 13.12 & 10.99 & 9.99 & 12.55 & 11.42 & 8.39 & 12.03 \\
FAIR-Pruner (50\%)
& 14.17 & 11.08 & 10.73 & 12.55 & 12.42 & 9.47 & 13.91 \\
Random (50\%)
& 8.19$\pm$2.33
& 8.97$\pm$1.99
& 7.91$\pm$2.61
& 9.57$\pm$1.55
& 9.94$\pm$2.00
& 7.59$\pm$2.72
& 9.60$\pm$3.19 \\
Frequency (50\%)
& 0.79 & 0.00 & 0.00 & 0.22 & 0.00 & 0.00 & 0.00 \\
\bottomrule
\end{tabular}
\end{table*}

\paragraph*{MoE evaluation protocol}
The MoE experiments use Qwen1.5-MoE-A2.7B-Chat and prune routed experts only;
non-expert parameters and shared components are kept unchanged. No recovery
procedure is applied after pruning: there is no fine-tuning, distillation,
expert merging, or parameter update. Score calibration uses the official
validation split, and all compared methods are evaluated in the same zero-shot
generation-based protocol with the same prompt template, decoding rule, answer
parser, and budget of kept routed experts. Parse failures are counted as
incorrect and reported through the parse rate.

Table~\ref{tab:mmlu_pro_subsets_keep30_strict} provides the per-category MMLU-Pro results under the prune-only 50\% routed-expert pruning setting. 
The detailed breakdown shows that the improvement of FAIR-Pruner over uniform soft-activation pruning is not driven by a single category. 
Instead, FAIR-Pruner tends to provide small but consistent gains across multiple subjects, which is consistent with the interpretation that ToD improves the allocation of the same total routed-expert budget.

\section{Comparison with the Lottery Ticket Hypothesis Paradigm}
\begin{table}[t]
	\caption{Comparison of the proposed FAIR-Pruner and LTH. The best-performing method is highlighted by bold text.}
	\centering
	\setlength{\tabcolsep}{1.5mm}
	\begin{tabular}{c|c|c|c c}
		\hline
		\multirow{2}{*}{\begin{tabular}[c]{@{}c@{}}Model\\(Dataset)\end{tabular}
		} & \multirow{2}{*}{Pruning rate} & \multicolumn{1}{c|}{One-shot} & \multicolumn{2}{c}{Fine-tuning(epochs)}  \\ \cline{3-5} 
		&   & \multicolumn{1}{c|}{FAIR} & \multicolumn{1}{c}{FAIR}   & LTH     \\ \hline
		\multirow{3}{*}{\begin{tabular}[c]{@{}c@{}}VGG16\\(CIFAR10)\end{tabular}
		}   & (Unpruned)  & \multicolumn{3}{c}{91.1\%}  \\ \cline{2-5}
		& 35.5\%    & \multicolumn{1}{c|}{\textbf{90.3\%}} & \multicolumn{1}{c}{\textbf{90.5\%}(10)} & 83.6\%(240) \\ 
		& 63.2\% & \multicolumn{1}{c|}{82.3\%} & \multicolumn{1}{c}{\textbf{90.1\%}(10)} & 84.6\%(292)   \\
		\hline 
		\multirow{3}{*}{\begin{tabular}[c]{@{}c@{}}VGG16\\(SVHN)\end{tabular}
		}  & (Unpruned)  & \multicolumn{3}{c}{94.3\%}    \\   \cline{2-5} 
		& 41.1\%   & \multicolumn{1}{c|}{\textbf{94.3\%}}  & \multicolumn{1}{c}{\textbf{94.4\%}(10)} & 87.9\%(126) \\ 
		& 70.8\%   & \multicolumn{1}{c|}{\textbf{90.2\%}}  & \multicolumn{1}{c}{\textbf{93.9\%}(10)} & 88.2\%(125) \\ \hline
		\multirow{3}{*}{\begin{tabular}[c]{@{}c@{}}AlexNet\\(SVHN)\end{tabular}
		}   & (Unpruned)  & \multicolumn{3}{c}{85.4\%}    \\   \cline{2-5}
		& 29.5\% & \multicolumn{1}{c|}{82.2\%}  & \multicolumn{1}{c}{\textbf{84.8\%}(10)} &  84.5\%(205) \\ 
		& 38.4\%   & \multicolumn{1}{c|}{80.8\%} & \multicolumn{1}{c}{84.8\%(10)} & \textbf{85.0\%}(229) \\ 
		\hline
		\multirow{3}{*}{\begin{tabular}[c]{@{}c@{}}AlexNet\\(CIFAR10)\end{tabular}
		} & (Unpruned) & \multicolumn{3}{c}{90.9\%}   \\  \cline{2-5} 
		& 10.3\%  & \multicolumn{1}{c|}{\textbf{90.4\%}} & \multicolumn{1}{c}{\textbf{92.0\%}(10)} &  84.9\%(264) \\ 
		& 31.4\%  & \multicolumn{1}{c|}{\textbf{84.8\%}} & \multicolumn{1}{c}{\textbf{91.4\%}(10)} &  84.4\%(260) \\ 
		\hline
	\end{tabular}
	\label{table:LTH-Comparison}
\end{table}
We further compare FAIR-Pruner with the \textbf{Lottery Ticket Hypothesis} (LTH) paradigm~\cite{frankle2018lottery}. This comparison is included not because LTH is a direct baseline for one-shot structured pruning, but because it represents a well-known iterative pruning-and-retraining strategy for identifying sparse subnetworks. In the standard LTH procedure, a dense model is repeatedly trained, pruned, and reset to its original initialization. While this process can discover strong sparse subnetworks, it is computationally expensive and does not naturally support flexible generation of multiple pruning levels from a single scoring pass.

By contrast, FAIR-Pruner incurs negligible overhead when sweeping across
compression levels once the U-Scores and R-Scores have been computed. This makes
the comparison informative from a practical perspective: it contrasts a
one-shot, score-based pruning framework with a canonical iterative pruning
paradigm.

Table~\ref{table:LTH-Comparison} summarizes the results across multiple datasets and architectures. To keep the comparison conservative, FAIR-Pruner is allowed only ten epochs of post-pruning fine-tuning. Even under this restricted setting, FAIR-Pruner performs strongly. In several configurations, the one-shot accuracy of FAIR-Pruner already exceeds the retrained accuracy of LTH; after ten epochs of fine-tuning, FAIR-Pruner outperforms LTH in most of the reported cases. For example, on VGG16/CIFAR-10 at 63.2\% pruning, FAIR-Pruner reaches 90.1\% after ten epochs, compared with 84.6\% for LTH after 292 epochs. On AlexNet/CIFAR-10 at 31.4\% pruning, FAIR-Pruner reaches 91.4\%, whereas LTH obtains 84.4\% after 260 epochs.

These results should not be interpreted as showing that FAIR-Pruner dominates all forms of iterative sparse training. Rather, they highlight a practically important point: a one-shot structured pruning framework guided by informative architectural and task-level signals can already produce subnetworks that are competitive with, and often stronger than, those obtained by much more expensive iterative procedures.

\begin{table}[t]
	\centering
	\caption{Normal inference efficiency on WiC. FLOPs are analytical estimates under the same generation protocol. Model size, GPU memory, and CPU RSS are reported in GB. FLOPs are reported as GFLOPs/example. GPU Mem. is PyTorch peak allocated CUDA memory during generation. CPU RSS denotes peak resident set size.}
	\label{tab:wic_normal_inference_efficiency}
	\footnotesize
	\setlength{\tabcolsep}{2pt}
	\begin{tabular}{lcccccc}
		\toprule
		Model & Kept & Acc. & FLOPs & Size & GPU Mem. & CPU RSS \\
		\midrule
		Full model & 60 (1440) & 57.68 & 347.0 & 26.67 & 26.76 & 48.03 \\
		FAIR 50\% & Non-unif. (720) & 56.74 & 346.8 & 15.06 & 15.09 & 31.06 \\
		FAIR 75\% & Non-unif. (360) & 48.28 & 346.7 & 9.26 & 9.29 & 19.47 \\
		FAIR 83.3\% & Non-unif. (240) & 36.83 & 346.6 & 7.32 & 7.36 & 15.55 \\
		\bottomrule
	\end{tabular}
\end{table}

\section{Pruning-time and Sample-size Sensitivity}
\label{app:sample_efficiency}

\begin{figure}[t]
	\centering
	\includegraphics[width=0.95\linewidth]{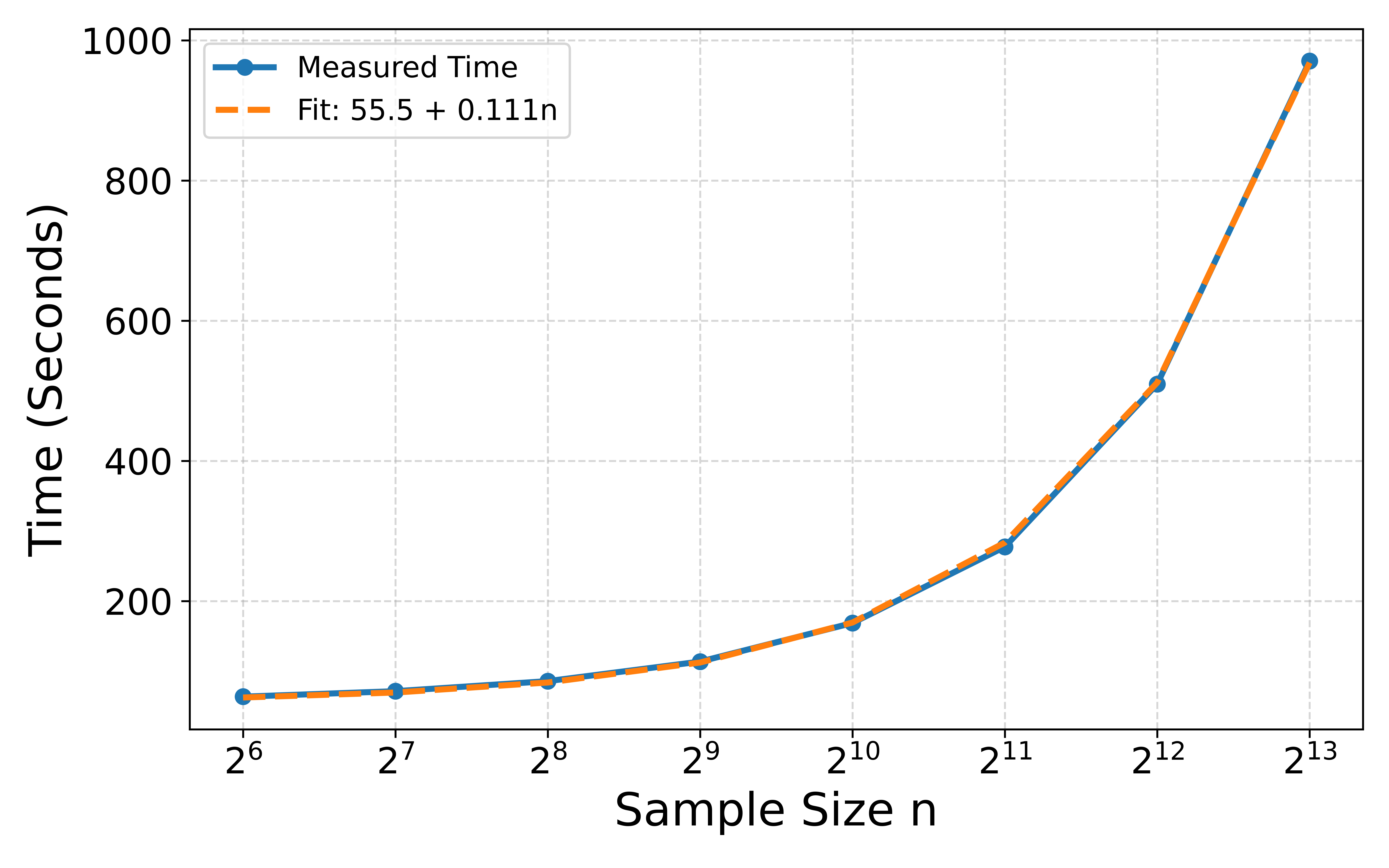}
	\caption{Sensitivity of FAIR-Pruner to the pruning-set sample size and the corresponding computational cost. Once U-Scores and R-Scores are computed, sweeping ToD levels is lightweight; the main cost comes from score estimation on the pruning subset.}
	\label{fig:sample_timecost_appendix}
\end{figure}

Figure~\ref{fig:sample_timecost_appendix} reports the sample-size and pruning-time sensitivity of FAIR-Pruner. 
The results show that score computation increases with the pruning-set size, while the ToD allocation step itself is lightweight after the scores have been obtained. 
This supports the practical use of FAIR-Pruner as a search-free pruning framework: multiple pruning budgets can be generated from the same score computation by varying the ToD level.

\vfill
\end{document}